\documentclass[10pt]{iopart}

\usepackage{cite}

\usepackage{subcaption}

\usepackage{setspace}
\usepackage{epstopdf}
\usepackage{trimclip}

\usepackage{iopams}
\expandafter\let\csname equation*\endcsname\relax
\expandafter\let\csname endequation*\endcsname\relax
\usepackage{amsmath}

\usepackage[font=small]{caption}

\usepackage{amssymb}
\usepackage[export]{adjustbox}

\usepackage{color, soul}
\makeatletter
\AtBeginDocument{\let\hl\@firstofone}
\makeatother

\usepackage{url}

\usepackage[ruled,lined]{algorithm2e}
\SetKwComment{Comment}{/* }{ */}

\usepackage{booktabs}

\usepackage{listings}

\usepackage{numprint}

\usepackage{siunitx}
\usepackage[nolist]{acronym}

\usepackage{breqn}

\usepackage[utf8]{inputenc}

\usepackage{todonotes}

\usepackage[shortcuts]{extdash}

\usepackage{textcomp}

\makeatletter
\newcommand\footnoteref[1]{\protected@xdef\@thefnmark{\ref{#1}}\@footnotemark}
\makeatother

\makeatletter
\def\lst@makecaption{%
  \def\@captype{table}%
  \@makecaption
  }
  \makeatother

  \makeatletter
  \def\BState{\State\hskip-\ALG@thistlm}
  \makeatother

  \usepackage[unicode]{hyperref}
  \hypersetup{
    colorlinks,
    linkcolor=black,
  }

\DeclareMathOperator*{\argmax}{argmax}

\newcommand{\vect}[1]{\mathbf{#1}}
\newcommand{\matr}[1]{\mathbf{#1}}

\newcommand{\mset}[1]{\mathcal{#1}}

\newcommand{\eg}{\textit{e.g.,}}
\newcommand{\ie}{\textit{i.e.,}}

\usepackage{pgfplots}
\usepackage{filecontents}
\pgfplotsset{compat=newest}
\pgfdeclarelayer{background}
\pgfdeclarelayer{foreground}
\pgfsetlayers{background,main,foreground}

\usepackage{tikz}
\pgfdeclarelayer{foreground}
\pgfsetlayers{background,main,foreground}
\usetikzlibrary{quotes, angles, backgrounds, trees, chains, arrows.meta, decorations.pathmorphing, arrows, automata, positioning, calc, through, spy, fit, intersections, decorations.markings, patterns, external, shapes.geometric, decorations.pathreplacing, shapes, matrix, shapes.symbols}

\tikzset{
    imglabel/.style={
      rectangle,
      inner sep=2pt,
      text=black,
      minimum height=1em,
      text centered,
      fill=white,
      fill opacity=1.0,
      text opacity=1,
      anchor=south west,
    },
  }

\tikzset{
	state/.style={
		rectangle,
		draw=black, very thick,
		minimum height=1.0em,
		text centered,
	},
}

\tikzset{external/system call={latex \tikzexternalcheckshellescape -halt-on-error
-interaction=batchmode -jobname "\image" "\texsource";
dvips -o "\image".eps "\image".dvi;
ps2eps "\image.eps"}}

\tikzset{
connect/.style args={(#1) to (#2) over (#3) by #4}{
  insert path={
    let \p1=($(#1)-(#3)$), \n1={veclen(\x1,\y1)},
    \n2={atan2(\x1,\y1)}, \n3={abs(#4)}, \n4={#4>0 ?180:-180}  in
    (#1) -- ($(#1)!\n1-\n3!(#3)$)
    arc (\n2:\n2+\n4:\n3) -- (#2)
  }
},
}

\tikzset{
    state/.style={
      rectangle,
      draw=black, very thick,
      minimum height=1.0em,
      text centered,
    },
    final_state/.style={
      rectangle,
      rounded corners,
      draw=black, very thick,
      minimum height=2em,
      text centered,
    },
    initial_state/.style={
      rectangle,
      double=white,
      double distance=1pt,
      inner sep=2pt,
      draw=black, very thick,
      minimum height=2em,
      text centered,
    },
    point/.style={
      circle,
      inner sep=0pt,
      minimum size=3pt,
      fill=red
    },
    adder/.style={
      circle,
      inner sep=2pt,
      minimum size=0.3in,
      draw=black, very thick,
      text centered
    }
}

\usepackage{enumerate}

\tikzset{
connect/.style args={(#1) to (#2) over (#3) by #4}{
  insert path={
    let \p1=($(#1)-(#3)$), \n1={veclen(\x1,\y1)},
    \n2={atan2(\x1,\y1)}, \n3={abs(#4)}, \n4={#4>0 ?180:-180}  in
    (#1) -- ($(#1)!\n1-\n3!(#3)$)
    arc (\n2:\n2+\n4:\n3) -- (#2)
  }
},
}

\usepackage{scalerel}
\usetikzlibrary{svg.path}
\definecolor{orcidlogocol}{HTML}{A6CE39}
\tikzset{
orcidlogo/.pic={
  \fill[orcidlogocol] svg{M256,128c0,70.7-57.3,128-128,128C57.3,256,0,198.7,0,128C0,57.3,57.3,0,128,0C198.7,0,256,57.3,256,128z};
  \fill[white] svg{M86.3,186.2H70.9V79.1h15.4v48.4V186.2z}
  svg{M108.9,79.1h41.6c39.6,0,57,28.3,57,53.6c0,27.5-21.5,53.6-56.8,53.6h-41.8V79.1z M124.3,172.4h24.5c34.9,0,42.9-26.5,42.9-39.7c0-21.5-13.7-39.7-43.7-39.7h-23.7V172.4z}
  svg{M88.7,56.8c0,5.5-4.5,10.1-10.1,10.1c-5.6,0-10.1-4.6-10.1-10.1c0-5.6,4.5-10.1,10.1-10.1C84.2,46.7,88.7,51.3,88.7,56.8z};
}
}
\newcommand\orcidicon[1]{\href{https://orcid.org/#1}{\mbox{\scalerel*{
\begin{tikzpicture}[yscale=-1,transform shape]
  \pic{orcidlogo};
\end{tikzpicture}
}{|}}}}

\newcommand\copyrighttext{%
	\small \begin{center} \color{red} \textcopyright\,2022 IOP Science. Personal use of this material is permitted. Permission from IOP Science must be obtained for all other uses, in any current or future media, including reprinting/republishing this material for advertising or promotional purposes, creating new collective works, for resale or redistribution to servers or lists, or reuse of any copyrighted component of this work in other works. \end{center}}

\begin{document}



\copyrighttext \vspace{-2em}
\title{PACNav: A Collective Navigation Approach for UAV Swarms Deprived of Communication and External Localization}


\author{
Afzal Ahmad$^{1\star{\orcidicon{0000-0002-5889-0320}}}$, Daniel Bonilla Licea$^{1\orcidicon{0000-0002-1057-816X}}$, Giuseppe Silano$^{2,1\orcidicon{0000-0002-6816-6002}}$, \hl{Tom\'{a}\v{s} B\'{a}\v{c}a}$^{1\orcidicon{0000-0001-9649-8277}}$, and Martin Saska$^{1\orcidicon{0000-0001-7106-3816}}$
}

\address{$^1$Department of Cybernetics, Faculty of Electrical Engineering, Czech Technical University in Prague, 16636, Prague 6, Czech Republic.}
\address{$^2$Department of Generation Technologies and Materials, Ricerca sul Sistema Energetico (RSE) S.p.A., 20134, Milan, Italy.}
\address{$^\star$Author to whom any correspondence should be addressed.}
\eads{
\mailto{ahmadafz@fel.cvut.cz}, \mailto{bonildan@fel.cvut.cz}, \mailto{giuseppe.silano@fel.cvut.cz}, \mailto{tomas.baca@fel.cvut.cz}, and
\mailto{martin.saska@fel.cvut.cz}
}
\vspace{10pt}
\begin{indented}
    \item[] Received 20 July 2022
    \item[] Revised 26 September 2022
    \item[] Accepted 7 October 2022
    \item[] Published 11 October 2022
\end{indented}




\begin{abstract}


This article proposes~\ac{PACNav} as an approach for achieving decentralized collective navigation of~\ac{UAV} swarms. The technique is based on the flocking and collective navigation behavior observed in natural swarms, such as cattle herds, bird flocks, and even large groups of humans. As global and concurrent information of all swarm members is not available in natural swarms, these systems use local observations to achieve the desired behavior. Similarly,~\ac{PACNav} relies only on local observations of relative positions of~\acp{UAV}, making it suitable for large swarms deprived of communication capabilities and external localization systems. We introduce the novel concepts of \textbf{path persistence} and \textbf{path similarity} that allow each swarm member to analyze the motion of other members in order to determine its own future motion.~\ac{PACNav} is based on two main principles: (1)~\acp{UAV} with little variation in motion direction have high \textbf{path persistence}, and are considered by other~\acp{UAV} to be reliable leaders; (2) groups of~\acp{UAV} that move in a similar direction have high \textbf{path similarity}, and such groups are assumed to contain a reliable leader. The proposed approach also embeds a reactive collision avoidance mechanism to avoid collisions with swarm members and environmental obstacles. This collision avoidance ensures safety while reducing deviations from the assigned path. Along with several simulated experiments, we present a real-world experiment in a natural forest, showcasing the validity and effectiveness of the proposed collective navigation approach in challenging environments. The source code is released as open-source, making it possible to replicate the obtained results and facilitate the continuation of research by the community. 

\end{abstract}


\vspace{2pc}

\submitto{\BB}

\maketitle

\ioptwocol



\section{Introduction}
\label{sec:introduction}

\begin{figure}[tb]
  \centering
  \includegraphics[width=0.48\textwidth]{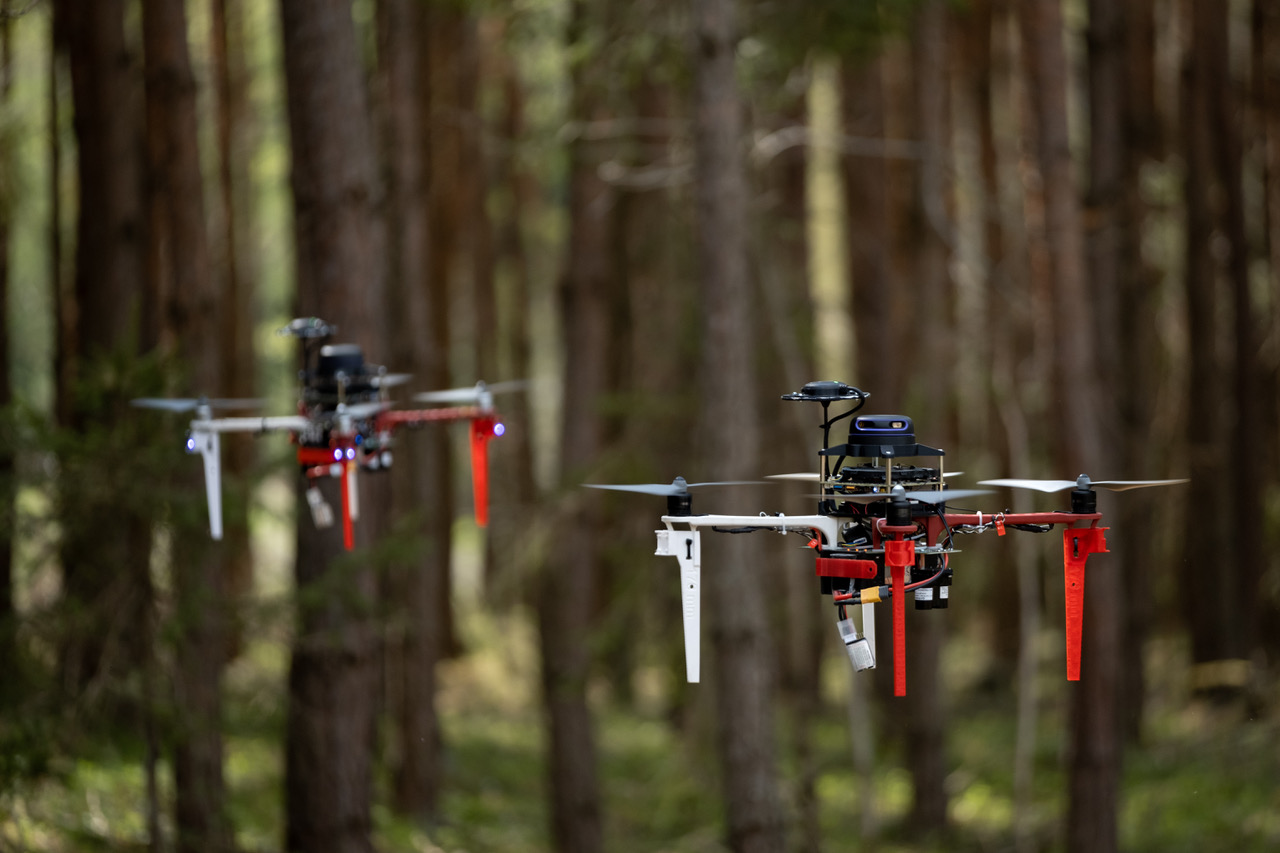}
  \caption{Picture from the flight experiment performed in a natural forest.}
  \label{fig:forest_flight}
\end{figure}

The collective motion of a tightly cooperating group of~\acp{UAV} has been intensively investigated in recent years~\cite{ZhouScienceRobotics2022, kumar_2018_tro}. The use of a group of~\acp{UAV} can reduce mission time and provide the redundancy and safety that is critical in many real-world applications, such as mapping large areas~\cite{Honig2018TRO}, construction~\cite{Augugliaro2014CSM}, agriculture~\cite{ElmokademIFAC2019}, and search-and-rescue missions~\cite{McGuireScienceRobotics2019}. These applications have further motivated research on the collective motion of a group of~\acp{UAV}~\cite{tagliabue_2019_sage, Inada2010IFAC, Novak2021Bioinspired}. Indeed, large groups of~\acp{UAV} are effective in some scenarios where facing the problem with a single robot may be unfeasible or difficult to solve.~\ac{UAV} swarms have also been useful for applications where redundancy is needed to cope with failure of individual~\acp{UAV}~\cite{rabbath_2012_autocontrol} and in cases when multiple~\acp{UAV} need to gather data simultaneously \cite{petracek_2020_ral}.

The deployment of a group of~\acp{UAV} requires a complex system composed of several intricate subsystems handling vehicle control, environment perception, absolute or relative localization, mapping, navigation, and communication. Therefore, the use of centralized control architectures which require~\acp{UAV} to communicate and exchange information over a shared network may explode in complexity as the number of vehicles and the size of the working area increases. Thus, it is important to introduce decentralized control architectures, fault detection systems, and feasible information sharing over a low-to-none bandwidth network.

Animals, including fish and birds, are an important source of inspiration to tackle this problem due to their use of decentralized decision making for collective motion~\cite{vicsek_2010_nature, couzin_2007_nature, yomosa_2015_plos}. Reynolds~\cite{reynolds_boids} described this motion using a set of simple rules addressing attraction to the group, repulsion from neighbors, and alignment to move in the same direction as the group. These rules have since been adapted for use in robotic swarms with additional components for obstacle avoidance~\cite{olfati_2006_tac}. The simplicity of the rules and the decentralized nature of the control strategy provides robustness against the failure of individual robots. In many cases, the decentralized decision making system only depends on local information about the neighbors, which makes these methods scalable to a large number of robots.

Decentralized control of a swarm of~\acp{UAV} can also be achieved using predictive controllers and trajectory planning. The swarm of~\acp{UAV} in~\cite{cheng_2017_iros} uses reactive avoidance and~\ac{MPC} strategies for collective navigation. On the other hand,~\cite{gao_2021_icra} presented a nonlinear optimization method for decentralized trajectory planning. However, these methods require a shared reference frame to localize all the~\acp{UAV} and often rely on shared information about the position and future trajectory of other~\acp{UAV}. Establishing a shared reference frame and communication links is often unfeasible or difficult to achieve in practice due to unreliable~\ac{GNSS} signals, as well as losses in wireless communication caused by environmental occlusions and reflections. Thus, dependence on localization and communication infrastructure can severely limit the use of these methods in environments cluttered with obstacles. In cluttered spaces such as forests and construction sites, it is common practice to use onboard sensors for the localization and relative pose estimation of other~\acp{UAV}. Although these sensors often have a higher computational demand than those used for global localization and communication, they can provide information at a rate sufficient for stable motion.
The work in~\cite{vicek_2018_scirob} presents a decentralized swarm that does not use communication, but still relies on a shared global reference frame (\ac{GNSS}). In our previous works~\cite{afzal_2021_icra, Dmytruk2021ICUAS}, we described a~\ac{UAV} swarm that navigates towards a goal completely independent of any shared reference frame and communication infrastructure. However, both~\cite{afzal_2021_icra, Dmytruk2021ICUAS} do not perform well in the presence of occlusions from the surrounding obstacles and scale poorly with a growing number of~\acp{UAV}.

Dependence on communication has a negative influence on the scalability of the swarm. Moreover, sharing a reference frame may require exchange of mapping information and synchronization in \ac{GNSS}-denied environments~\cite{Petracek2021RAL}, which has adverse effects on robustness. Therefore, this article proposes a bioinspired decentralized approach for collective navigation of a swarm of~\acp{UAV} without using~\ac{GNSS} and any communication. The~\acp{UAV} are controlled using only onboard sensor data, which is used for detecting and localizing obstacles and other team members. The proposed approach builds upon the analysis of collective motion of groups of animals and humans~\cite{wang_2017_aaai} in order to design path similarity and path persistence metrics for comparing the trajectories of~\acp{UAV}. Target~\acp{UAV} are selected based on these metrics and consequently followed by individual~\acp{UAV}, resulting in a collective motion. The approach does not rely on any prior information about the environment or the swarm and can be deployed in an unknown environment cluttered with obstacles. The collision avoidance mechanism proposed as part of the approach places high emphasis on safety and is specifically designed to operate in complex unknown environments. In particular, collision avoidance is designed in such a way so as to reduce deviations from the assigned paths by only reacting to the~\acp{UAV} showing an immediate threat of collision. This is similar to the emergency instincts which are often observed in swarms of animals in nature. Simulated experiments and real-world flight in a natural forest have been used to validate and analyze the performance and robustness of the proposed approach. Compared to our previous works~\cite{afzal_2021_icra, Dmytruk2021ICUAS}, the approach proposed in this article uses relative localization information, recorded over a finite time horizon, to overcome the challenges of occlusion of~\acp{UAV} by obstacles. The source code related to this work has been released as open-source\footnote{\url{https://github.com/ctu-mrs/pacnav}\label{fotnote:code}} for easy replication and future work by the community. Additionally, videos from the simulations and real-world experiments have been provided as supplementary multimedia material and made available at~\url{http://mrs.felk.cvut.cz/pacnav}.



\subsection{Related works}
\label{sec:sota}

\ac{UAV} swarms have been a topic of several studies and have recently gained a lot of attention due to their useful properties, such as adaptability, scalability, reliability, and fault-tolerance~\cite{kumar_2018_tro, swarm_survey2, swarm_survey_mean_field}. The idea of using group intelligence stems from natural biological systems, such as flocks of birds, schools of fish, or swarms of bees. One of the first simulated models of the natural flocking behavior of birds was introduced in~\cite{reynolds_boids}, where the motion of a swarm of dimensionless particles was controlled by a simple set of rules.~\cite{kumar_2018_tro} describes several different methods to control a swarm of~\acp{UAV}, including physics-based models and~\ac{MPC}. However, most of the recent research has been limited to laboratory-like conditions, as presented in~\cite{lablike_swarm1, lablike_swarm2, lablike_swarm3, non_obst_swarm1}.

\subsubsection{No obstacle avoidance}
\label{subsec:noObstacleAvoidance}

Recent works presented in~\cite{vicek_2018_scirob, comm_swarm, vio_swarm_kumar} study the real-world deployment challenges of~\ac{UAV} swarms.~\cite{vicek_2018_scirob} uses the method developed in~\cite{reynolds_boids} for controlling a swarm of~\acp{UAV} in a real-world environment, where each~\ac{UAV} uses~\ac{GNSS} for localization and shares this position information with other~\acp{UAV} over a communication network. The swarm in~\cite{comm_swarm} also uses~\ac{GNSS} and communication for flocking, but has additional constraints on energy consumption. The reliance of~\cite{vicek_2018_scirob, comm_swarm} on~\ac{GNSS} makes them unsuitable for several real-world scenarios where~\ac{GNSS} is unreliable or unavailable,~\eg~forests and indoor construction sites.~\cite{vio_swarm_kumar} presented a swarm of~\acp{UAV} for~\ac{GNSS}-denied environments that uses inter-\ac{UAV} communication to control the swarm. However, inter-\ac{UAV} communication scales poorly with a growing number of~\acp{UAV} in the swarm and can become a bottleneck in cluttered environments where obstacles reduce the communication range. The work presented in~\cite{non_obst_swarm1} uses a  computer vision based technique for relative localization of~\acp{UAV}. The use of on-board sensors for relative localization makes this method independent of any communication infrastructure and associated scalability issues. However,~\cite{non_obst_swarm1,vicek_2018_scirob, comm_swarm, vio_swarm_kumar} do not implement any collision avoidance, thus limiting the deployment to only obstacle-free environments.

\subsubsection{Obstacle avoidance with relative localization and communication}
\label{subsec:obstacleAvoidanceRelativeLocalizationCommunication}

The swarm system presented in~\cite{gao_2021_icra} uses cameras for relative localization and obstacle avoidance in a cluttered environment. The motion planning problem of moving the~\acp{UAV} towards a goal is solved as a nonlinear optimization problem on-board each vehicle. However, this method uses a broadcast communication network to share~\acp{UAV}' trajectories for collision avoidance between vehicles. The method presented in~\cite{pavel_swarm} uses a~\ac{UV} light based visual relative localization system for the~\acp{UAV}. Cameras placed on the~\acp{UAV} localize other~\acp{UAV} in a relative reference frame, removing the need for sharing position information. Although the swarm in~\cite{pavel_swarm} can avoid static obstacles, the real-world experiments use artificial obstacles and very low environment density. Moreover, the obstacle positions are known \textit{a priori}, which is difficult to achieve in arbitrary real-world deployment scenarios. The bioinspired method introduced in~\cite{olfati_2006_tac} models the obstacles as agents or as a group of agents depending on the obstacles number. This approach is easy to implement, but relies on precise obstacle positions and shape estimates which are not trivial to obtain in complex environments. Furthermore, such an approach can often create a virtual deadlock when the~\ac{UAV} agent is surrounded by several obstacles, as is common in cluttered real-world environments, such as forests or indoor construction sites.





\subsection{Contributions}

This article proposes~\acf{PACNav} as a decentralized approach to navigating a~\ac{UAV} swarm without communication and without global localization infrastructure from its initial position to a goal point. We address several challenges related to the real-world deployment of a swarm of~\acp{UAV} and their collective motion in a cluttered environment. The main contributions going beyond the previously presented literature include:

\begin{itemize}

  \item a bioinspired decentralized approach to infer goal direction using motion of~\acp{UAV} observed directly from on-board sensors, similar to the sensory organs of animals. This approach is based on novel concepts of path persistence and path similarity which make it resilient to sensor uncertainties observed during real-world deployment;
  
  \item an approach for motion planning and reactive collision avoidance to safely navigate the swarm in cluttered environments without communication among the~\acp{UAV} and any global localization system.
  %
  

\end{itemize}





\subsection{Notation}
\label{sec:notation}

This article will use $[\matr{A}]_{i}$ to denote the $i$-th column of a matrix $\mathbf{A}$, and $[\matr{A}]_{ij}$ to denote the $i$-th row and the $j$-th column of the matrix $\mathbf{A}$. For any two matrices $\mathbf{A}$ and $\mathbf{B}$, $[\mathbf{A},\mathbf{B}]$ is used to represent column concatenation. If $\mathcal{S}$ is a set, then $|\mathcal{S}|$ denotes its cardinality. For any two vectors $\mathbf{a}$ and $\mathbf{b}$, their inner product is written as $\mathbf{a} \cdot \mathbf{b}$. The symbol $\lVert \bullet \rVert$ denotes the Euclidean norm.




\section{System Model}
\label{sec:sys_model}

We consider a swarm composed of $N$-\acp{UAV} that move on a horizontal plane ($XY$-plane) and assume that we can directly control their velocities along these axes.
Thus, the system dynamics of the $i$-th~\ac{UAV}, at discrete time index $k$, can be described with a point-mass model as:
\begin{align}
    \mathbf{\dot{p}}_{i}[k] = \vect{u}_{i}[k],
\label{eq:dyn}
\end{align}
where $\mathbf{p}_{i}[k] \in \mathbb{R}^2$ is the position in the world coordinate frame $\mathcal{F}_W$, and $\mathbf{u}_{i}[k] \in  \mathbb{R}^2$ is the velocity control input expressed in the same reference system ($\mathcal{F}_W)$.

We assume that~\acp{UAV} in the swarm do not have prior information about other~\acp{UAV} and cannot communicate among themselves. Each~\ac{UAV} is equipped with an omnidirectional camera, allowing for position estimation of the other~\acp{UAV} using an image processing method~\cite{WalterIEEECASE2018, WalterIEEERAL2019, VrbaIEEERAL2020}. Obstacles in the  environment may occlude the camera view and, consequently, direct observation of other~\acp{UAV}' positions is not always possible. Thus, the estimate $\mathbf{\check{p}}_{ij}[k]$ of the $j$-th~\ac{UAV} position when observed by the $i$-th~\ac{UAV}, at discrete time $k$, is modeled as: 

\begin{subequations}\label{eq:pos_est}
    \begin{align}
    \mathbf{\check{p}}_{ij}[k] & = \mathbf{q}_{ij}[k] + \mathbf{\eta}_{ij}[k], \\
    \mathbf{q}_{ij}[k] & = \mathbf{p}_{ij}[k]f_{ij}[k] + \mathbf{\check{p}}_{ij}[k-1] \bar{f}_{ij}[k],
    \end{align}
\end{subequations}

where $\mathbf{p}_{ij}[k] \in \mathbb{R}^2$ is the relative ground truth position of the $j$-th~\ac{UAV} with respect to the $i$-th~\ac{UAV}. $f_{ij}[k]=1$ if there is~\ac{LoS} between the $i$-th and $j$-th~\acp{UAV} and $f_{ij}[k]=0$ otherwise. $\bar{f}_{ij}[k]$ is the complementary function of $f_{ij}[k]$. $\mathbf{\eta}_{ij}[k]\in\mathbb{R}^2$ is the estimation error which is assumed to be a Gaussian random process with zero-mean and covariance matrix $\sigma^2[k]\mathbf{I}$, where $\mathbf{I} \in \mathbb{R}^{2 \times 2}$ represents the identity matrix, and:
\begin{equation}\label{eq:noise_type}
\sigma^2[k] = f_{ij}[k] \sigma_{\mathrm{LoS}}^2 + \bar{f}_{ij}[k] \sigma_{\mathrm{NLoS}}^2.    
\end{equation}
The terms $\sigma_{\mathrm{LoS}}$ and $\sigma_{\mathrm{NLoS}}$ capture the different sources of estimation error. In the case of~\ac{LoS}, the estimation error is mostly influenced by the sensor noise. However, when the~\ac{LoS} is lost, factors, including UAV motion or other random processes, further contribute to the estimation error. For simplicity, we assume omnidirectional sensing, although \eqref{eq:pos_est} can be adapted to consider any other directional sensor.

When there is~\ac{LoS} between the $i$-th and the $j$-th~\acp{UAV}, the error in the position estimate is modeled by the additive Gaussian noise (see~\eqref{eq:pos_est}) with zero-mean and covariance matrix $\sigma_{\mathrm{LoS}}^2 \mathbf{I}$. Depending on the available~\ac{LoS}, the $i$-th~\ac{UAV} performs two actions: i) it adds the index $j$ to the set of neighbors $\mathcal{N}_i[k]$ whenever~\ac{LoS} is obtained between them. However, when the~\ac{LoS} is lost, the estimate $\mathbf{\check{p}}_{ij}$ becomes an autoregressive process which provides feedback to the noise process $\mathbf{\eta}_{ij}$. Its expected value is the last known position of the $j$-th~\ac{UAV}, and its covariance matrix grows linearly at a rate of $\sigma_{\mathrm{NLoS}}^2$ per sampling instant; ii) the $i$-th~\ac{UAV} removes the index $j$ from $\mathcal{N}_i[k]$ if the~\ac{LoS} with the $j$-th~\ac{UAV} is lost for more than $K^m$ time instants, where $K^m \in \mathbb{R}$ is a design parameter. A $\delta_{ij}$ variable keeps track of the latest known time instant when there was~\ac{LoS} between the $i$-th and $j$-th~\acp{UAV}. Algorithm~\ref{alg:neighbor_update} describes the update of $\mathcal{N}_i[k]$ in more detail.



\begin{algorithm}[tb]
\caption{Updating set $\mathcal{N}_i[k]$}
\label{alg:neighbor_update}
\KwData{$\mathcal{N}_i[k-1], f_{ij}[k], \delta_{ij}[k-1]$}
\KwResult{$\mathcal{N}_i[k]$}
\vspace{0.25cm}
\Comment{add indices of~\acp{UAV} that have~\ac{LoS} with the $i$-th~\ac{UAV} at time instant $k$}
$\mathcal{N}_i[k] \leftarrow \mathcal{N}_i[k-1] \cup \{ j : f_{ij}[k] = 1,j\notin\mathcal{N}_i[k-1] \}$ \\    
\vspace{0.15cm}
\For{$j \in \mathcal{N}_i[k]$}{
  \eIf{$f_{ij}[k] = 1$}{
        \Comment{$\delta_{ij}$: latest recorded time when there was~\ac{LoS} between the $i$-th and $j$-th~\acp{UAV}}
    $\delta_{ij}[k] \leftarrow k$
  }{
    $\delta_{ij}[k] \leftarrow \delta_{ij}[k-1]$
  }
  \vspace{0.15cm}
  \Comment{remove $j$ from $\mathcal{N}_i[k]$ if~\ac{LoS} with the $i$-th~\ac{UAV} has been lost for more than $K^m$ time instants}
  \If{$ k - \delta_{ij}[k] > K^m$}{
    $\mathcal{N}_i[k] \leftarrow \mathcal{N}_i[k] \setminus j$
  }
}
\end{algorithm}



The approach proposed in this article relies on a sequence of position estimates which are stored in a matrix $\mathbf{H}_{ij}[k]$, called the \textit{path history} matrix. When index $j$ is added to $\mathcal{N}_i[k]$, the $i$-th~\ac{UAV} starts to store the position estimates $\mathbf{\check{p}}_{ij}[k]$ into a path history matrix $\mathbf{H}_{ij}[k]$, along with the corresponding time instant of the estimate into the matrix $\mathbf{\Gamma}_{ij}[k]$. Algorithm~\ref{alg:path_hist} describes the update process for $\mathbf{H}_{ij}[k]$, where the estimates are sequentially concatenated in $\mathbf{H}_{ij}[k]$. $[\mathbf{H}_{ij}]_1$ contains the newest position estimate of the $j$-th~\ac{UAV}, and $[\mathbf{H}_{ij}]_{L_{ij}}$ the oldest one, where ${L_{ij}} \in \mathbb{N}_{>0}$ is the number of columns of the matrix $\mathbf{H}_{ij}$. When an estimate is older than the design parameter $K^p \in \mathbb{Z}_{>0}$, it is removed from $\mathbf{H}_{ij}[k]$, and so ${L_{ij}}\leq K^p$.

\begin{algorithm}[tb]
\caption{Updating $\mathbf{H}_{ij}[k]$}
\label{alg:path_hist}
\KwData{$\mathbf{H}_{ij}[k-1], \mathbf{\Gamma}_{ij}[k-1], \mathbf{\check{p}}_{ij}[k]$}
\KwResult{$\mathbf{H}_{ij}[k]$}
\vspace{0.25cm}
\eIf{$j \in \mset{N}_i[k]$}{
  $\mathbf{H}_{ij}[k] \leftarrow \Big[ \mathbf{\check{p}}_{ij}[k], \mathbf{H}_{ij}[k-1] \Big]$ \\
  $\mathbf{\Gamma}_{ij}[k] \leftarrow \Big[ k, \mathbf{\Gamma}_{ij}[k-1] \Big]$
}{
  $\mathbf{H}_{ij}[k] \leftarrow \mathbf{H}_{ij}[k-1]$ \\
  $\mathbf{\Gamma}_{ij}[k] \leftarrow \mathbf{\Gamma}_{ij}[k-1]$
}
\vspace{0.15cm}
\Comment{remove last column if older than $K^p$ time instances}
\If{$ k - \Big[ \mathbf{\Gamma}_{ij}[k] \Big]_{L_{ij}} > K^p$}{
  \vspace{0.10cm}
  remove$\Bigg( \Big[ \mathbf{H}_{ij}[k] \Big]_{L_{ij}}, \Big[ \mathbf{\Gamma}_{ij}[k] \Big]_{L_{ij}} \Bigg)$\\
}

\end{algorithm}


We now use the previously described data structures to introduce some metrics and notation that will be helpful for the description of the proposed~\ac{PACNav} approach detailed in the following sections.

\textbf{Definition 1 (Path Similarity):} This metric describes the similarity between the motion of two different~\acp{UAV}. We define the path similarity between the $j$-th and $l$-th~\acp{UAV} when observed by the $i$-th~\ac{UAV} as follows:
\begin{align}\label{eq:path_sim}
    \sigma_{ijl} &\triangleq \frac{1}{L-1}  \sum_{m = 1}^{L-1}  \frac{\mathbf{h}^m_{ij} \cdot \mathbf{h}^m_{il}}{\lVert \vect{h}^m_{ij} \rVert \lVert\vect{h}^m_{il} \rVert},\\
    \mathbf{h}^m_{ij} &= [\mathbf{H}_{ij}]_{m} - [\mathbf{H}_{ij}]_{m+1}, 
\end{align}
where $\{\mathbf{h}^m_{ij}\}_{m=1}^{L-1}$ is the recent history of the displacement of the $j$-th~\ac{UAV}, estimated by the $i$-th~\ac{UAV}, and $L = \min(L_{ij}, L_{il})$. The path similarity $\sigma_{ijl} \in [-1, 1]$ in~\eqref{eq:path_sim} is the moving average of the inner product between the estimated displacement vectors of the $j$-th and $l$-th~\acp{UAV}. $\blacksquare$

Note that, when both~\acp{UAV} move in the same direction during the previous $L$ time instants, then $\sigma_{ijl}=1$; if they move in opposite directions, then $\sigma_{ijl}=-1$; and, if they move in orthogonal directions, then $\sigma_{ijl}=0$.

The \textit{path similarity} $\sigma_{ijl}$ compares two different paths, but it does not provide information about the individual path. To this end, we introduce the following metric.

\textbf{Definition 2 (Path Persistence):} This metric describes variability in the direction of motion of a~\ac{UAV}. We define the \textit{path persistence} of the $j$-th~\ac{UAV} when observed by the $i$-th~\ac{UAV} as:
\begin{align}\label{eq:path_pers}
    \gamma_{ij} &\triangleq \frac{1}{L_{ij}-2} \sum_{m = 1}^{L_{ij}-2} \frac{\mathbf{h}^{m+1}_{ij}\cdot \mathbf{h}^m_{ij}}{\lVert \mathbf{h}^{m+1}_{ij} \rVert \lVert \mathbf{h}^m_{ij} \rVert}.
\end{align}

The path persistence $\gamma_{ij}$ is the moving average of the inner product between all consecutive displacement estimations of the $j$-th~\ac{UAV} observed by the $i$-th~\ac{UAV}. It measures how much the $j$-th~\ac{UAV} motion direction has changed recently. $\blacksquare$

When the $j$-th~\ac{UAV} moves in a straight line, then~\eqref{eq:path_pers} is maximized and $\gamma_{ij}=1$. On the other hand, when it moves in a random fashion and the velocity often changes, $\gamma_{ij}$ will have lower values. From~\eqref{eq:path_pers}, we can see that the argument of the sum is the inner product between the normalized vectors $\mathbf{h}^{m+1}_{ij}$ and $\mathbf{h}^m_{ij}$. Thus, $\frac{\mathbf{h}^{m+1}_{ij}\cdot \mathbf{h}^m_{ij}}{\lVert \mathbf{h}^{m+1}_{ij} \rVert \lVert \mathbf{h}^m_{ij} \rVert}\in[-1,1]$. Since~\eqref{eq:path_pers} is the average of these normalized inner products, we obtain $\gamma_{ij} \geq -1$.

For the purpose of the proposed approach, each~\ac{UAV} in the swarm will belong to one of the two categories:

\begin{enumerate}
    
    \item \textbf{Informed~\ac{UAV}} which knows the goal location $\mathbf{g} \in \mathbb{R}^2$ and plans a path to reach it. Because of this information, its trajectory will present, in general, a small number of direction changes (as small as the environment allows), \ie~a high path persistence $\gamma_{ij}$.
    
    \item \textbf{Uninformed~\ac{UAV}} which does not know the goal $\mathbf{g}$. It will observe the motion of the other~\acp{UAV} within the swarm and use it to adaptively move in order to reach the goal. As a consequence, the uninformed UAVs will initially have more irregular motion with many directional changes, resulting in low path persistence, \ie~low values of $\gamma_{ij}$.
\end{enumerate}

Finally, we denote $\mathcal{I}$ the set that contains the indices of all the informed~\acp{UAV}, and $\mathcal{\bar{I}}$ the set that contains the indices of all the uninformed~\acp{UAV}.



\section{Problem Statement}
\label{sec:problem}

We consider the problem of navigating a~\ac{UAV} swarm, deprived of communication and global localization, in an environment with randomly distributed obstacles. The~\acp{UAV} start at random locations inside a circle of radius $R^s \in \mathbb{R}_{>0}$. Their mission is accomplished once all the~\acp{UAV} are inside the disk of radius $R^g \in \mathbb{R}_{>0}$, centered at the goal location $\mathbf{g}$. Some randomly selected~\acp{UAV} belong to the informed category described before, while the rest of the~\acp{UAV} belong to the uninformed category. The~\ac{UAV} categories remain fixed during the whole mission execution.

As mentioned before, we assume that the~\acp{UAV} are equipped with on-board omnidirectional sensors~\cite{WalterIEEECASE2018, WalterIEEERAL2019, VrbaIEEERAL2020, Horyna2022ICUAS_UVDAR} that allow for estimating the position of the surrounding \acp{UAV}, as well as detecting their IDs (IDs simplify the separation of multiple observed neighbor~\acp{UAV} and retrieval of their relative position). We also assume that the individual~\acp{UAV} are equipped with the necessary sensors to implement~\ac{SLAM}~\cite{Kohlbrecher2011ISSSRR} for localization and navigation.

We aim to design a decentralized control approach for~\acp{UAV} that only uses on-board sensors and computational resources to complete the collective navigation mission described above. Since the~\acp{UAV} are deprived of communication, they cannot exchange information like the goal position $\mathbf{g}$ or the category of individual~\acp{UAV} (informed or uninformed). Thus, the uninformed~\acp{UAV} must use the observed~\acp{UAV}' trajectories to devise a motion plan to reach $\mathbf{g}$. In addition,~\acp{UAV} must avoid collisions not only with the obstacles populating the environment, but also among themselves.

To this end, this article proposes the decentralized control approach~\ac{PACNav}, which is composed of two modules. The first module, described in Section~\ref{sec:dyn_target}, iteratively determines a target to be followed by the~\ac{UAV}. The second module, described in Section~\ref{sec:track_target}, controls the velocity of the~\ac{UAV} ($\mathbf{u}_i$) to reach the target provided by the previous module while avoiding collisions.




\section{Iterative Target Selection}
\label{sec:dyn_target}

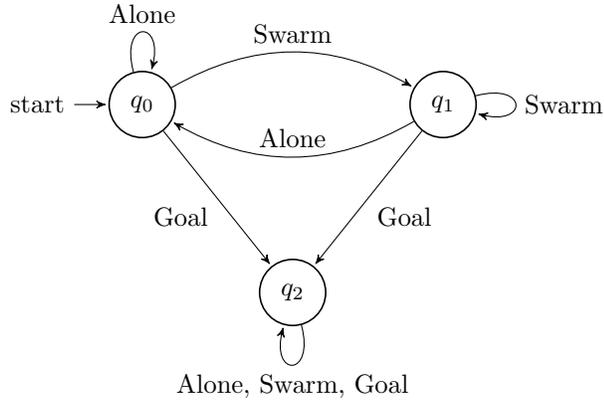
\begin{figure}[tb]
    \begin{center}
	\begin{tikzpicture}[>=stealth', 
	                    state/.append style = {circle, semithick, align=center, inner sep=2pt, minimum size=25pt}, 
	                    shorten >=1pt, 
	                    node distance=3.5cm, 
	                    on grid, 
	                    auto] 
	
	\node[state, initial] (q_0)   {$q_0$}; 
	\node[state] (q_1) [right=0.7cm and 4cm of q_0] {$q_1$}; 
	\node[state] (q_2) [below right=2.5cm and 2cm of q_0] {$q_2$};
	
	\path[->] 
		(q_0) edge [bend left]	node 		      {\text{Swarm}} (q_1)
		      edge [loop above] node              {\text{Alone}} ()
		      edge              node[below left]  {\text{Goal}}  (q_2)
		(q_1) edge [bend left]	node[above]       {\text{Alone}} (q_0)
		      edge [loop right]	node    	      {\text{Swarm}} ()
		      edge 				node 		      {\text{Goal}} (q_2)
		(q_2) edge [loop below]	node 		      {\text{Alone},~\text{Swarm},~\text{Goal}} ();
					
	\end{tikzpicture}
    \caption{Finite state machine $\mathcal{M}$ for dynamic target selection.}
    \label{fig:target_machine}
    \end{center}
\end{figure}

At each time instant $k$, the $i$-th~\ac{UAV} determines a target location $\mathbf{d}_i[k] \in \mathbb{R}^2$ and devises a path to reach it. This target can be the goal position $\mathbf{g}$ or the position of a neighboring~\ac{UAV} potentially moving towards the goal $\mathbf{g}$. Thus, moving towards a target~\ac{UAV} can lead the $i$-th UAV to the goal. This section discusses how to determine the target $\mathbf{d}_i[k]$. Let us start by defining the finite state machine $\mathcal{M}~=~( \mathcal{S}, \mathcal{Q}, \Delta, q_0, \mathcal{L} )$, where:

\begin{itemize}
    \item $\mathcal{S} = \{ \text{Alone}, \text{Swarm}, \text{Goal} \}$ is the input alphabet,
    \item $\mathcal{Q} = \{ q_0, q_1, q_2 \}$ represents the set of all possible states of the state machine $\mathcal{M}$ (illustrated in Fig.~\ref{fig:target_machine}),
    \item $\Delta$ is the state transition function of $\mathcal{M}$,  
    \item $q_0$ is the initial state, and
    \item $\mathcal{L} = \varnothing$ is the set of final states. As the machine $\mathcal{M}$ is designed to run indefinitely, there are no final states.
\end{itemize}  

$x_i[k] \in \mathcal{S}$ is the input to the state machine $\mathcal{M}$ of $i$-th~\ac{UAV} at time $k$, and it is given by:
\begin{align}\label{eq:m_state}
    x_i[k] = 
    \begin{cases}
      \text{Alone} &, \, \lvert \mset{T}_i[k] \rvert = 0 \wedge i \in \mathcal{\bar{I}}, \\
      \text{Swarm} &, \, \lvert \mset{T}_i[k] \rvert > 0 \wedge i \in \mathcal{\bar{I}}, \\
      \text{Goal} &, \, i\in\mathcal{I},
    \end{cases}
\end{align}
where $\mathcal{T}_i[k] \subseteq \mset{N}_i[k]$ contains the indices of \acp{UAV} that are the potential targets of the uninformed $i$-th~\ac{UAV}. The objective is to design $\mathcal{T}_i[k]$ so that it contains the indices of the~\acp{UAV} which are potentially moving towards the goal $\mathbf{g}$. These~\acp{UAV} are determined based on the following criteria:

\begin{itemize}
    
    \item \acp{UAV} that are not in close proximity of the $i$-th~\ac{UAV}. The trajectories of the~\acp{UAV} close to the $i$-th~\ac{UAV} are mostly influenced by the collision avoidance mechanism (to avoid collision with the $i$-th~\ac{UAV}). Consequently, at that moment, such trajectories have little information about the motion towards a target. So, the $i$-th~\ac{UAV} discards any~\ac{UAV} that is too close, and it considers only~\acp{UAV} beyond a certain distance. In other words, the $j$-th~\ac{UAV} has to satisfy $\lVert [\mathbf{H}_{ij}[k]]_1 - \mathbf{p}_i[k] \rVert \geq R^f$, with $R^f \in \mathbb{R}_{>0}$ being a design parameter, to be considered a potential target by the $i$-th~\ac{UAV}.
      
    \item \acp{UAV} that are not moving towards the previous target position $\mathbf{d}_i[k-1]$. Since the target is one of the~\acp{UAV} in $\mathcal{T}_i[k]$, any~\acp{UAV} moving towards the previous target $\mathbf{d}_i[k-1]$ will not change the current direction of motion of the $i$-th~\ac{UAV}. Thus, the $i$-th~\ac{UAV} discards any $j$-th~\ac{UAV} that satisfies $\lVert [\mathbf{H}_{ij}[k]]_1 - \mathbf{d}_i[k-1] \rVert < \lVert [\mathbf{H}_{ij}[k]]_{L_{ij}} - \mathbf{d}_i[k-1] \rVert$.
    
    \item \acp{UAV} whose path history $\mathbf{H}_{ij}[k]$ contains at least three elements. The target is determined using the path persistence metric ($\gamma_{ij}$), which needs at least three elements in the path history (see~\eqref{eq:path_pers}).

\end{itemize}
The set of potential targets $\mathcal{T}_i[k]$ contains the indices of the~\acp{UAV} that satisfy all three conditions described above.  

Given the current input $x_i[k]$, the state machine $\mathcal{M}$ transitions into a state $s_i[k] \in \mathcal{Q}$. If $s_i[k]= q_2$ ($i$-th~\ac{UAV} is informed), then the target is the goal position, \ie~$\mathbf{d}_i[k]=\mathbf{g}$. If $s_i[k]= q_0$ ($i$-th~\ac{UAV} is uninformed and its set of potential targets is empty), then the~\ac{UAV} does not move, \ie~$\mathbf{d}_i[k]=\mathbf{p}_i[k]$. However, if $s_i[k]= q_1$ ($i$-th~\ac{UAV} is uninformed and its set of potential targets is not empty), then it proceeds to select $\mathbf{d}_i[k]$ using the potential target set $\mathcal{T}_i[k]$. This selection process is based on the path similarity ($\sigma_{ijl}$) and persistence ($\gamma_{ij})$ metrics defined in Section~\ref{sec:sys_model}. First, note that the goal $\mathbf{g}$ is the target position for informed~\acp{UAV} and they are following a path that will lead them towards it. Any changes of direction in their movement is mainly due to the collision avoidance mechanism reacting to the environment. As a result, the informed~\ac{UAV} path history, in general, will present high path persistence. Since all informed~\acp{UAV} have $\mathbf{g}$ as their target, then their path similarity will also be large. Therefore, if an uninformed~\ac{UAV} follows a~\ac{UAV} which has large path persistence and path similarities with other~\acp{UAV}, then it would very likely reach the goal $\mathbf{g}$. Thus, the target $\mathbf{d}_i[k]$, when $s_i[k]= q_1$, is obtained as:
\begin{equation}\label{eq:heuristicPresented}
    \mathbf{d}_i[k] = \Big[ \mathbf{H}_{ij^\star}[k] \Big]_1, 
\end{equation}
with
\begin{equation} \label{eq:t_select2}
    j^\star = \argmax_{j \in \mathcal{T}_i[k]} \left( \gamma_{ij} + \sum_{l \in \mset{T}_i[k] \backslash j} \sigma_{ijl} \right),
\end{equation}
where the first term in~\eqref{eq:t_select2} is the path persistence ($\gamma_{ij}$) of the $j$-th candidate~\ac{UAV}, and the second term is the sum of path similarities between the path of $j$-th~\ac{UAV} and the rest of the potential~\ac{UAV} targets.

The selection heuristic presented in~\eqref{eq:heuristicPresented} is based on the following observation. Initially, when the swarm starts to move, only the informed~\acp{UAV} will have high path persistence. As the uninformed~\acp{UAV} select the informed ones as their target, the path similarity between the~\acp{UAV} will increase. Additionally, as the uninformed~\acp{UAV} keep moving towards the informed ones, their path persistence will also increase. As a result, the number of~\acp{UAV} with high path similarity and persistence increases over time. Thus, the uninformed~\acp{UAV} can select other uninformed~\acp{UAV} as targets and do not necessarily need the information about the path history of informed~\acp{UAV}. This is crucial when the number of~\acp{UAV} in the swarm is large and all uninformed~\acp{UAV} cannot observe the informed ones. Experimental results in Section~\ref{sec:exp_analysis} verify and demonstrate this selection process. To summarize, we have:
\begin{align}\label{eq:3.4}
\mathbf{d}_i[k] = 
\begin{cases}
  \mathbf{p}_i[k], & s_i[k]=q_0, \\
  \Big[ \mathbf{H}_{ij^\star}[k] \Big]_1, & s_i[k]=q_1, \\
  \mathbf{g}, & s_i[k]=q_2. \\
\end{cases}
\end{align}






\section{Target Tracking}
\label{sec:track_target}

After selecting the target, \ie~$\mathbf{d}_i[k]$, the $i$-th~\ac{UAV} constructs and follows a path to the target while avoiding collisions with the obstacles in the environment and with the surrounding~\acp{UAV}. This is achieved by controlling the~\ac{UAV} velocity, \ie~the control input $\mathbf{u}_i$ in~\eqref{eq:dyn}, as:
\begin{equation}\label{eq:controlSignal}
\mathbf{u}_i[k] = \mathbf{n}_i[k] + \mathbf{c}_i[k],
\end{equation}
where $\mathbf{n}_i[k]$ is the navigation control vector that will move the $i$-th~\ac{UAV} towards the desired target $\mathbf{d}_i[k]$, and $\mathbf{c}_i[k]$ is the collision control vector that provides the $i$-th~\ac{UAV} with collision avoidance capability. In the rest of this section, we will explain the process from the $i$-th~\ac{UAV} perspective, and to lighten the notation, we will drop the subscript $i$ from $\mathbf{n}_i[k]$ and $\mathbf{c}_i[k]$ variables. 




\subsection{Navigation control vector}
\label{sec:track_target:navigation}

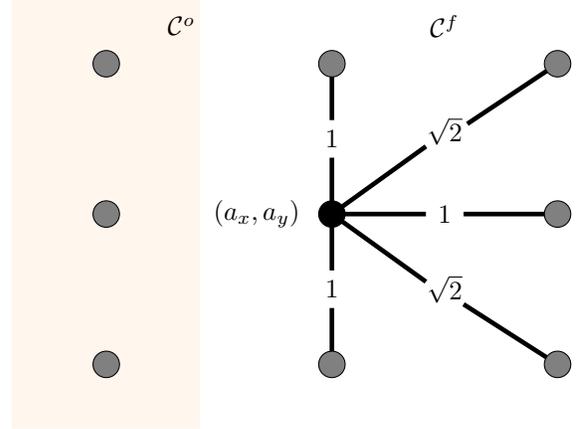
\begin{figure}[tb]
  \centering
  \begin{tikzpicture}
  
    \node (Region-Box) at (-3,2) [fill=orange!7, minimum height=5.7cm, minimum width=2.5cm]{}; 
  
    \draw (-2,4.5) node[text centered]{$\mathcal{C}^o$};
    \draw (1.5,4.5) node[text centered]{$\mathcal{C}^f$};
    
    \draw[-, line width=1.75pt] (0,2) -- (0,2.75); \node at (0,3.0) [text centered]{$1$}; \draw[-, line width=1.75pt] (0,3.25) -- (0,4);
    \draw[-, line width=1.75pt] (0,2) -- (0,1.25); \node at (0,1.0) [text centered]{$1$}; \draw[-, line width=1.75pt] (0,0.75) -- (0,0);
    \draw[-, line width=1.75pt] (0,2) -- (1.25,2); \node at (1.5,2) [text centered]{$1$}; \draw[-, line width=1.75pt] (1.75,2) -- (3,2);
    \draw[-, line width=1.75pt] (0,2) -- (1.25,1.1); \node at (1.5,1.0) [text centered]{$\sqrt{2}$}; \draw[-, line width=1.75pt] (1.75,0.8) -- (3,0);
    \draw[-, line width=1.75pt] (0,2) -- (1.25,2.9); \node at (1.5,3.1) [text centered]{$\sqrt{2}$}; \draw[-, line width=1.75pt] (1.8,3.2) -- (3,4);

    \node at (3,0) [circle,fill=gray,draw,inner sep=0pt,minimum size=10pt, text centered]{};
    \node at (0,0) [circle,fill=gray,draw,inner sep=0pt,minimum size=10pt, text centered]{};
    \node at (-3,0) [circle,fill=gray,draw,inner sep=0pt,minimum size=10pt, text centered]{};
    
    \node at (3,2) [circle,fill=gray,draw,inner sep=0pt,minimum size=10pt, text centered]{};
    \node at (0,2) [circle,fill,draw,inner sep=0pt,minimum size=10pt, text centered]{};
    \draw (-1,2) node[text centered]{$(a_x, a_y)$};
    \node at (-3,2) [circle,fill=gray,draw,inner sep=0pt,minimum size=10pt, text centered]{};
    
    \node at (3,4) [circle,fill=gray,draw,inner sep=0pt,minimum size=10pt, text centered]{};
    \node at (0,4) [circle,fill=gray,draw,inner sep=0pt,minimum size=10pt, text centered]{};
    \node at (-3,4) [circle,fill=gray,draw,inner sep=0pt,minimum size=10pt, text centered]{};
    
  \end{tikzpicture}
  \caption{Graph representation of the continuous space. The edges $(b_{x_i}, b_{y_i})$ are shown for the node at coordinate $(a_x, a_y)$ and depicted in gray. The free ($\mathcal{C}^f$) and occupied ($\mathcal{C}^o$) sets are represented using white and yellow colors, respectively. }
  \label{fig:map_grid}%
\end{figure}

The navigation control vector $\mathbf{n}[k]$ directs the~\ac{UAV} towards the target position $\mathbf{d}_i[k]$. This vector is obtained from the shortest collision-free path from the~\ac{UAV} position ($\mathbf{p}_i[k]$) to the target ($\mathbf{d}_i[k]$). To compute this path, we model the environment as a $2$-dimensional grid, resulting in a discretization of the continuous space in $\mathbb{R}^2$. Each grid coordinate either belongs to a set of occupied (obstacles) points $\mathcal{C}^o$ or free points $\mathcal{C}^f$. The occupied points are obtained by creating an online map of the environment. This grid represents a weighted undirected graph $\mathcal{G}$ with edges connecting coordinates to their neighbors, as shown in Fig.~\ref{fig:map_grid}. The graph $\mathcal{G}$ has an edge between the generic coordinates $(a_x, a_y)$ and $(b_x, b_y)$ if, and only if, both coordinates belong to the free set $\mathcal{C}^f$. The weight matrix $\mathbf{W}$ for the coordinates in the set $\mathcal{C}^f$ is defined as:
\begin{align}
[\mathbf{W}]_{ab} = 
\begin{cases}
  \sqrt{2} &,~b_x = a_x \pm 1, b_y = a_y \pm 1, \\
  1 &,~b_x = a_x \pm 1, b_y = a_y, \\
  1 &,~b_x = a_x, b_y = a_y \pm 1. \\
\end{cases}
\end{align}

The shortest path problem is formulated on the graph $\mathcal{G}$ with weights given by $\mathbf{W}$. To find the shortest path from the~\ac{UAV} position $\mathbf{p}=(p_x, p_y)$ to the target position $\mathbf{d}=(d_x, d_y)$ on $\mathcal{G}$, we use the A$^\star$ algorithm~\cite{orig_astar} with the following heuristic:
\begin{equation}
h(a_x, a_y) = \sqrt{(a_x - d_x)^2 + (a_y - d_y)^2}.
\end{equation}

The simultaneous mapping and planning needed for creating the graph is achieved using the methods described in~\cite{afzal_2021_icra}. The shortest path $\mathcal{P}$ (this path is collision-free since the graph $\mathcal{G}$ does not have any edges connecting the occupied coordinates in the set $\mathcal{C}^o$)  generated by the A$^\star$ algorithm is an ordered set of waypoints connecting the $i$-th~\ac{UAV} position $\mathbf{p}_i[k]$ to the target point $\mathbf{d}_i[k]$, whose closest waypoint to the current~\ac{UAV} position is denoted with the vector $\mathbf{a}_n \in \mathcal{P}$. Therefore, the navigation vector $\mathbf{n}[k]$ is given by:
\begin{equation}
\label{eq:n_vector}
    \mathbf{n}[k] = f_{ig}\mathbf{n}_I  + \bar{f}_{ig}\mathbf{n}_U,
\end{equation}
where $f_{ig} = 1$ if, and only if, the $i$-th~\ac{UAV} has goal information, while $\bar{f}_{ig}$ is its complementary function. So, $\mathbf{n}_I$ is the navigation vector if the~\ac{UAV} is informed; otherwise, the navigation vector is $\mathbf{n}_U$.

As discussed in Section~\ref{sec:dyn_target}, the target selection process relies on the high path persistence of the informed~\acp{UAV}. Thus, these~\acp{UAV} must always move with a finite velocity towards their target. However, in order to remain in the~\ac{LoS} of other~\acp{UAV}, the informed~\acp{UAV} must slow down for the swarm to catch up. Thus, the navigation vector for informed~\acp{UAV} is given as:
\begin{equation}\label{eq:nav_vec_informed}
    \mathbf{n}_I = \max \left( V^m, 1 - \frac{\sum\limits_{j \in \mathcal{N}_i} \lVert \mathbf{\check{p}}_{ij} - \mathbf{p}_i \rVert }{2R^f \lvert \mset{N}_i \rvert}  \right) K^n(\mathbf{a}_n - \mathbf{p}_i), 
\end{equation}
where $V^m \in (0,1)$ is the minimum normalized velocity of the informed~\ac{UAV}, and $K^n \in \mathbb{R}$ is a scaling coefficient to rescale the position vector to form the velocity control input ($\mathbf{u}$). The second term in the max function depends on the average distance from the~\acp{UAV} in $\mathcal{N}_i$. The magnitude of vector $\mathbf{n}_I$ decreases as this average distance increases, which prevents the informed~\ac{UAV} from wandering far away from the swarm. The $\max$ function ensures that the~\ac{UAV} always moves with a minimum velocity of $V^m$.

Unlike the informed~\acp{UAV}, the target of uninformed~\acp{UAV} is the position of other~\acp{UAV} in the swarm. Hence, the navigation vector must be designed to prevent collisions when moving towards other~\acp{UAV}. The vector $\mathbf{n}_U$ for uninformed~\acp{UAV} is obtained as described in Algorithm~\ref{alg:nav_vec}. The projection parallel to the relative position vector ($\mathbf{\check{p}}_{ij} - \mathbf{p}_i$) is iteratively scaled down if $\lVert \mathbf{\check{p}}_{ij} - \mathbf{p}_i \rVert$ is less than parameter $R^f$. This scaling ensures that the navigation vector will have a smaller component towards the $j$-th~\ac{UAV} as the $i$-th~\ac{UAV} moves closer to it, which is essential to prevent collisions. The parameter $\alpha \in \mathbb{R}_{>0}$ determines the rate of change of the scaling term. The orthogonal component $\mathbf{o}^\perp$ is not scaled; thus, the motion in direction orthogonal to ($\mathbf{\check{p}}_{ij} - \mathbf{p}_i$) remains unaffected.


\begin{algorithm}[tb]
\caption{Regulating navigation vector}
\label{alg:nav_vec}
\KwData{$\mathcal{N}_i, \mathbf{a}_n, \mathbf{p}_i$}
\KwResult{$\mathbf{n}_U$}

$\mathbf{n}_U \leftarrow K^n(\mathbf{a}_n - \mathbf{p}_i)$ \\
\For{$j \in \mathcal{N}_i$}{
  \Comment{projection parallel to $\mathbf{\check{p}}_{ij} - \mathbf{p}_i$}
  $\mathbf{s} \leftarrow \frac{\mathbf{n}_U \cdot ( \mathbf{\check{p}}_{ij} - \mathbf{p}_i )}{ \lVert \mathbf{\check{p}}_{ij} - \mathbf{p}_i \rVert^2}(\mathbf{\check{p}}_{ij} - \mathbf{p}_i)$ \\
  \Comment{projection orthogonal to $\mathbf{\check{p}}_{ij} - \mathbf{p}_i$}
  $\mathbf{o}^\perp \leftarrow \mathbf{n}_U - \mathbf{s}$ \\
  $\mathbf{n}_U \leftarrow \min \Bigg(1, \Big(\frac{ \lVert \mathbf{\check{p}}_{ij} - \mathbf{p}_i \rVert} {R_f} \Big)^\alpha \Bigg) \mathbf{s} + \mathbf{o}^\perp$
  }
\end{algorithm}



\subsection{Collision control vector}
\label{sec:track_target:collision}

The obstacles in the environment pose a great challenge when moving towards the target. The collision free path $\mathcal{P}$ used to obtain the navigation vector is safe in an ideal case, but it can fail to prevent collisions in the presence of motion and sensor uncertainties. 
As the environment is unknown, it is critical for the~\ac{UAV} to react to obstacles as soon as they are detected by the sensors. Thus, reactive collision avoidance is essential to ensure safety of the~\ac{UAV}. 
However, the~\acp{UAV} do not need to avoid all detected obstacles, but only those that pose an immediate threat of collision. We define with $\mathcal{O}_i[k]$  the set containing the position of all the obstacles (\eg~trees, other~\acp{UAV}) whose distance to $\mathbf{p}_i$ at time instant $k$ is smaller than the reaction distance $R^o \in \mathbb{R}_{>0}$. Increasing the value of $R^o$ would result in the~\ac{UAV} avoiding even far away obstacles. Thus $R^o$ is a design parameter that needs to be tuned according to the obstacle density and the number of~\acp{UAV} that make up the swarm.

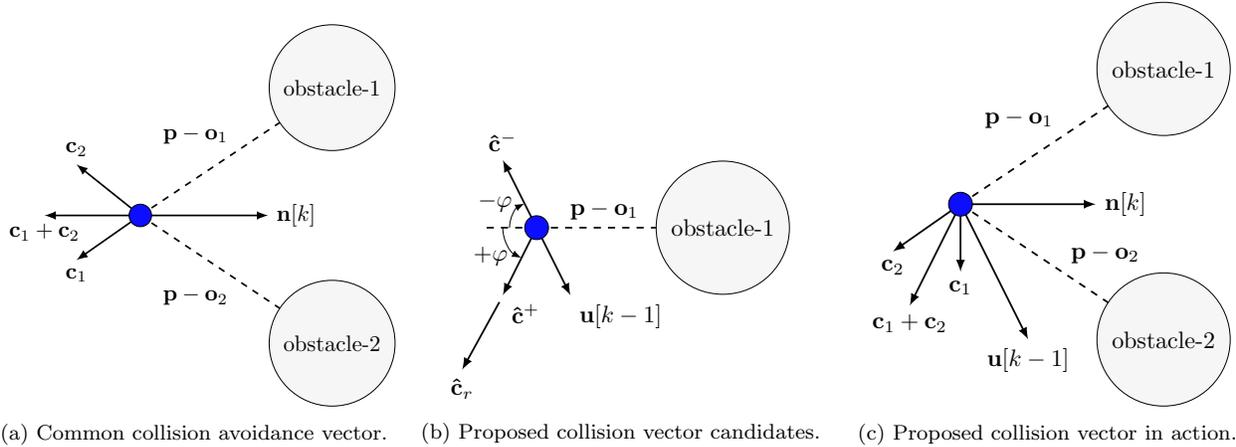
\begin{figure*}[tb]
\begin{center}
  \begin{subfigure}[t]{0.31\textwidth}
    \centering
    \scalebox{0.85}{
    \begin{tikzpicture}
    
      \draw[line width=0.75pt, dashed] (-3,-2) -- node[above left, text centered]{$\mathbf{p}-\mathbf{o}_1$} (0,0);
      \draw[line width=0.75pt, solid, -latex] (-3,-2) -- (-4,-2.7) node[below]{$\mathbf{c}_1$};;
      \draw[line width=0.75pt, solid, -latex] (-3,-2) -- (-4,-1.2) node[above]{$\mathbf{c}_2$};
      \draw[line width=0.75pt, dashed] (-3,-2) -- node[below left, text centered]{$\mathbf{p}-\mathbf{o}_2$} (0,-4);
      
      \draw[line width=0.75pt, solid, -latex] (-3,-2) -- (-1,-2) node[right]{$\mathbf{n}[k]$};

      \draw[line width=0.75pt, solid, -latex] (-3,-2) -- (-4.5,-2) node[below]{$\mathbf{c}_1 + \mathbf{c}_2$};;
      
      \node at (0,0) [circle,fill=gray!7,draw,inner sep=5pt,minimum size=10pt, text centered]{obstacle-1};
      \node at (0,-4) [circle,fill=gray!7,draw,inner sep=5pt,minimum size=10pt, text centered]{obstacle-2};
      
      \node at (-3,-2) [circle,fill=blue!95,draw,inner sep=0pt,minimum size=10pt, text centered]{};

    \end{tikzpicture}
    }
    \vspace{-1.2em}
    \caption{Common collision avoidance vector.}
    \label{fig:c_vect}%
  \end{subfigure} 
  \hfil
  \begin{subfigure}[t]{0.31\textwidth}
     \centering
    \scalebox{0.9}{
    \begin{tikzpicture}
    
      \node (x1) at (-3.75,-2)[text centered]{};  \node (O1) at (-3,-2)[text centered]{};  \node (z1) at (-3.5,-3)[text centered]{}; 
      \node (x2) at (-3.75,-2)[text centered]{};  \node (O2) at (-3,-2)[text centered]{};  \node (z2) at (-3.5,-1)[text centered]{}; 
      \pic [draw, -latex, angle radius=5mm, angle eccentricity=1.6, "$+\varphi$"] {angle = x1--O1--z1};
      \pic [draw, latex-, angle radius=4mm, angle eccentricity=1.8, "$-\varphi$"] {angle = z2--O2--x2};

      \draw[line width=0.75pt, dashed] (-3,-2) -- (-0.25,-2); \draw (-2.0,-2) node[above]{$\mathbf{p}-\mathbf{o}_1$};
      \draw[line width=0.75pt, dashed] (-3,-2) -- (-3.75,-2);
      \draw[line width=0.75pt, solid, -latex] (-3,-2) -- (-2.5,-3) node[below right]{$\mathbf{u}[k-1]$};
      \draw[line width=0.75pt, solid, -latex] (-3,-2) -- (-3.5,-1) node[above]{$\mathbf{\hat{c}}^-$};
      
      \draw[line width=0.75pt, solid, -latex] (-3,-2) -- (-3.5,-3) node[below right]{$\mathbf{\hat{c}}^+$};
      \draw[line width=0.75pt, solid, -latex] (-3.55,-3.1) -- (-4.1,-4.1) node[below]{$\mathbf{\hat{c}}_r$};
      
      \node at (-0.25,-2) [circle,fill=gray!7,draw,inner sep=5pt,minimum size=10pt, text centered]{obstacle-1};

      \node at (-3,-2) [circle,fill=blue!95,draw,inner sep=0pt,minimum size=10pt, text centered]{};
      
    \end{tikzpicture}
    }
    \vspace{-0.05em}
    \caption{Proposed collision vector candidates.}
    \label{fig:proposed_c_vect}%
  \end{subfigure} 
  \hfil
  \begin{subfigure}[t]{0.31\textwidth}
    \centering
    \scalebox{0.9}{
    \begin{tikzpicture}
    
      \draw[line width=0.75pt, dashed] (-3,-2) -- node[above left, text centered]{$\mathbf{p}-\mathbf{o}_1$} (0,0) ;
      \draw[line width=0.75pt, solid, -latex] (-3,-2) -- (-4,-2.7) node[below]{$\mathbf{c}_2$};;
      \draw[line width=0.75pt, solid, -latex] (-3,-2) -- (-3,-3.0) node[below]{$\mathbf{c}_1$};
      \draw[line width=0.75pt, dashed] (-3,-2) -- node[above right, text centered]{$\mathbf{p}-\mathbf{o}_2$} (0,-4);
      
      \draw[line width=0.75pt, solid, -latex] (-3,-2) -- (-1,-2) node[right]{$\mathbf{n}[k]$};
      \draw[line width=0.75pt, solid, -latex] (-3,-2) -- (-2,-4) node[below]{$\mathbf{u}[k-1]$};
      
      \draw[line width=0.75pt, solid, -latex] (-3,-2) -- (-3.75,-3.5) node[below]{$\mathbf{c}_1 + \mathbf{c}_2$};;
      
      \node at (0,0) [circle,fill=gray!7,draw,inner sep=5pt,minimum size=10pt, text centered]{obstacle-1};
      \node at (0,-4) [circle,fill=gray!7,draw,inner sep=5pt,minimum size=10pt, text centered]{obstacle-2};
      
      \node at (-3,-2) [circle,fill=blue!95,draw,inner sep=0pt,minimum size=10pt, text centered]{};
      
    \end{tikzpicture}
    }
    \caption{Proposed collision vector in action.}
    \label{fig:proposed_deadlock_avoid}%
  \end{subfigure}
  \caption{Collision avoidance vectors. The~\ac{UAV} is denoted in blue and the obstacles in gray. Fig.~(\ref{fig:c_vect})~Common collision avoidance vector with components parallel to the obstacle position vector. Fig.~(\ref{fig:proposed_c_vect}) Collision avoidance candidates described in~\eqref{eq:collision_avoidance_candidatesa} and~\eqref{eq:collision_avoidance_candidatesb}. Fig.~(\ref{fig:proposed_deadlock_avoid})~Proposed collision avoidance vector, as described in~\eqref{eq:collision_avoidance}, resulting from the interaction with two obstacles.}
  \label{fig:dist_time}%
  \vspace{-1.7em}
\end{center}
\end{figure*}

The reactive collision avoidance method used by several state-of-the-art real-world systems~\cite{pot_obst_avoid,afzal_2021_icra} often suffers from deadlocks. Some examples are shown in Fig.~\ref{fig:c_vect}. To prevent such deadlocks, we propose a novel method to move away from the obstacles while avoiding the deadlocks as much as possible. Unlike~\cite{pot_obst_avoid}, the proposed collision avoidance vector has components in both parallel and orthogonal directions relative to the position vector $(\mathbf{p}_i - \mathbf{o}_r)$, where $\mathbf{o}_r \in \mathbb{R}^2$ is the position of the obstacle considered. This collision vector is depicted in Fig.~\ref{fig:proposed_c_vect}. The parallel component takes the~\ac{UAV} away from the obstacle, while the orthogonal component moves it tangentially to the circle of radius $\lVert \mathbf{p}_i - \mathbf{o}_r \rVert$ and centered at $\mathbf{o}_r$. The combined parallel and orthogonal motions not only move the~\ac{UAV} away from the obstacle, but also around it. This allows for avoiding deadlocks, such as those shown in Fig.~\ref{fig:proposed_deadlock_avoid}.

The collision avoidance vector for an obstacle $\mathbf{o}_r \in \mathcal{O}_i[k]$ is obtained by first generating two candidate unit vectors $\mathbf{\hat{c}}^+$ and $\mathbf{\hat{c}}^-$ as:
\begin{subequations}
    \begin{align}
    \label{eq:collision_avoidance_candidatesa}
        \mathbf{\hat{c}}^+ &= \mathbf{R}(+\varphi) \frac{(\mathbf{p}_i - \mathbf{o}_r)}{ \lVert \mathbf{p}_i - \mathbf{o}_r \rVert}, \\
    \label{eq:collision_avoidance_candidatesb}    
        \mathbf{\hat{c}}^- &= \mathbf{R}(-\varphi) \frac{(\mathbf{p}_i - \mathbf{o}_r)}{ \lVert \mathbf{p}_i - \mathbf{o}_r \rVert},
    \end{align}
\end{subequations}
where $\mathbf{R}(\pm\varphi) \in \mathbb{R}^{2\times 2}$ denotes the rotation matrix along the $z$-axis of the world frame $\mathcal{F}_W$, which is used to rotate the relative vector $(\mathbf{p}_i - \mathbf{o}_r)$ by angle $\pm\varphi$ in the \emph{counterclockwise} direction. The angle $\varphi$ is given as:
\begin{equation}
    \varphi = \frac{\pi}{2R^o} \lVert \mathbf{p}_i - \mathbf{o}_r \rVert.
\end{equation}
The angle $\varphi$ varies with the distance to the obstacle. As a result, $\mathbf{\hat{c}}^+$ and $\mathbf{\hat{c}}^-$ have large components parallel to $(\mathbf{p}_i - \mathbf{o}_r)$ when the~\ac{UAV} is close to the obstacle. This large parallel component makes the reactive collision avoidance more focused in the parallel direction, thus preventing collisions. However, as the~\ac{UAV} moves away, $\varphi$ increases and $\mathbf{\hat{c}}^+$ and $\mathbf{\hat{c}}^-$ have larger orthogonal components. The orthogonal components focus on moving the~\ac{UAV} around the obstacle.

The vectors $\mathbf{\hat{c}}^+$ and $\mathbf{\hat{c}}^-$ denote two possible directions of motion to avoid collision with the obstacle $\mathbf{o}_r$ (see Fig.~\ref{fig:proposed_c_vect}). In order to keep the motion smooth, the vector with the least angular distance to the previous control input $\mathbf{u}_i[k-1]$ is used as the collision avoidance vector $\mathbf{c}_r$ for obstacle at $\mathbf{o}_r$. Thus, the vector $\mathbf{c}_r$ is obtained as:
\begin{equation}\label{eq:collision_avoidance_vec}
    \mathbf{c}_r = \max \left( 0, \frac{1}{ \lVert \mathbf{p}_i - \mathbf{o}_r \rVert}  - \frac{1}{R^o} \right) \mathbf{\hat{c}}_r,
\end{equation}
with
\begin{equation}
    \mathbf{\hat{c}}_r = \argmax_{\mathbf{\hat{b}}\in\{\mathbf{\hat{c}}^+, \mathbf{\hat{c}}^-\}} \left( \frac{\mathbf{\hat{b}} \cdot \mathbf{u}_i[k-1]}{ \lVert \mathbf{u}_i[k-1] \rVert}   \right), 
\end{equation}
where $\mathbf{\hat{c}}_r$ is a unit vector. The magnitude of $\mathbf{c}_r$ is inversely proportional to the relative distance to the obstacle. Thus, the~\ac{UAV} reacts more strongly to nearby obstacles in comparison to farther ones. Note that, $\mathbf{{c}}_r$ is a function of the obstacle position $\mathbf{o}_r\in\mathcal{O}_i[k]$. The collision control vector $\mathbf{c}[k]$ (see~\eqref{eq:controlSignal}) is a superposition of collision avoidance vectors of all the obstacles in $\mathcal{O}_i[k]$ and is obtained as:
\begin{equation}\label{eq:collision_avoidance}
    \mathbf{c}[k] = K^c  \sum_{\mathcal{O}_i[k]} \mathbf{c}_r,
\end{equation}
where $K^c$ is a scaling coefficient to rescale the summation. The preventive collision avoidance incorporated in $\mathbf{n}[k]$, along with the reactive collision avoidance from $\mathbf{c}[k]$, makes the control input safe even in the presence of imperfect sensor data.




\section{Simulations and Experiments}
\label{sec:exp_analysis}

In this section, we analyze the proposed collective navigation approach~\ac{PACNav}, as presented in Sections~\ref{sec:dyn_target} and~\ref{sec:track_target}. First, we evaluate the swarm behavior using realistic simulations in Gazebo~\cite{gazebo_2004_iros}, exploiting the advantages of software-in-the-loop simulations~\cite{SilanoSCM19}. Then, we present the results of real-world experiments carried out in a natural forest. Videos with the simulated and real-world experiments are available at~\url{http://mrs.felk.cvut.cz/pacnav}, while the source code has been made available as open-source\footnoteref{fotnote:code}.



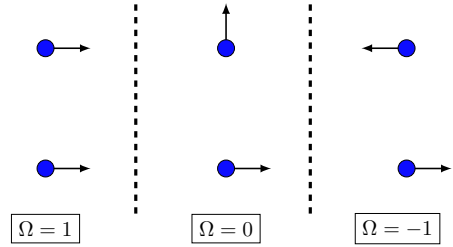
\begin{figure}[tb]
    \centering
    \scalebox{0.8}{
    \begin{tikzpicture}
      
      
      \draw[line width=0.75pt, solid, black, -latex] (0,1) -- (0.75,1);
      \draw[line width=0.75pt, solid, black, -latex] (0,-1) -- (0.75,-1);
      
      \node at (0,1) [circle,fill=blue!95,draw,inner sep=0pt,minimum size=8pt, text centered]{};
      
      \node at (0,-1) [circle,fill=blue!95,draw,inner sep=0pt,minimum size=8pt, text centered]{};
      
      
      
      \draw[black, dashed, line width=0.15em] (1.5,-1.75) coordinate -- (1.5,1.75) coordinate;
      
      \node at (0,-2) [draw, black, text centered]{$\Omega = 1$};
      
      
      \draw[line width=0.75pt, solid, black, -latex] (3,1) -- (3,1.75);
      \draw[line width=0.75pt, solid, black, -latex] (3,-1) -- (3.75,-1);
      
      \node at (3,1) [circle,fill=blue!95,draw,inner sep=0pt,minimum size=8pt, text centered]{};
      
      \node at (3,-1) [circle,fill=blue!95,draw,inner sep=0pt,minimum size=8pt, text centered]{};
      
      
      
      \draw[black, dashed, line width=0.15em] (4.4,-1.75) coordinate -- (4.4,1.75) coordinate;
      
      \node at (3.0,-2) [draw, black, text centered]{$\Omega = 0$};
      
      
      \draw[line width=0.75pt, solid, black, -latex] (6,1) -- (5.25,1);
      \draw[line width=0.75pt, solid, black, -latex] (6,-1) -- (6.75,-1);
      
      \node at (6,1) [circle,fill=blue!95,draw,inner sep=0pt,minimum size=8pt, text centered]{};
      
      \node at (6,-1) [circle,fill=blue!95,draw,inner sep=0pt,minimum size=8pt, text centered]{};
      
      \node at (5.85,-2) [draw, black, text centered]{$\Omega = -1$};
      
      
      
      
      
      

      
    \end{tikzpicture}
    }
    \caption{Illustrative scenarios of the \textit{order metric}~\eqref{eq:orderMetric} for two \acp{UAV}. \acp{UAV} are depicted in blue, while arrows denote their direction.}
    \label{fig:orderMetric}
\end{figure}

%
We introduce the \textit{order metric} $\Omega[k]$~\cite{order_def} that will be used to analyze the collective motion of the swarm as: 
%
%
\begin{equation}\label{eq:orderMetric}
    \Omega[k] = \frac{1}{N(N-1)} \sum_{i,j \in \mathcal{N}_i} \frac{\mathbf{v}_{i}[k]\cdot \mathbf{v}_{j}[k]}{\lVert \mathbf{v}_{i}[k] \rVert \lVert \mathbf{v}_{j}[k] \rVert},
\end{equation}
where $\mathbf{v}_{i}[k]$ is the instantaneous velocity of the $i$-th~\ac{UAV}. 
This metric captures the correlation between the movements of the agents and provides an indication about the overall alignment of the member~\acp{UAV}.
The value of $\Omega \in [-1,1]$, where $\Omega=1$, means that all members of the swarm are moving in the same direction. $\Omega < 1$ implies misalignment between the~\acp{UAV}. 
Figure~\ref{fig:orderMetric} illustrates the order between two agents.

\subsection{Simulated experiments}
\label{ssec:sim_experiments}

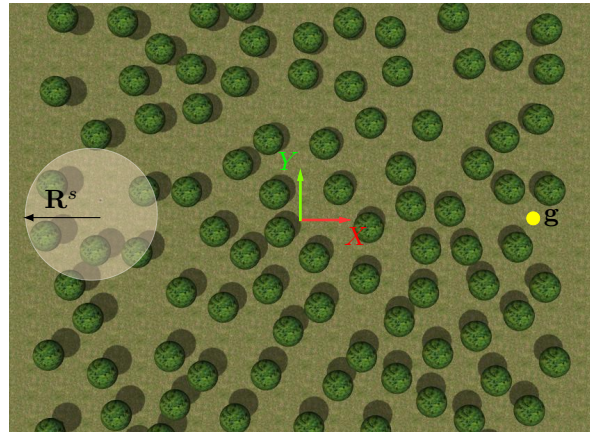
\begin{figure}[tb]
  \centering
  \begin{tikzpicture}
    \node at (0,0) [text centered]{\adjincludegraphics[width=0.45\textwidth, keepaspectratio, trim={{.0\width} {.125\height} {.0\width} {.125\height}}, clip]{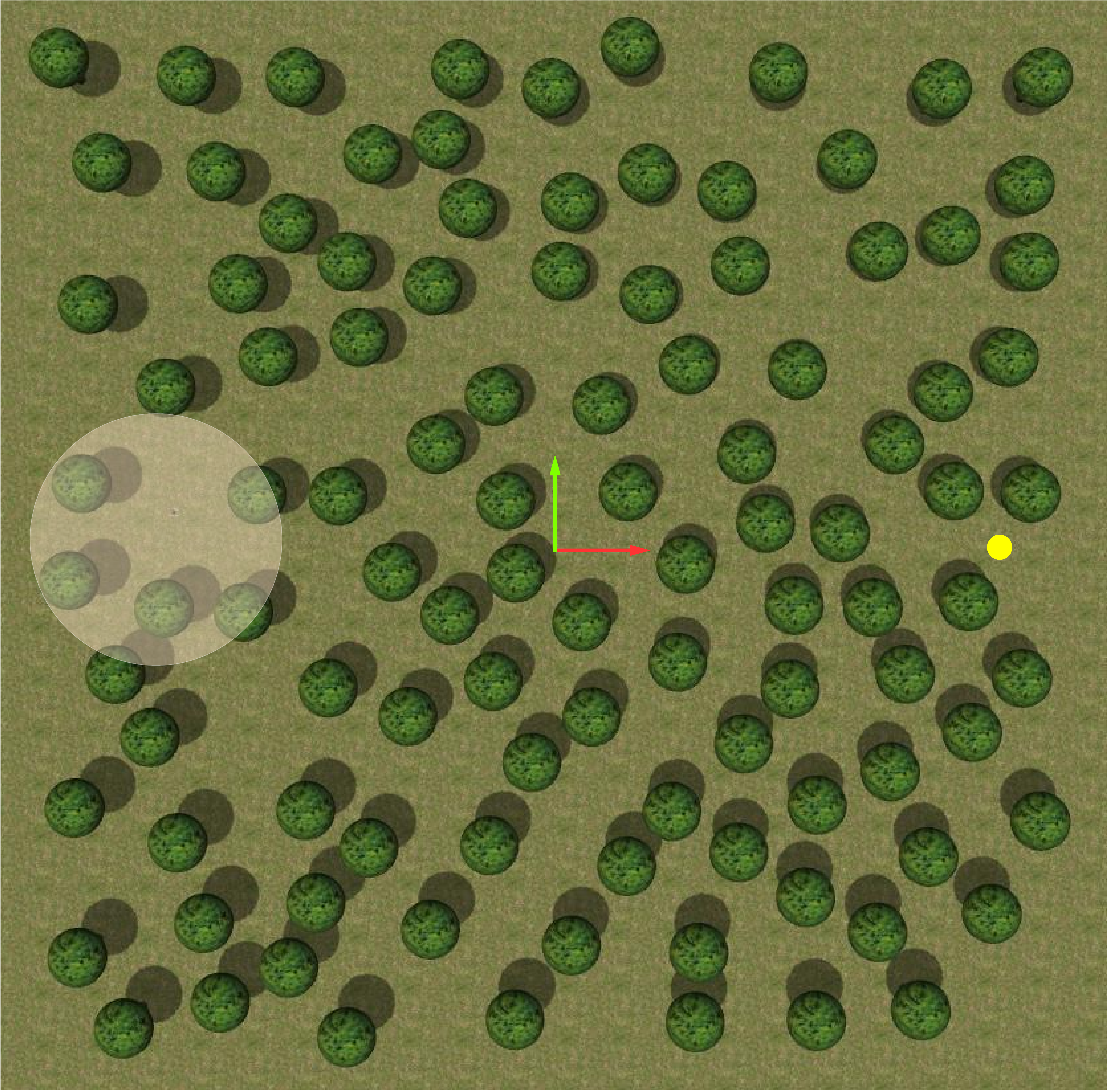}};  
    
    \draw (3.35,0) node[text centered, black]{$\mathbf{g}$};
    
    \draw (0.75,-0.25) node[text centered, red]{$X$};
    \draw (-0.15,0.75) node[text centered, green]{$Y$};
    
    \draw[latex-] (-3.68,0) -- node[above]{$\mathbf{R}^s$} (-2.65,0);
    
  \end{tikzpicture}
  \caption{Simulated forest in the Gazebo simulator. The~\acp{UAV} are initialized in the shaded region, while the goal is marked as the yellow dot. The red and green arrows denote the $X$- and $Y$-axis of the reference system $\mathcal{F}_W$, respectively.}
  \label{fig:sim_forest}%
\end{figure}

We simulate four different cases: (case 1A) a swarm composed of $N=3$~\acp{UAV} with only one informed~\ac{UAV}; (case 1B) a swarm composed of $N=3$~\acp{UAV} with two informed~\acp{UAV}; (case 2A) a swarm composed of $N=6$~\acp{UAV} with two informed~\acp{UAV}; and (case 2B) a swarm composed of $N=6$~\acp{UAV} with four informed~\acp{UAV}. For each case scenario, we run $10$ simulated experiments by varying the spatial distribution of the trees in the simulated natural forest, while initializing the~\acp{UAV} in the same shaded region and using the same fixed goal $\mathbf{g}$. A realization of the overall scenario is depicted in Fig.~\ref{fig:sim_forest}. Such an approach allowed us to statically characterize the behavior of the swarm with respect to the changes in the environment. 


The density of the forest with randomly distributed trees over an area $A$ can be described as:
\begin{equation}
 \rho = \frac{N^t \pi (R^o)^2}{A},
\end{equation}
where $N^t$ is the number of trees in the area $A$. We selected a forest of dimensions $\SI{50}{\meter} \times \SI{50}{\meter}$ with density of $\rho=0.4$. This density allows six~\acp{UAV} to simultaneously navigate through the forest. 

After trying various simulation setups, we observed that densities larger than $\rho=0.6$ make navigation in the forest extremely difficult. Due to the high number of trees, the~\acp{UAV} are trying to avoid collisions most of the time. Thus, the effect of the navigation vector $\mathbf{n}[k]$ in~\eqref{eq:controlSignal} is overtaken by the collision avoidance vector $\mathbf{c}[k]$. 

\begin{figure*}[tb]
  \begin{center}
    \hspace{-2.5em}
    \begin{subfigure}[t]{0.47\textwidth}
      \centering
      \includegraphics[width=\textwidth, keepaspectratio]{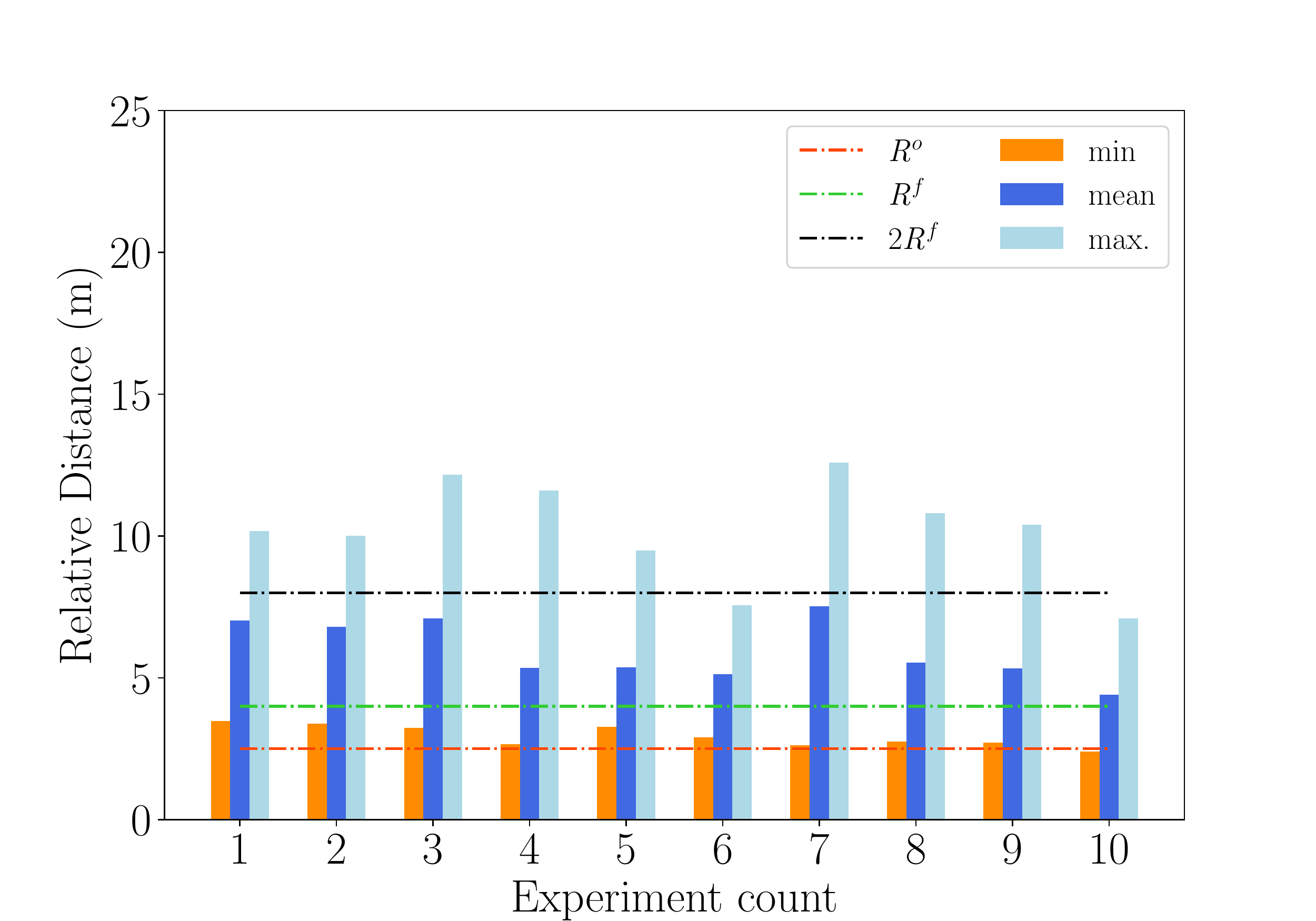}
      \caption{Relative distance between~\acp{UAV}.}
      \label{fig:3_uav_1_informed_a}
    \end{subfigure} 
    \hspace{-1.2em}
    \begin{subfigure}[t]{0.47\textwidth}
      \centering
      \includegraphics[width=1.2\textwidth, keepaspectratio]{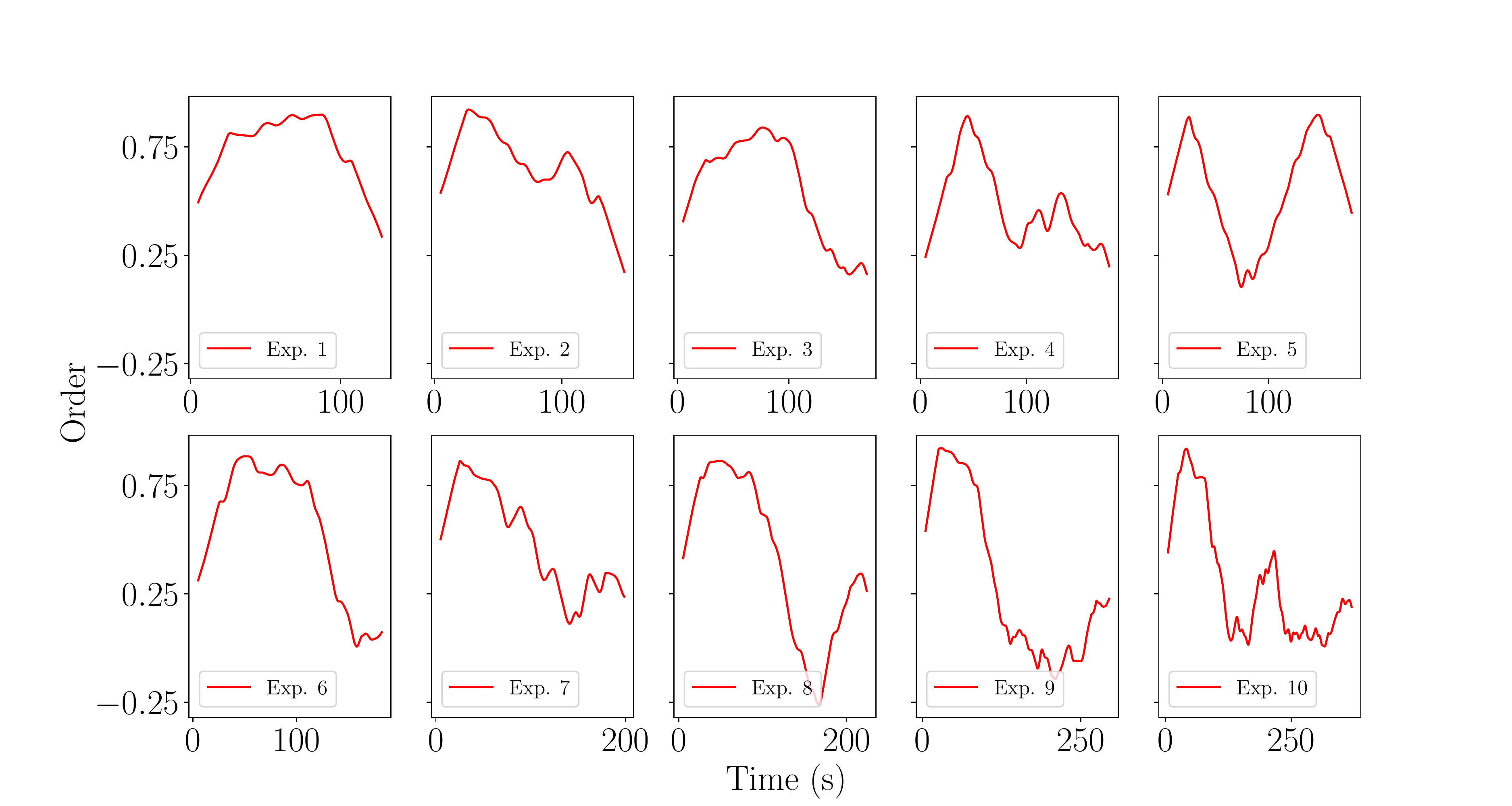}
      \caption{Order between~\acp{UAV}.}
      \label{fig:3_uav_1_informed_b}
    \end{subfigure} 
    \hfill
    \caption{Relative distance and Order between~\acp{UAV} in $10$ independent simulated experiments with three randomly initialized~\acp{UAV}, where one~\ac{UAV} had goal information (case 1A) with $R^s = \SI{3.0}{\meter}$ and $R^g=\SI{6.0}{\meter}$.}
    \label{fig:3_uav_1_informed}%
    \vspace{-1.7em}

  \end{center}
\end{figure*}

\begin{figure*}[tb]
  \begin{center}
    \hspace{-2.5em}
    \begin{subfigure}[t]{0.47\textwidth}
      \centering
      \includegraphics[width=\textwidth, keepaspectratio]{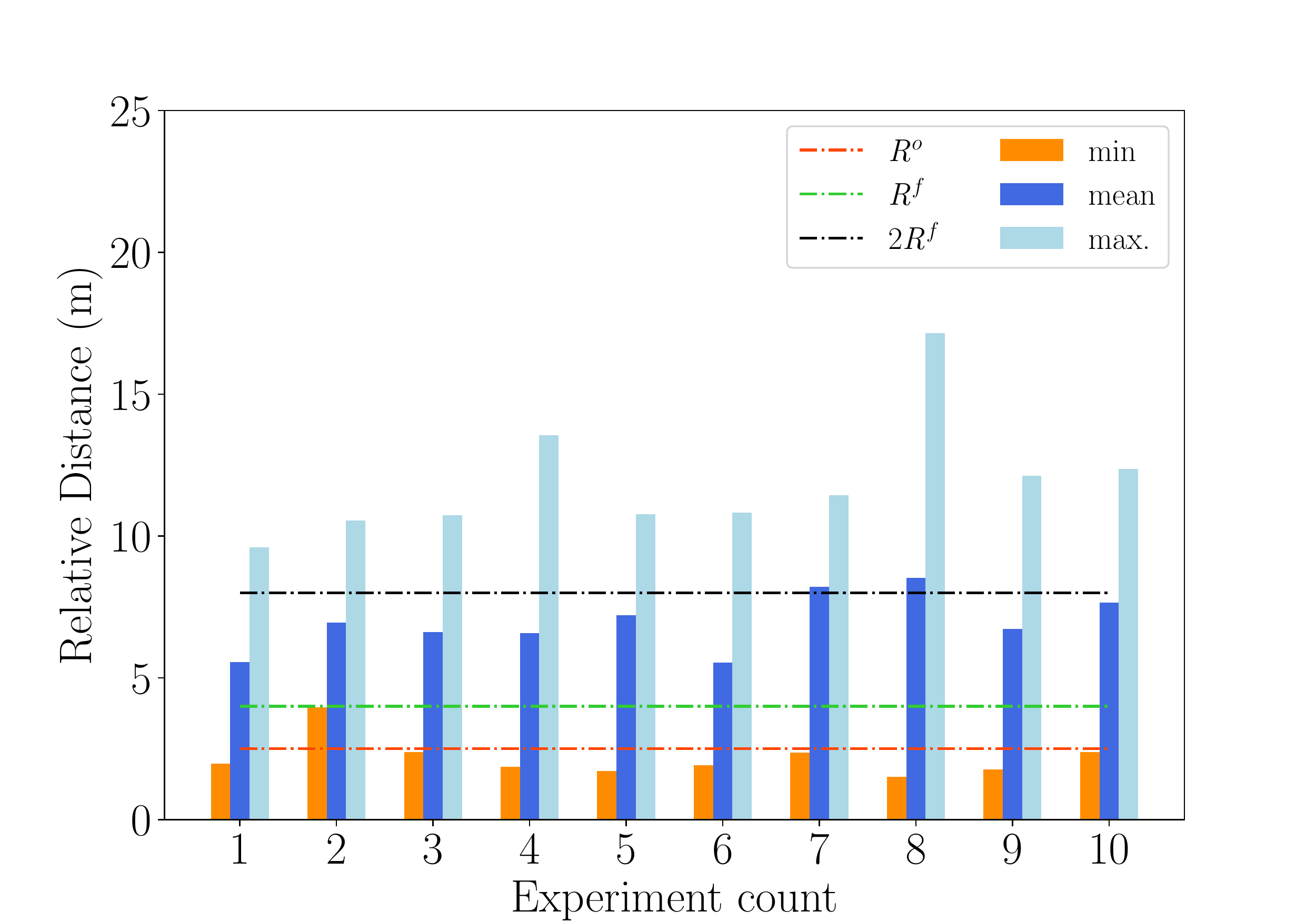}
      \caption{Relative distance between~\acp{UAV}.}
      \label{fig:3_uav_2_informed_a}%
    \end{subfigure} 
    \hspace{-1.2em}
    \begin{subfigure}[t]{0.47\textwidth}
      \centering
      \includegraphics[width=1.2\textwidth, keepaspectratio]{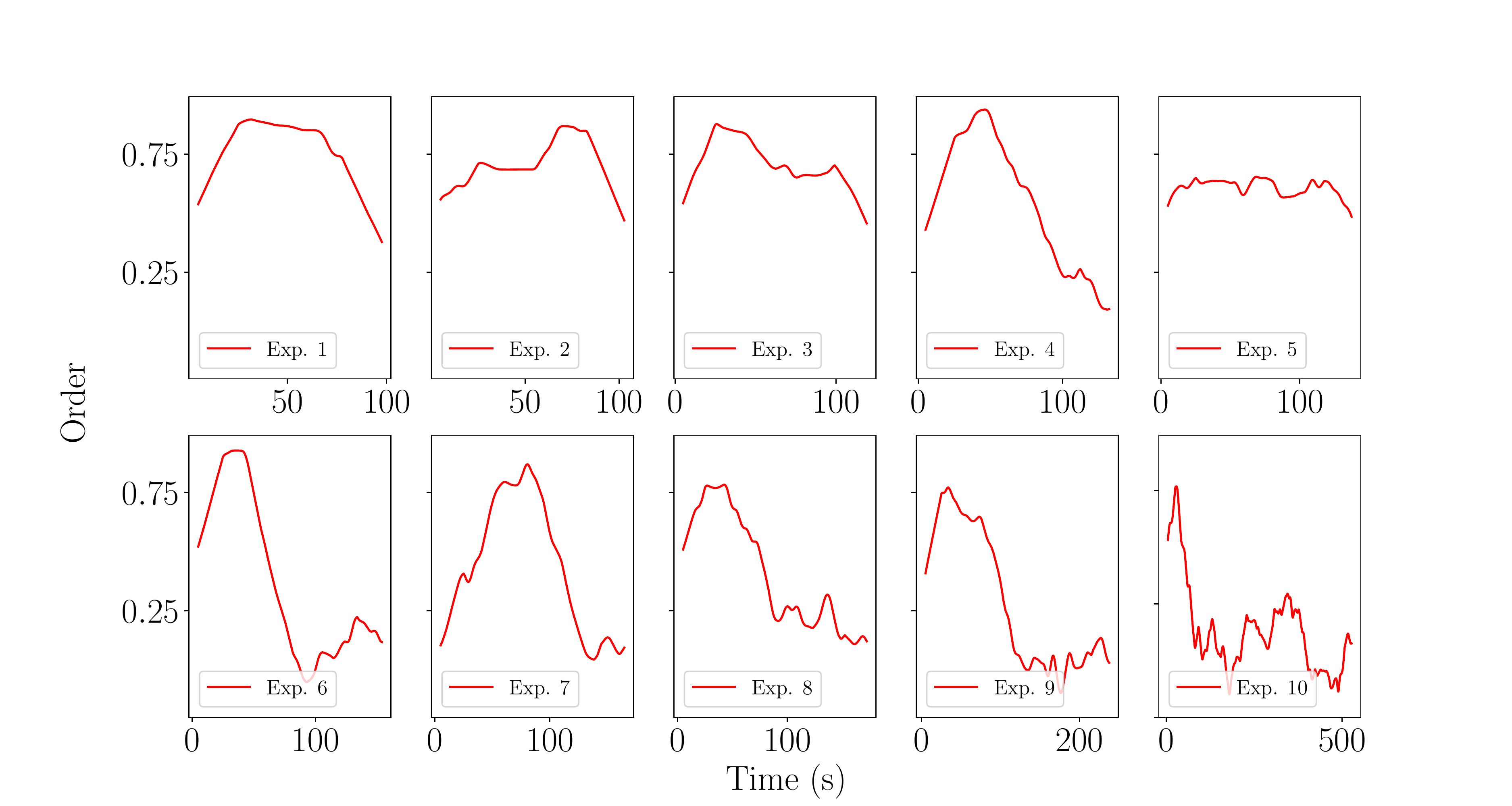}
      \caption{Order between~\acp{UAV}.}
      \label{fig:3_uav_2_informed_b}%
    \end{subfigure} 
    \hfill
    \caption{Relative distance and Order between~\acp{UAV} in $10$ independent simulated experiments with three randomly initialized~\acp{UAV}, where two~\acp{UAV} had goal information (case 1B) with $R^s = \SI{3.0}{\meter}$ and $R^g=\SI{6.0}{\meter}$.}
    \label{fig:3_uav_2_informed}%
    \vspace{-1.7em}
  \end{center}
\end{figure*}

\begin{figure*}[tb]
\begin{center}
  \hspace{-2.5em}
  \begin{subfigure}[t]{0.47\textwidth}
    \centering
    \includegraphics[width=\textwidth, keepaspectratio]{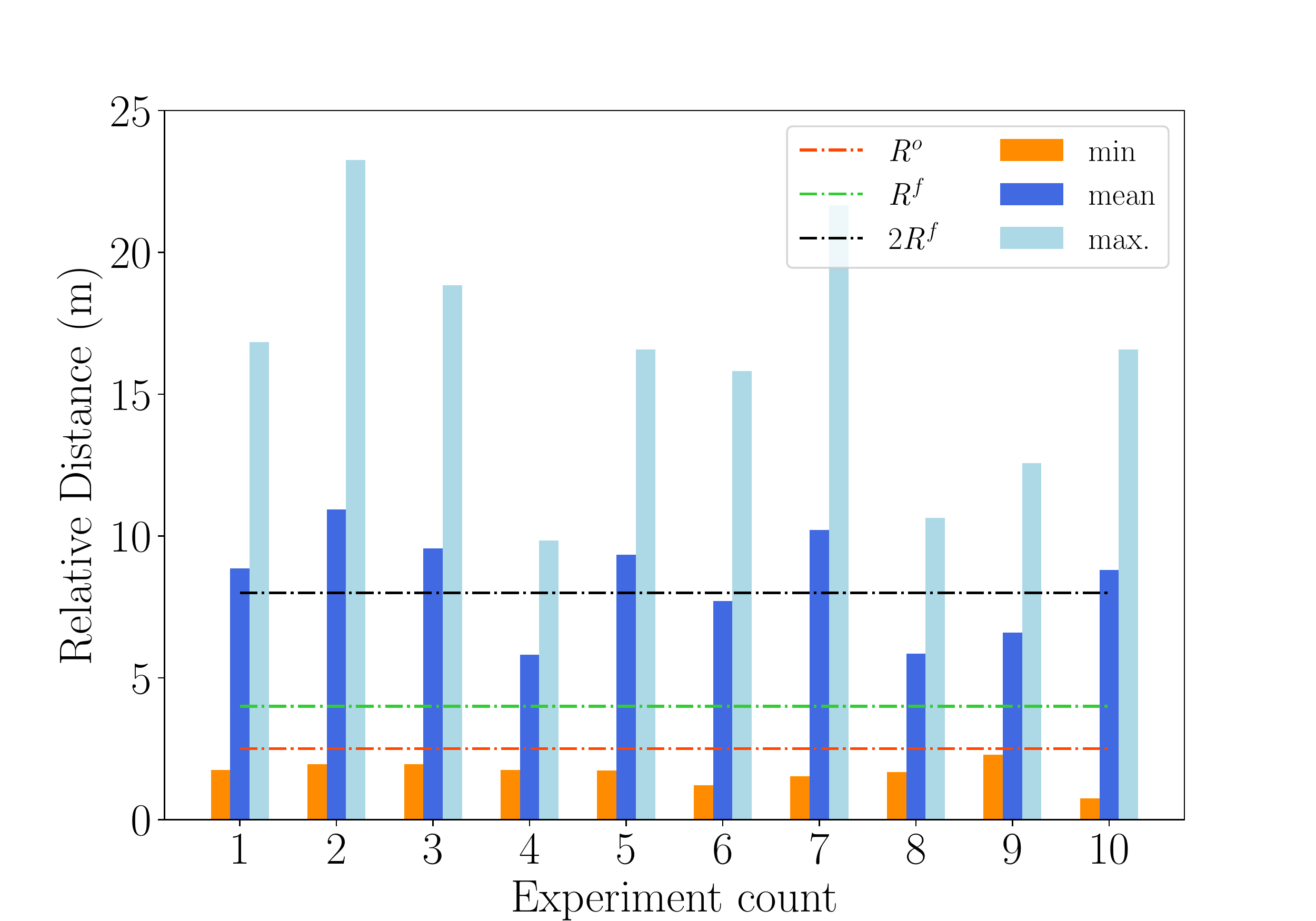}
    \caption{Relative distance between~\acp{UAV}.}
    \label{fig:6_uav_2_informed_a}%
  \end{subfigure} 
    \hspace{-1.2em}
  \begin{subfigure}[t]{0.47\textwidth}
    \centering
    \includegraphics[width=1.2\textwidth, keepaspectratio]{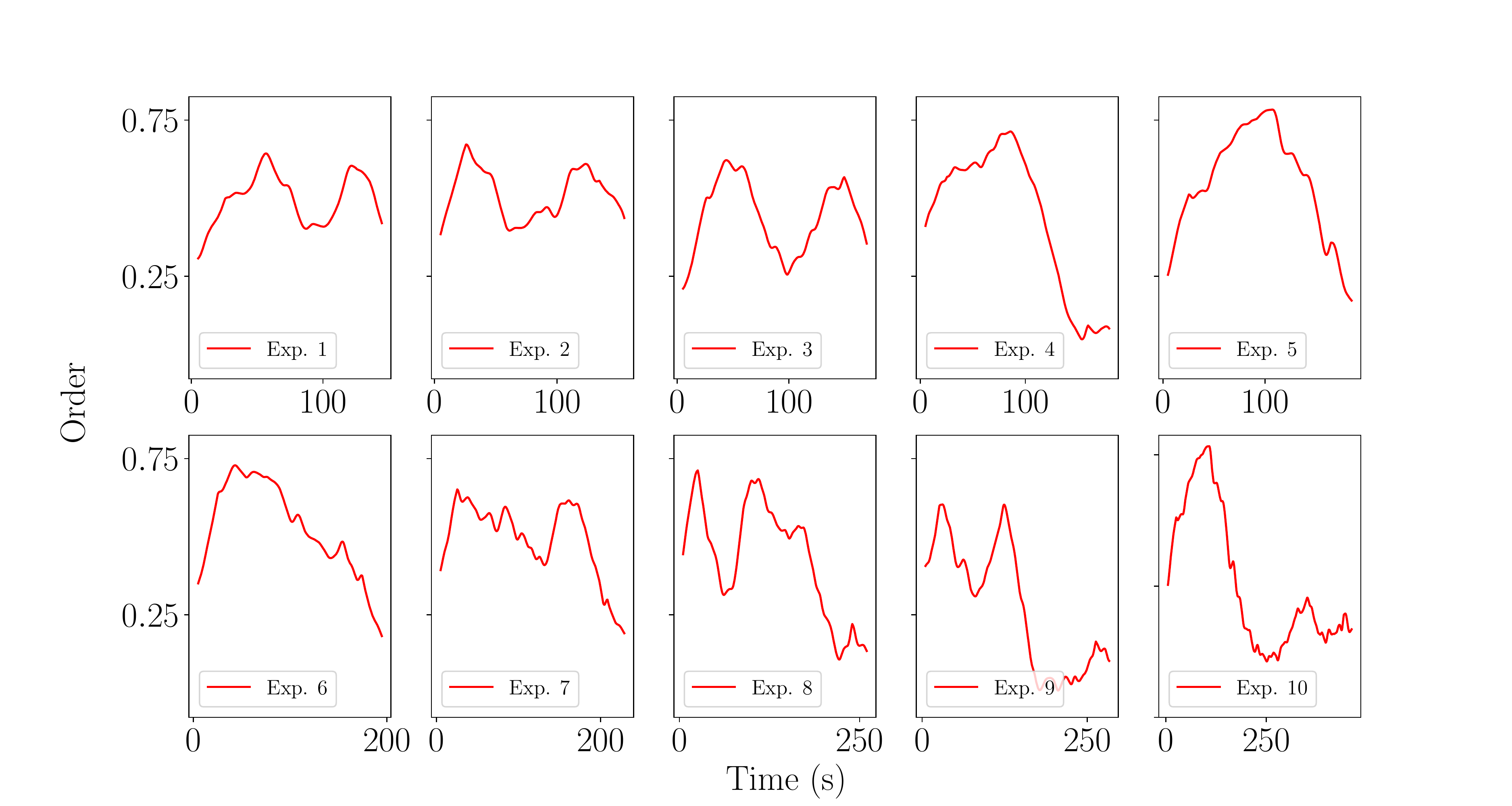}
    \caption{Order between~\acp{UAV}.}
    \label{fig:6_uav_2_informed_b}%
  \end{subfigure} 
  \caption{Relative distance and Order between~\acp{UAV} in $10$ independent simulated experiments with six randomly initialized~\acp{UAV}, where two~\acp{UAV} had goal information (case 2A) with $R^s = \SI{4.5}{\meter}$ and $R^g=\SI{8.5}{\meter}$.}
  \hfill
  \label{fig:6_uav_2_informed}%
  \vspace{-1.7em}
\end{center}
\end{figure*}

\begin{figure*}[tb]
\begin{center}
  \hspace{-2.5em}
  \begin{subfigure}[t]{0.47\textwidth}
    \centering
    \includegraphics[width=\textwidth, keepaspectratio]{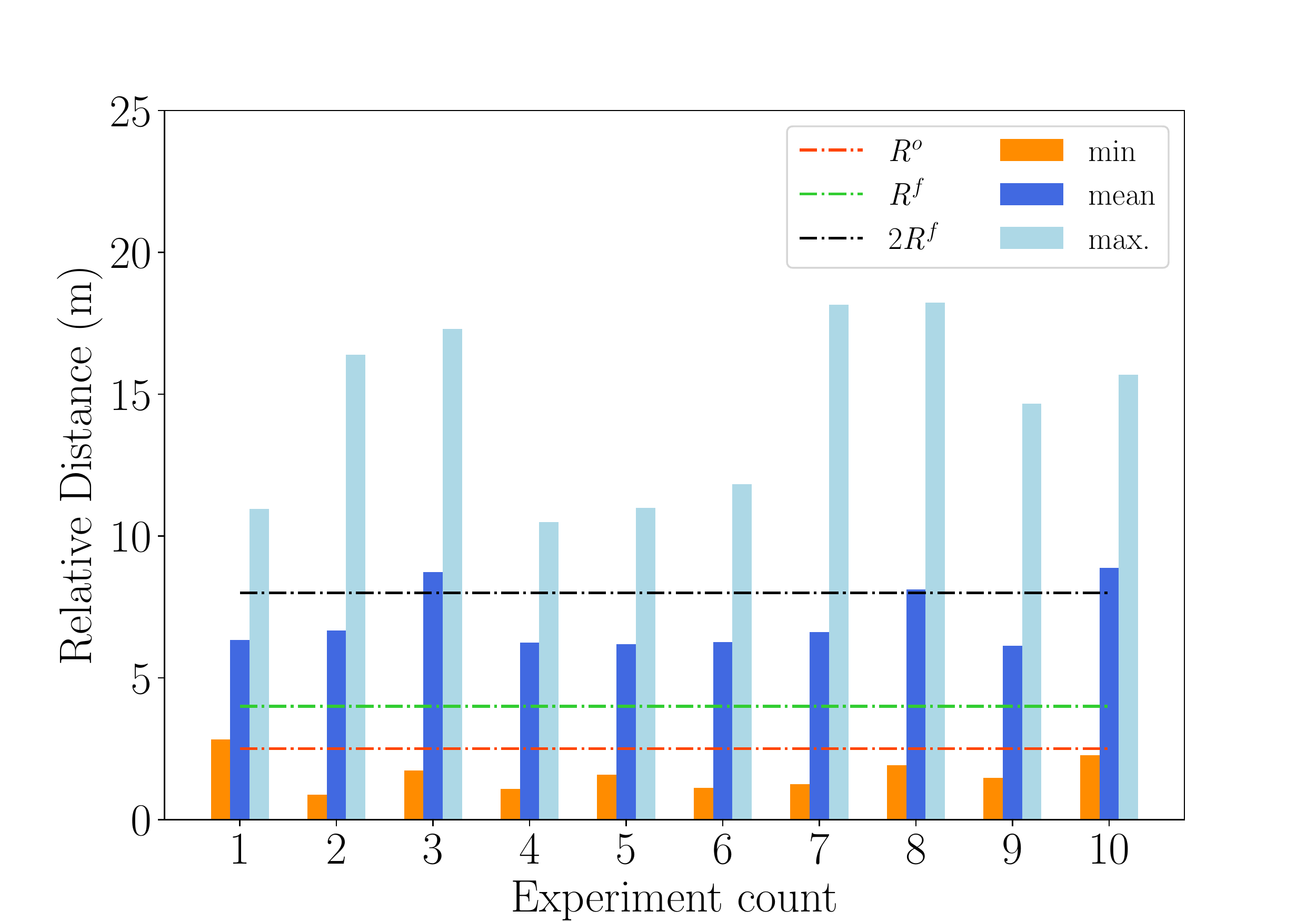}
    \caption{Relative distance between~\acp{UAV}.}
    \label{fig:6_uav_4_informed_a}%
  \end{subfigure} 
    \hspace{-1.2em}
  \begin{subfigure}[t]{0.47\textwidth}
    \centering
    \includegraphics[width=1.2\textwidth, keepaspectratio]{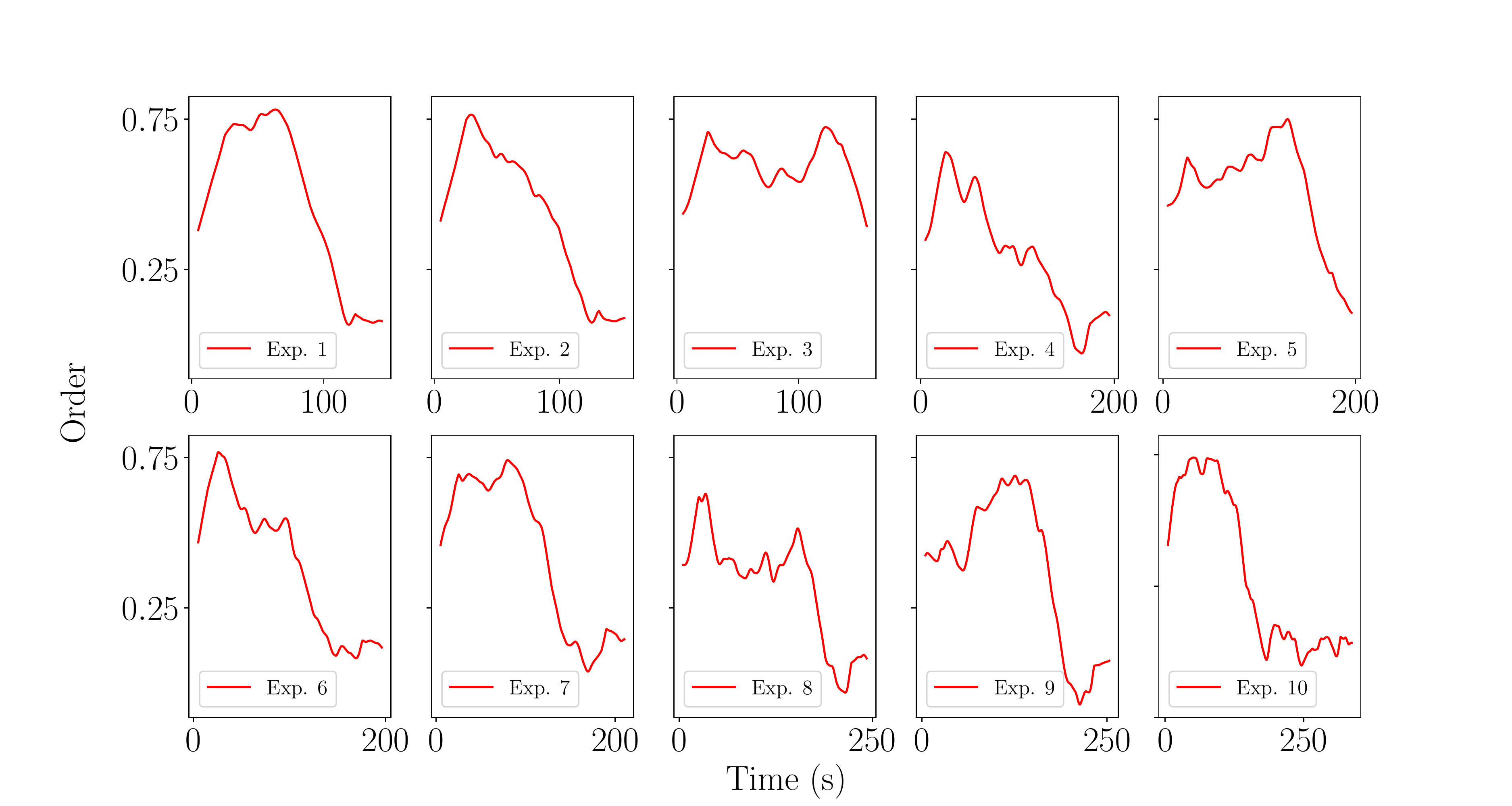}
    \caption{Order between~\acp{UAV}.}
    \label{fig:6_uav_4_informed_b}%
  \end{subfigure} 
  \caption{Relative distance and Order between~\acp{UAV} in $10$ independent simulated experiments with six randomly initialized~\acp{UAV}, where four~\acp{UAV} had goal information (case 2B) with $R^s = \SI{4.5}{\meter}$ and $R^g=\SI{8.5}{\meter}$.}
  \hfill
  \label{fig:6_uav_4_informed}%
  \vspace{-1.7em}
\end{center}
\end{figure*}

The task objective consists of navigating a swarm composed of $N$-\acp{UAV} from a random initial position (inside the shaded circle of radius $R^g$ in Fig.~\ref{fig:sim_forest}) to the goal $\mathbf{g}$, while trying to keep the mean distance between~\acp{UAV} below $2R^f$ most of the time.
The experiment is complete once all the $N$-\acp{UAV} are within a $R^g$ distance from the goal position. 
Figures~\ref{fig:3_uav_1_informed}-\ref{fig:6_uav_4_informed} present the results for the case 1A, case 1B, case 2A, and case 2B, respectively. The plots in each figure are arranged according to the increase completion time of the experiment. 


As described in Section~\ref{sec:dyn_target}, only the~\acp{UAV} farther than $R^f$ are considered as potential targets by uninformed~\acp{UAV}. Thus, to complete the mission, the uninformed~\acp{UAV} must remain at a distance larger than $R^f$ from the informed ones. This is one of the reasons why the mean distances between~\acp{UAV} in the swarm are larger than $R^f$ in Fig.~\ref{fig:3_uav_1_informed_a}, Fig.~\ref{fig:3_uav_2_informed_a}, Fig.~\ref{fig:6_uav_2_informed_a}, and Fig.~\ref{fig:6_uav_4_informed_a}. We also note that the minimum distance between~\acp{UAV} is sometimes slightly smaller than $R^o$, which is when the reactive collision avoidance mechanism of the~\acp{UAV} becomes active.

Comparing case 1A and case 1B, we observe that as the number of informed~\acp{UAV} increases, the mean completion time is reduced from $\SI{212.4}{\second}$ to $\SI{189.5}{\second}$. This is due to the fact that as the number of informed ~\acp{UAV} is increased, it becomes more likely that an uninformed \ac{UAV} directly tracks an informed~\ac{UAV}. Similarly, increasing the number of informed~\acp{UAV} from case 2A to case 2B also reduces the completion time from $\SI{231.4}{\second}$ to $\SI{213.3}{\second}$. We also observe that as the number of~\acp{UAV} are increased from three to six, the maximum and minimum distance between grows larger (see Fig.~\ref{fig:3_uav_1_informed_a}, Fig.~\ref{fig:3_uav_2_informed_a}, Fig.~\ref{fig:6_uav_2_informed_a}, and Fig.~\ref{fig:6_uav_4_informed_a}). This is because the~\acp{UAV} need to avoid collisions more often when there are more~\acp{UAV} moving around. This slows down their motion which leads to some of them lagging behind the others. However, the proposed approach successfully navigates the swarm in all the experiments with three and even six~\acp{UAV}.

Let us focus now on the order metric~\eqref{eq:orderMetric} plotted in Fig.~\ref{fig:3_uav_1_informed_b}, Fig.~\ref{fig:3_uav_2_informed_b}, Fig.~\ref{fig:6_uav_2_informed_b}, and Fig.~\ref{fig:6_uav_4_informed_b}. This metric measures the mean alignment between the velocity of~\acp{UAV}~\cite{order_def}. When all~\acp{UAV} move in the same direction, the order takes values close to $1$. However, when the~\acp{UAV} move in a \textit{disordered} manner and each~\ac{UAV} moves in a different direction, then the order metric takes lower values. From the plots, we observe that, in general, the order profile behaves in the following manner. In the first stage, the uninformed~\acp{UAV} are trying to figure out where to go and which~\ac{UAV} to follow. Here, the order value is low, but as they start to follow informed~\acp{UAV} (or other uniformed~\acp{UAV} which are already following informed~\acp{UAV}), the order starts to increase. When all the uninformed~\acp{UAV} are \textit{locked} and are tracking the informed~\acp{UAV}, the order metric remains almost constant for some time. At this stage all the~\acp{UAV} are moving, more or less, in the same direction at a similar speed. In the next stage, when the~\acp{UAV} start to reach the goal, the~\acp{UAV} lose alignment and the order metric decreases. This is due to the fact that, at the end of the experiment, the~\acp{UAV} slow down and move in different directions to avoid collisions. Finally, when most of the~\acp{UAV} have already reached the goal, but some~\acp{UAV} are lagging behind due to collision avoidance with the trees, we observe a fourth stage. In this stage, the order metric remains almost constant at low values, as seen in the last three subplots in Fig.~\ref{fig:3_uav_1_informed_b}, Fig.~\ref{fig:3_uav_2_informed_b}, Fig.~\ref{fig:6_uav_2_informed_b}, and Fig.~\ref{fig:6_uav_4_informed_b}. Table~\ref{tab:modelParameterValuesSimulatedExp} reports the values of the parameters used for the simulated experiments.

\begin{table}[tb]
    \centering
    \scalebox{0.9}{
    \begin{tabular}{|l|l|l|l|l|l|}
    \hline
    \textbf{Sym.} & \textbf{Value} & \textbf{Sym.} & \textbf{Value} & \textbf{Sym.} & \textbf{Value} \\

    \hline \hline
    $R^f$ & $\SI{4.0}{\meter}$ & $R^o$ & $\SI{2.5}{\meter}$ & $N^t$ & $\SI{104}{}$\\
    $K^c$ & $\SI{1.0}{\second^{-1}}$ & $K^n$ & $\SI{1.2}{\second^{-1}}$ & $\mathbf{g}$ & $(\SI{20}{\meter}, \SI{0}{\meter})^\top$\\
    %
    \hline
    \end{tabular}
    }
    \caption{List of parameters and their values for the simulated experiments.}
    \label{tab:modelParameterValuesSimulatedExp}
\end{table}





\subsection{Real-world experiments}
\label{ssec:real_experiments}

\begin{figure}[tb]
  \centering
  \includegraphics[width=0.40\textwidth]{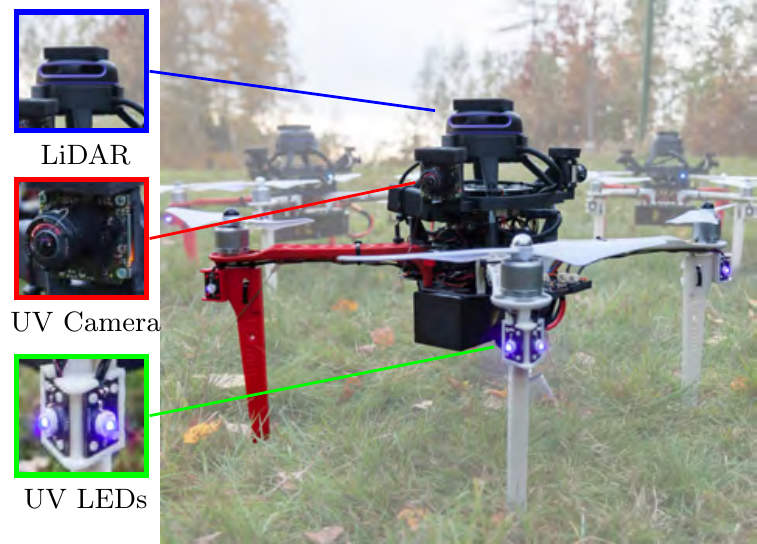}
    \caption{Quadrotor platform highlighting the~\ac{UV} LEDs and the UV cameras used by the~\ac{UVDAR} system for direct relative localization of~\acp{UAV}, and the 2D~\ac{LiDAR} for~\ac{SLAM}.}
    \label{fig:uav_platform}%
\end{figure}

\begin{figure}[tb]
  \centering
  \includegraphics[width=0.5\textwidth, keepaspectratio]{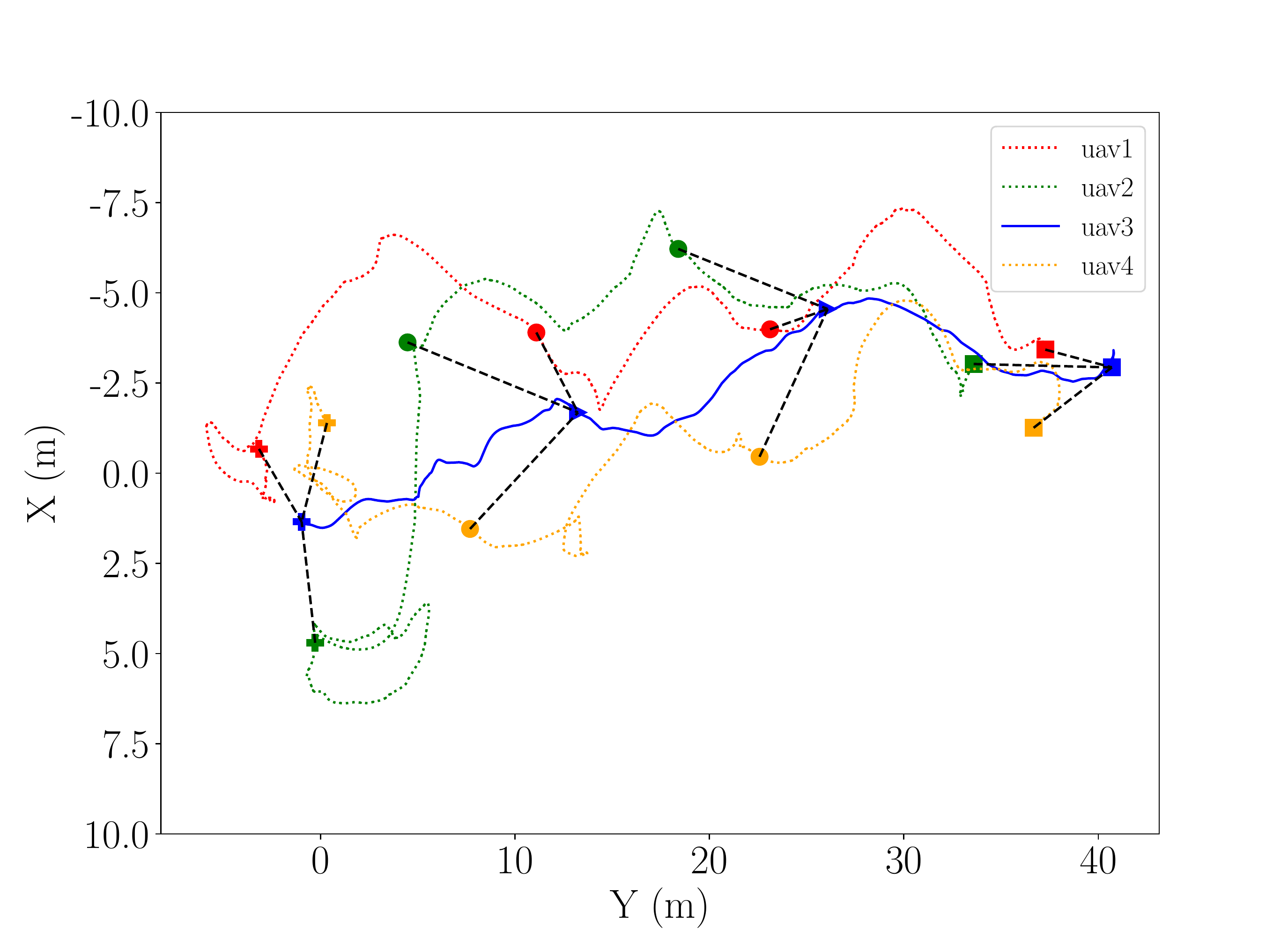}
  \caption{Recorded trajectory from the real-world flight experiment in the forest. The solid line represent the trajectory of the \texttt{uav3} (informed about the goal), while dashed lines represent the trajectories of all other~\acp{UAV} (\texttt{uav1}, \texttt{uav2}, and \texttt{uav4}). Triangle and circle marks depict the position of the~\acp{UAV} at time instants $\SI{105}{\second}$ and $\SI{180}{\second}$, while plus and square marks show the initial ($\SI{0}{\second}$) and final ($\SI{300}{\second}$) positions of the~\acp{UAV}, respectively.}
  \label{fig:forest_traj}%
\end{figure}

\begin{figure*}[tb]
    \hspace{0.1cm}
    \begin{subfigure}{0.45\columnwidth}
      \centering
      \begin{tikzpicture}
        \node[anchor=south west,inner sep=0] (img) at (0,0) { 
        \adjincludegraphics[trim={{.0\width} {.3\height} {.0\width} {.1\height}}, clip, scale=0.195]{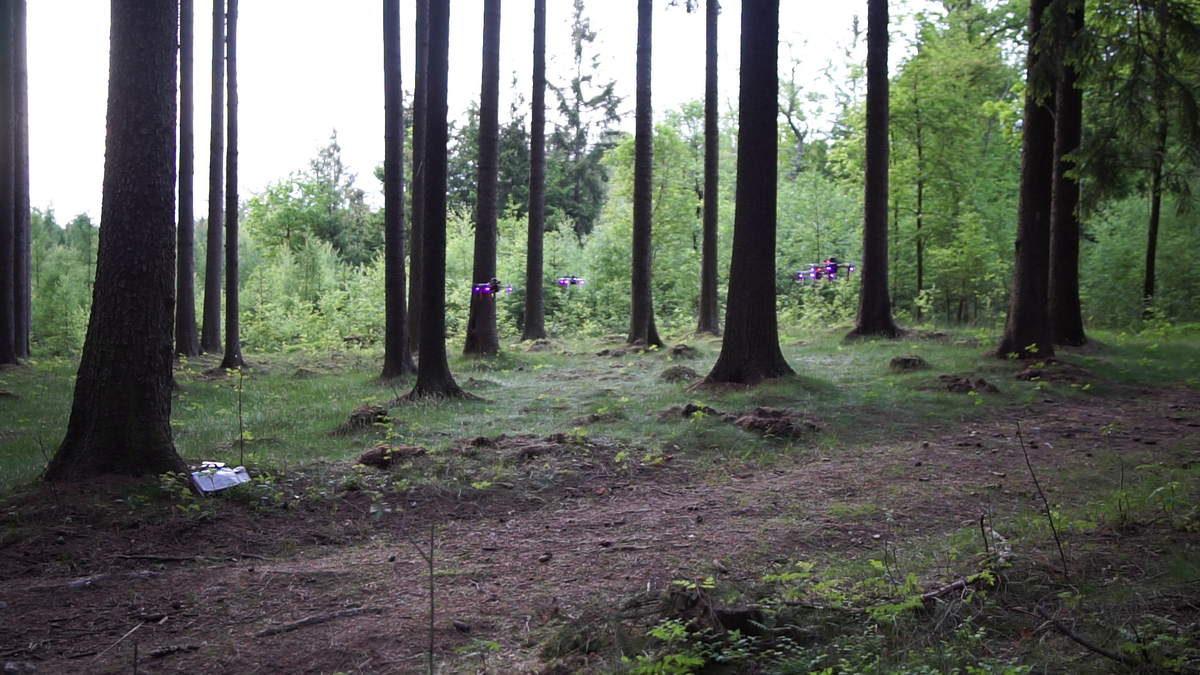}};
        \begin{scope}[x={(img.south east)},y={(img.north west)}]
          \draw [blue, solid, thick] (0.405, 0.465) circle (0.04); 
          \draw [green, solid, thick] (0.475, 0.48) circle (0.04); 
          \draw [yellow, solid, thick] (0.68, 0.50) circle (0.04);
          \draw [red, solid, thick] (0.69, 0.50) circle (0.04);
          \node[imglabel,text=black] (label) at (img.south west) {\scriptsize $\SI{25}{\second}$};
        \end{scope}
      \end{tikzpicture}
    \end{subfigure}
    \hspace{4.5cm}
    \begin{subfigure}{0.45\columnwidth}
      \centering
      \begin{tikzpicture}
        \node[anchor=south west,inner sep=0] (img) at (0,0) { 
        \adjincludegraphics[trim={{.0\width} {.3\height} {.0\width} {.1\height}},clip,scale=0.195]{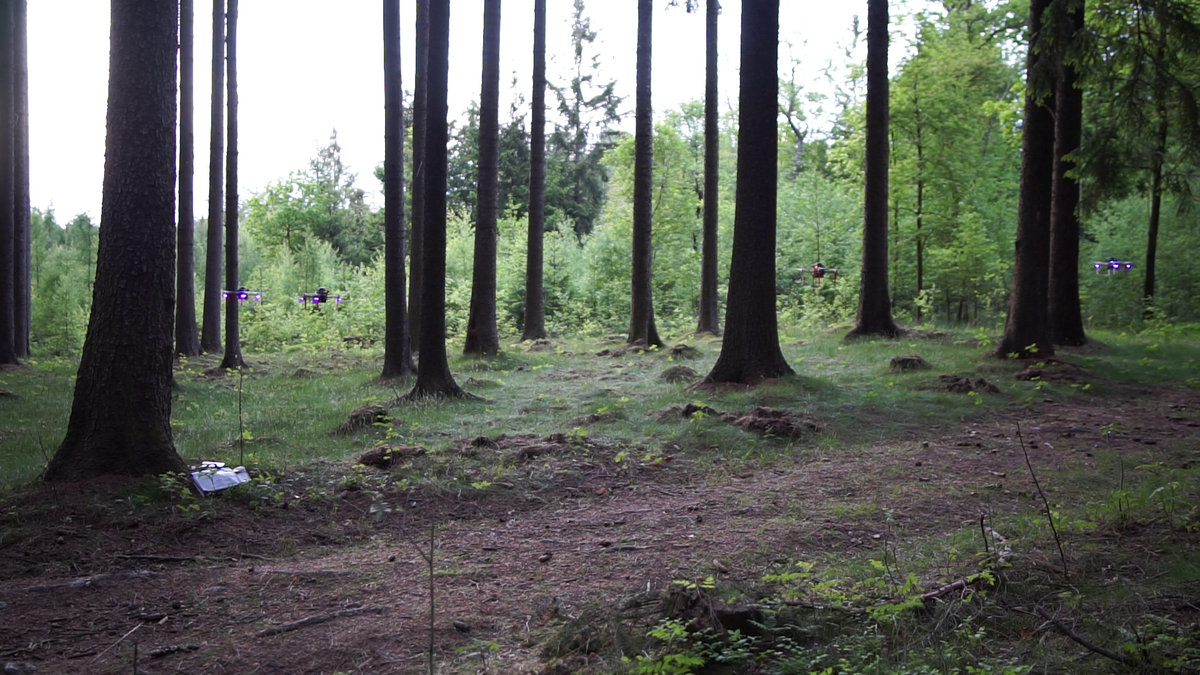}};
        \begin{scope}[x={(img.south east)},y={(img.north west)}]
          \draw [blue, solid, thick] (0.20, 0.44) circle (0.04);
          \draw [green, solid, thick] (0.27, 0.44) circle (0.04);
          \draw [yellow, solid, thick] (0.685, 0.50) circle (0.04);
          \draw [red, solid, thick] (0.925, 0.515) circle (0.04); 
          \node[imglabel,text=black] (label) at (img.south west) {\scriptsize $\SI{45}{\second}$};
        \end{scope}
      \end{tikzpicture}
    \end{subfigure}
    \vspace{0.25cm}
    \\
    \begin{subfigure}{0.45\columnwidth}
      \centering
      \begin{tikzpicture}
        \node[anchor=south west,inner sep=0] (img) at (0,0) {\hspace{0.1cm} 
        \adjincludegraphics[trim={{.0\width} {.3\height} {.0\width} {.1\height}}, clip, scale=0.195]{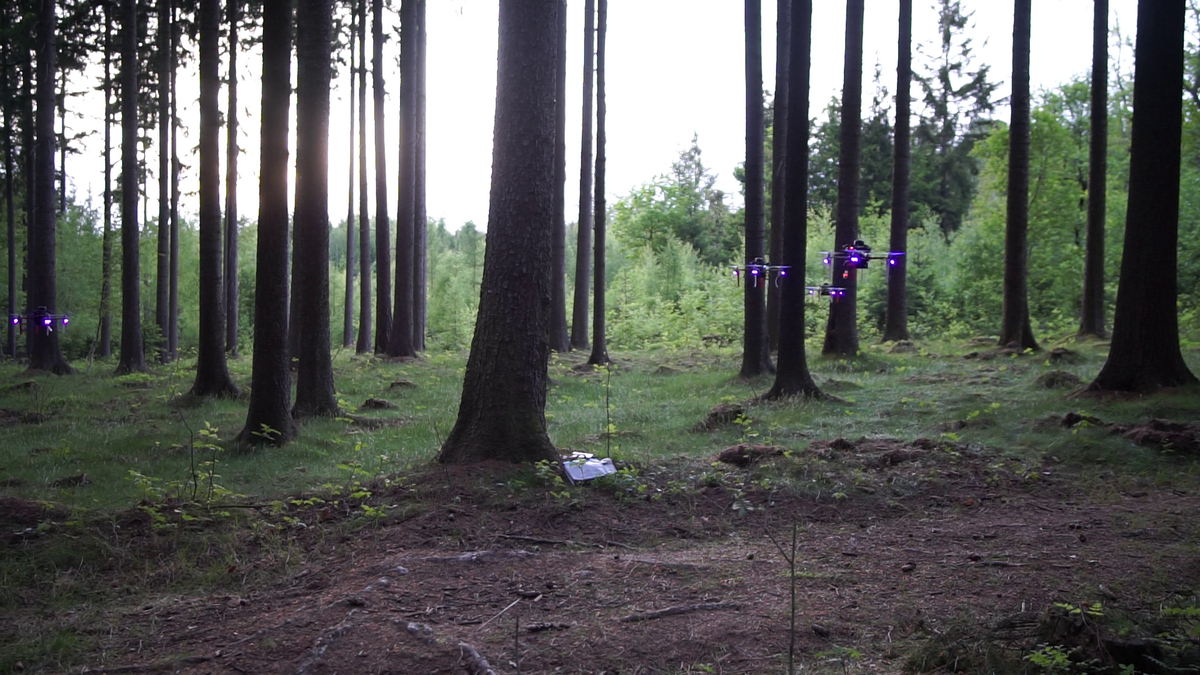}};
        \begin{scope}[x={(img.south east)},y={(img.north west)}]
          \draw [blue, solid, thick] (0.06, 0.375) circle (0.04);
          \draw [red, solid, thick] (0.64, 0.50) circle (0.04);
          \draw [yellow, solid, thick] (0.695, 0.45) circle (0.04);
          \draw [green, solid, thick] (0.72, 0.53) circle (0.045);
          \node[imglabel,text=black] (label) at (img.south west) {\hspace{0.2cm}\scriptsize $\SI{110}{\second}$};
        \end{scope}
      \end{tikzpicture}
    \end{subfigure}
    \hspace{4.5cm}
    \begin{subfigure}{0.45\columnwidth}
      \centering
      \begin{tikzpicture}
        \node[anchor=south west,inner sep=0] (img) at (0,0) {\hspace{0.1cm}
        \adjincludegraphics[trim={{.0\width} {.3\height} {.0\width} {.1\height}},clip,scale=0.195]{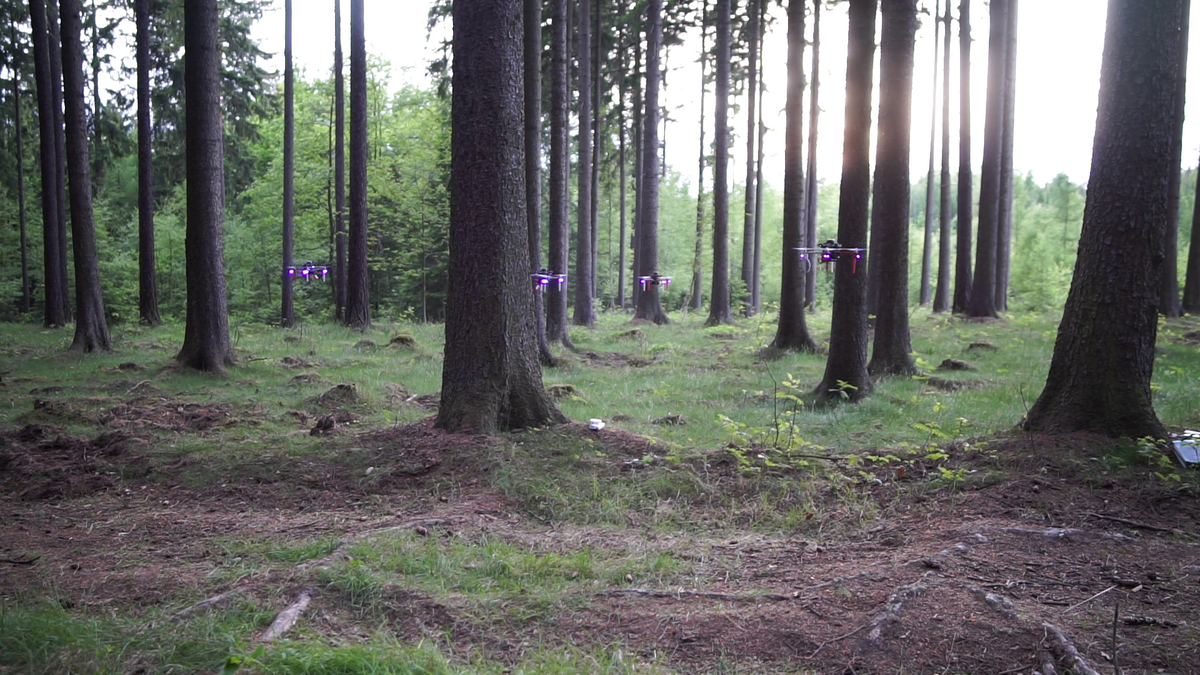}};
        \begin{scope}[x={(img.south east)},y={(img.north west)}]
          \draw [blue, solid, thick] (0.27, 0.50) circle (0.04); 
          \draw [red, solid, thick] (0.465, 0.4725) circle (0.0405); 
          \draw [yellow, solid, thick] (0.56, 0.48) circle (0.04);
          \draw [green, solid, thick] (0.70, 0.54) circle (0.045); 
          \node[imglabel,text=black] (label) at (img.south west) {\hspace{0.2cm}\scriptsize $\SI{150}{\second}$};
        \end{scope}
      \end{tikzpicture}
    \end{subfigure}
    \vspace{0.25cm}
    \\
    \begin{subfigure}{0.45\columnwidth}
      \centering
      \begin{tikzpicture}
        \node[anchor=south west,inner sep=0] (img) at (0,0) {\hspace{0.1cm}
        \adjincludegraphics[trim={{.0\width} {.3\height} {.0\width} {.1\height}},clip,scale=0.195]{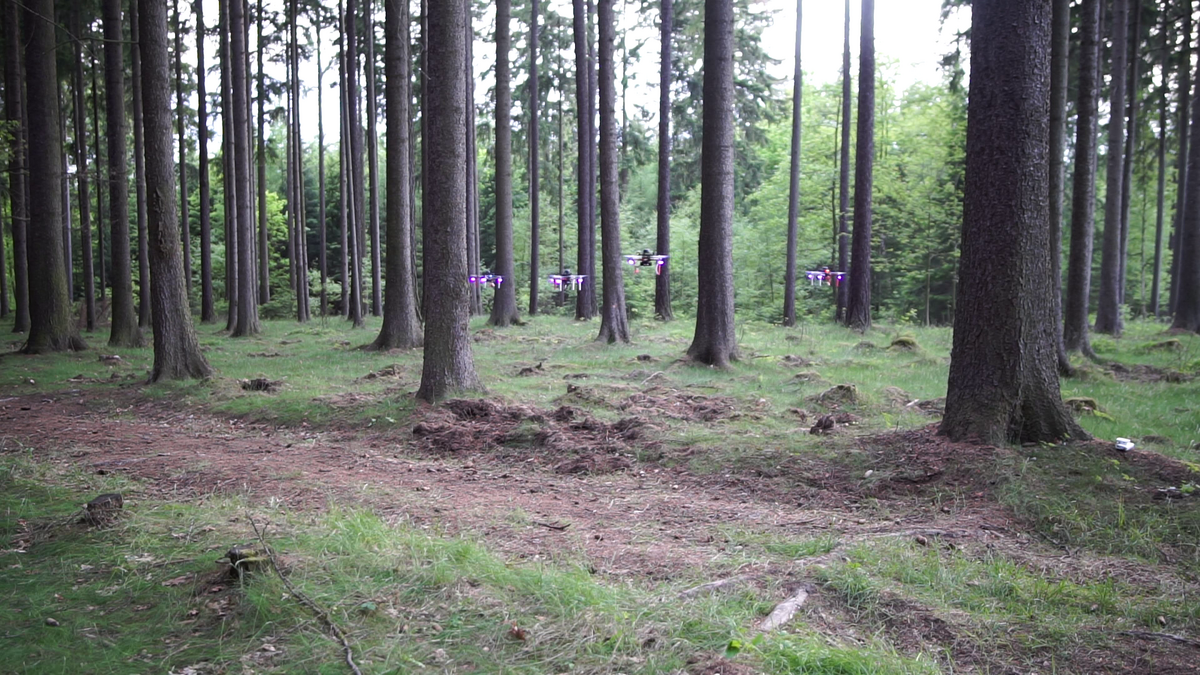}};
        \begin{scope}[x={(img.south east)},y={(img.north west)}]
          \draw [blue, solid, thick] (0.42, 0.48) circle (0.04);
          \draw [red, solid, thick] (0.48, 0.48) circle (0.04); 
          \draw [green, solid, thick] (0.55, 0.525) circle (0.04);
          \draw [yellow, solid, thick] (0.695, 0.485) circle (0.041);
          \node[imglabel,text=black] (label) at (img.south west) {\hspace{0.2cm}\scriptsize $\SI{184}{\second}$};
        \end{scope}
      \end{tikzpicture}
    \end{subfigure}
    \hspace{4.5cm}
    \begin{subfigure}{0.45\columnwidth}
      \centering
      \begin{tikzpicture}
        \node[anchor=south west,inner sep=0] (img) at (0,0) {\hspace{0.1cm}
        \adjincludegraphics[trim={{.0\width} {.3\height} {.0\width} {.1\height}},clip,scale=0.195]{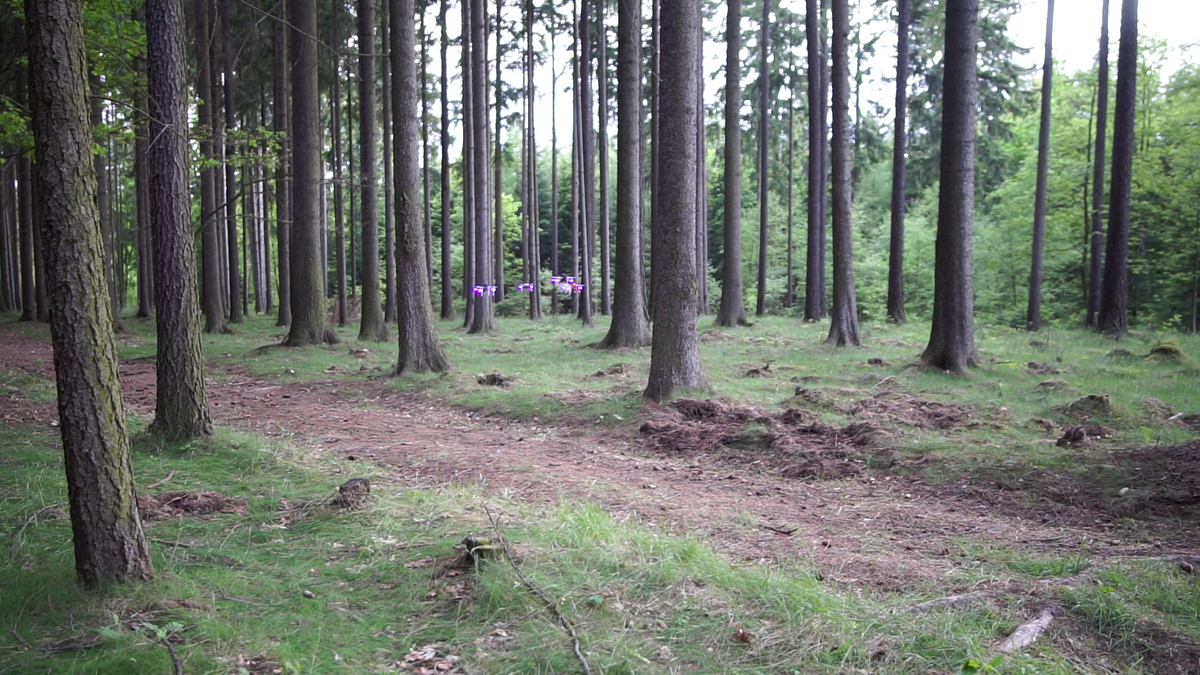}};
        \begin{scope}[x={(img.south east)},y={(img.north west)}]
          \draw [blue, solid, thick] (0.48, 0.48) circle (0.02);
          \draw [green, solid, thick] (0.49, 0.455) circle (0.0225);
          \draw [yellow, solid, thick] (0.45, 0.46) circle (0.0225);
          \draw [red, solid, thick] (0.42, 0.45) circle (0.0225);
          \node [imglabel,text=black] (label) at (img.south west) {\hspace{0.2cm}\scriptsize $\SI{266}{\second}$};
        \end{scope}
      \end{tikzpicture}
    \end{subfigure}
    \caption{Video snapshots from the real-world experiment. The system evolution at different time instances is reported. Colored solid circles highlight the~\ac{UAV} positions and their IDs according to the legend in Fig.~\ref{fig:forest_traj}. }
    \label{fig:snapshotExp}
\end{figure*}
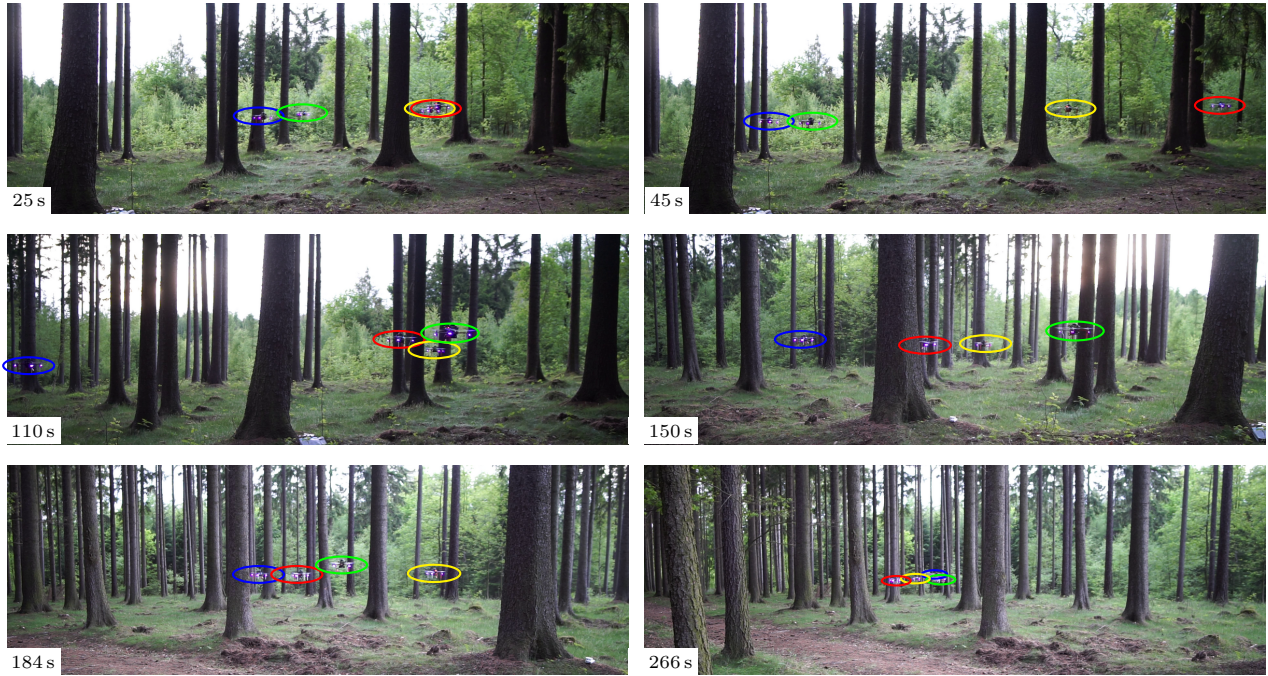

After validating the proposed approach in the simulated experiments, we performed a real-world experiment in a natural forest, as shown in Fig.~\ref{fig:snapshotExp}. The complex and random arrangement of trees in the forest serve as a challenging environment to validate the feasibility and robustness of the proposed approach.
The forest had a density of $\rho = 0.25$ (as calculated using (19)), and the~\acp{UAV} were initialized at arbitrary positions with $R^s=\SI{3.0}{\meter}$. The swarm was composed of four quadrotor~\ac{UAV} platforms (\texttt{uav1, uav2, uav3}, and \texttt{uav4}), each with a diameter of $\SI{0.5}{\meter}$ (the dimensions of a quadrotor can be approximated with a circle in the $XY$-plane) and a mass of $\SI{2.6}{\kilogram}$. The~\ac{UAV} platforms were based on the DJI F450 quadrotor, equipped with an Intel NUC on-board computer (an i7-8559U processor with 16GB of RAM) and the Pixhawk flight controller. The software stack~\cite{Baca2021mrs} was built on the Noetic Ninjemys release of the~\ac{ROS} running on Ubuntu $20.04$. The~\acp{UAV} were also equipped with a RPLIDAR rotary rangefinder for~\ac{SLAM}~\cite{Kohlbrecher2011ISSSRR}, a Garmin laser rangefinder used as an altimeter, and the~\ac{UVDAR} system~\cite{WalterIEEERAL2019} for direct relative localization of~\acp{UAV}. Further details on the hardware setup can be found in~\cite{MRS2022ICUAS_HW, Pritzl2022ICUAS_Swarm}. Fig.~\ref{fig:uav_platform} shows the sensory equipment layout on-board the~\acp{UAV}.

We selected \texttt{uav3} as the informed~\ac{UAV} and assigned the goal at $\mathbf{g} = (\SI{0}{\meter}, \SI{40}{\meter})^\top$ before the start of the experiment. The rest of the~\acp{UAV} were uninformed. Since only one of the~\acp{UAV} is informed, the experiment constitutes the worst case scenario, which is useful to test the limits of the proposed approach. 

\begin{figure}[tb]
  \includegraphics[width=0.5\textwidth, keepaspectratio]{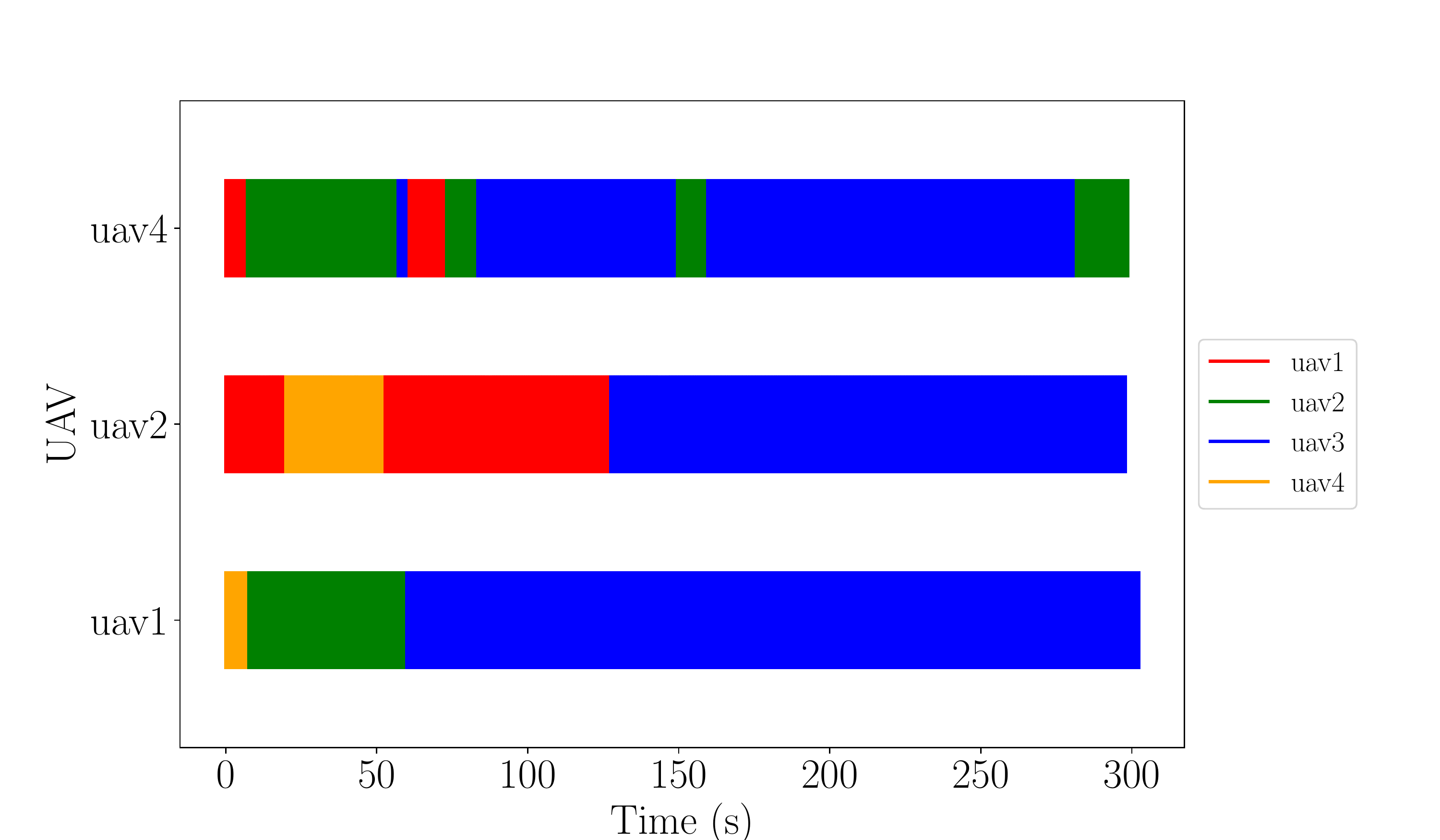}
  \caption{Target selected and tracked by each~\ac{UAV} during the real-world experiment in the forest. Colored lines represent the target (another~\ac{UAV}) selected by a particular~\ac{UAV}.}
  \label{fig:target_time}%
\end{figure}

Figure~\ref{fig:forest_traj} shows the trajectory of all four~\acp{UAV} in the forest, while Fig.~\ref{fig:snapshotExp} reports video snapshots from the real-world experiment. The marks of the~\acp{UAV} in Fig.~\ref{fig:forest_traj} correspond to the instants $\SI{0}{\second}$, $\SI{105}{\second}$, $\SI{180}{\second}$, and $\SI{300}{\second}$. At the beginning of the experiment ($\si{0}$-$\SI{105}{\second}$), all~\acp{UAV} except for \texttt{uav3}, move around to prevent any collisions with other~\acp{UAV} and surrounding trees.
As shown in Fig.~\ref{fig:target_time}, the~\acp{UAV} target each other at the beginning, but when their motion stabilizes, they select \texttt{uav3} as the target. As more and more~\acp{UAV} select \texttt{uav3} as their target, the swarm motion becomes more directed towards the goal. After $\SI{150}{\second}$, all of the~\acp{UAV} have selected and are tracking \texttt{uav3} as their target. The swarm then reaches its goal at~$\SI{300}{\second}$.

\begin{figure*}[tb]
\begin{center}
  \begin{subfigure}{0.45\textwidth}
    \adjincludegraphics[width=\textwidth, keepaspectratio, trim={{.0\width} {.0\height} {.0\width} {.1\height}}, clip]{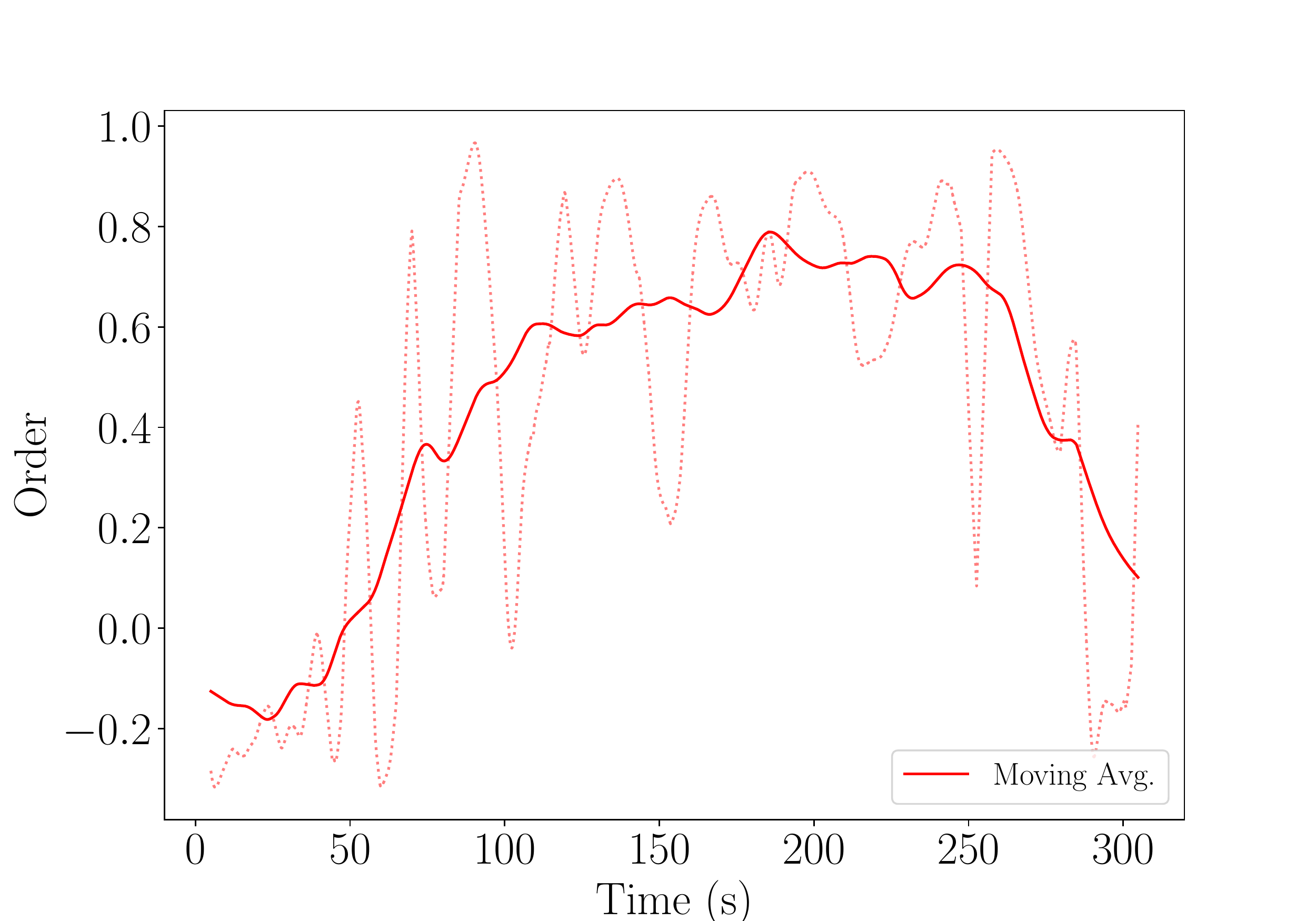}
    \caption{\texttt{uav1}.}
  \end{subfigure} 
  \hfil
  \begin{subfigure}{0.45\textwidth}
    \adjincludegraphics[width=\textwidth, keepaspectratio, trim={{.0\width} {.0\height} {.0\width} {.1\height}}, clip]{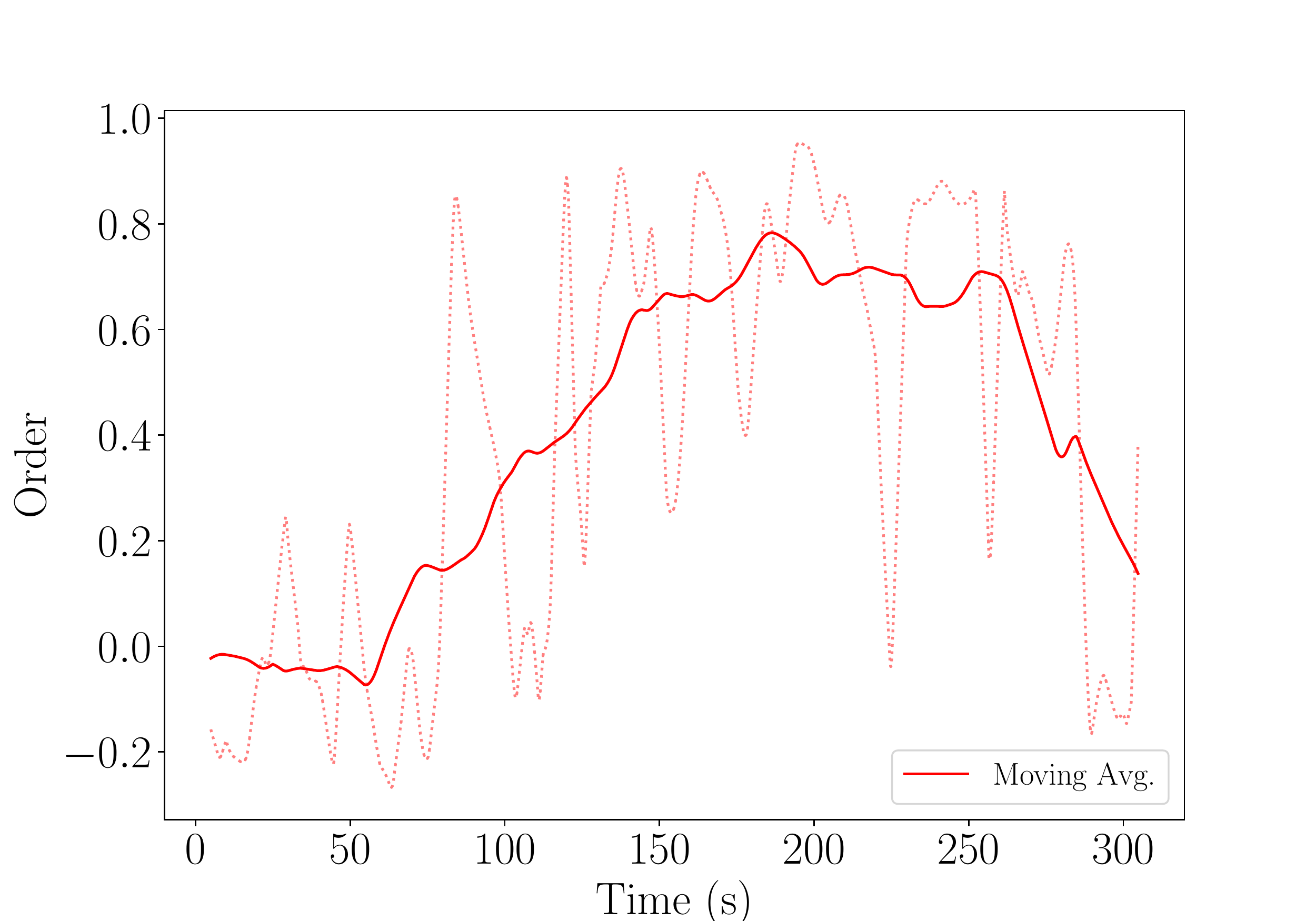}
    \caption{\texttt{uav2}.}
  \end{subfigure} 
  \\
  \begin{subfigure}{0.45\textwidth}
    \adjincludegraphics[width=\textwidth, keepaspectratio, trim={{.0\width} {.0\height} {.0\width} {.1\height}}, clip]{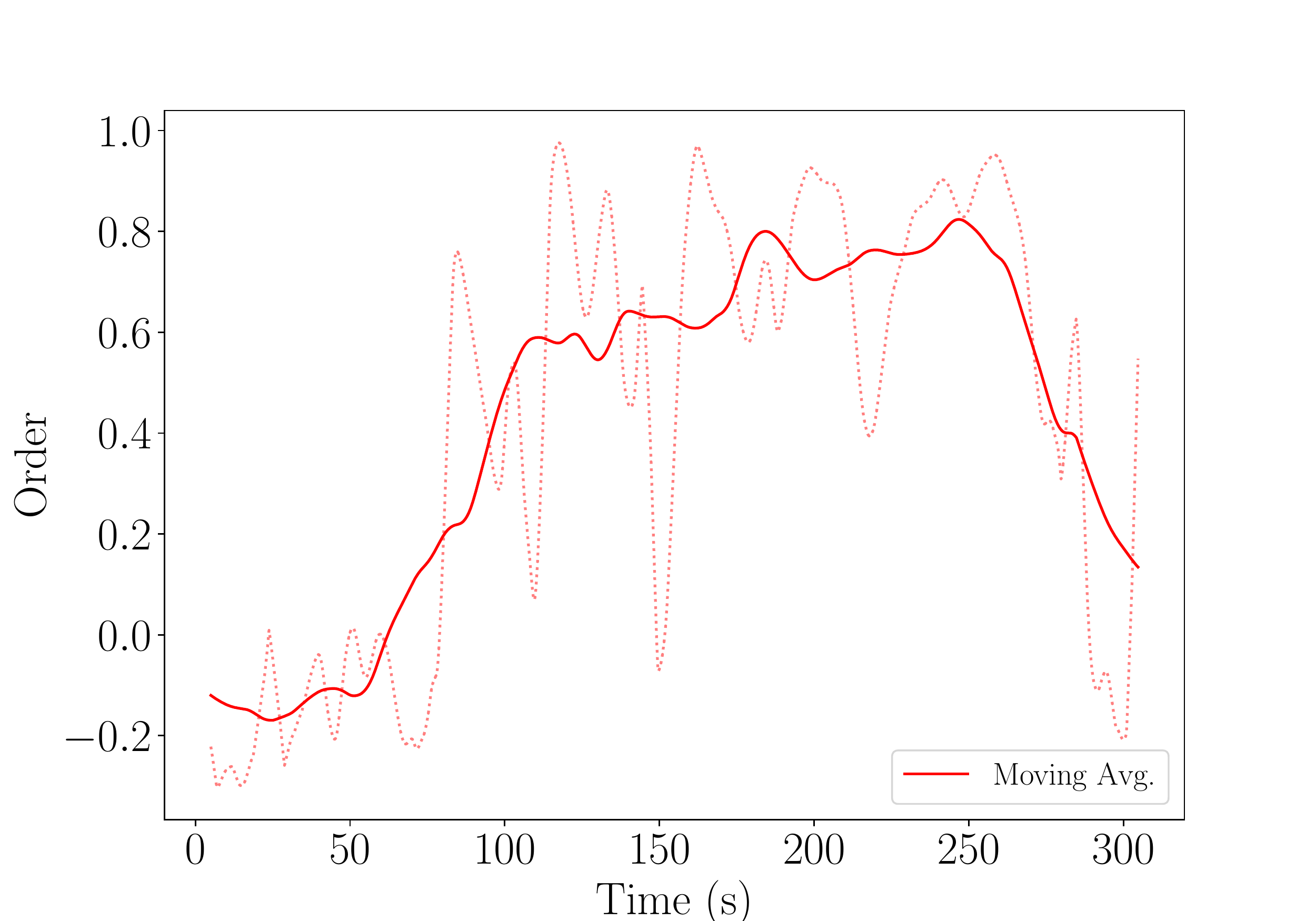}
    \caption{\texttt{uav3}.}
  \end{subfigure} 
  \hfil
  \begin{subfigure}{0.45\textwidth}
    \adjincludegraphics[width=\textwidth, keepaspectratio, trim={{.0\width} {.0\height} {.0\width} {.1\height}}, clip]{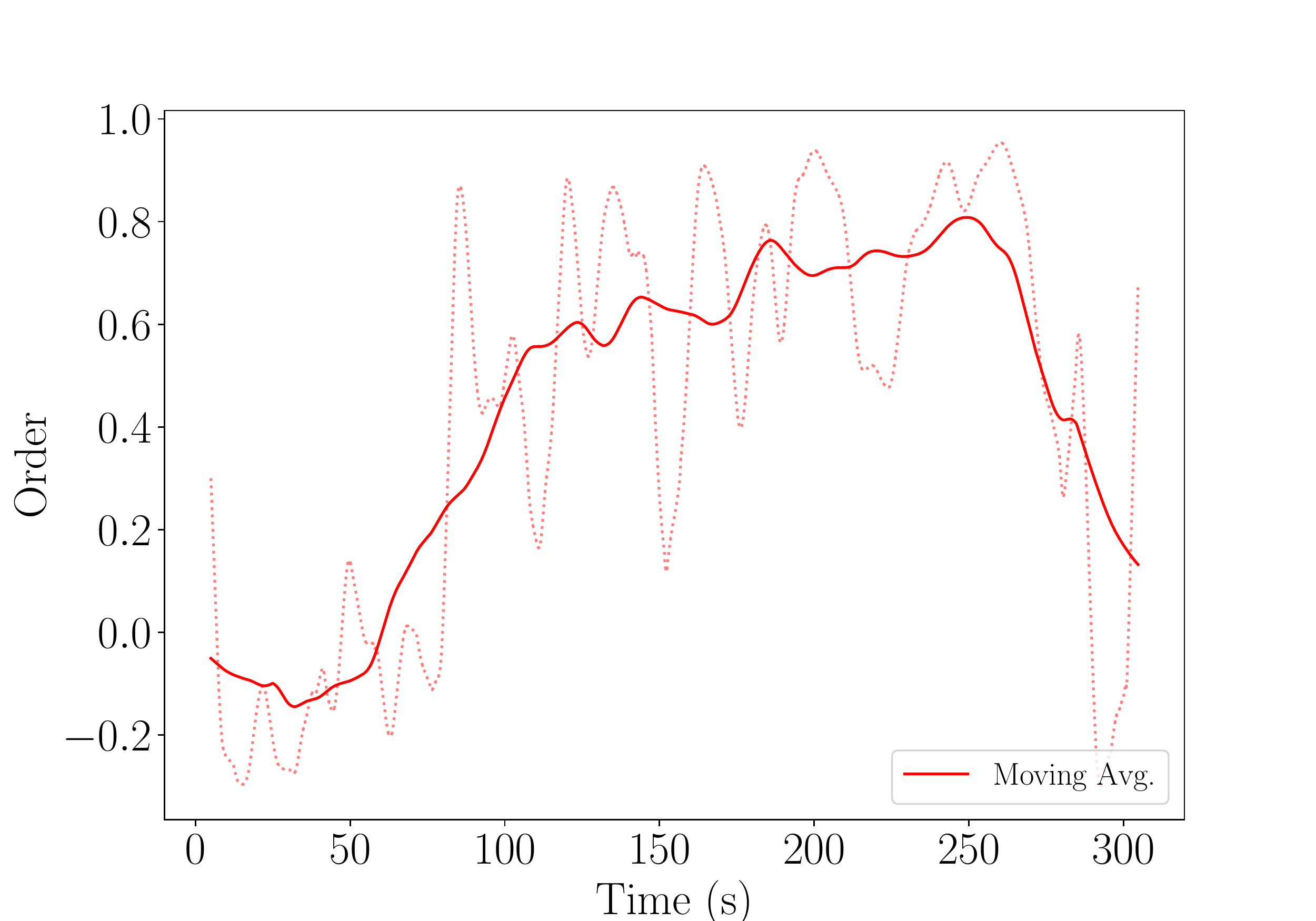}
    \caption{\texttt{uav4}.}
  \end{subfigure}
  \vspace{-0.5em}
  \caption{Recorded Order between the~\acp{UAV} during the real-world experiment. The dashed and solid lines represent the recorded order and the simple moving average, respectively.}
  \label{fig:order_time}%
  \vspace{-1.7em}
\end{center}
\end{figure*}

As shown in Figures~\ref{fig:forest_traj} and~\ref{fig:snapshotExp}, the~\acp{UAV} move through the forest as a cohesive group, while avoiding obstacles and each other. Fig.~\ref{fig:dist_time} shows the relative distance of the swarm as observed by each~\ac{UAV}. As can be seen from the figure, after the initial $\SI{105}{\second}$, the mean relative distance for \texttt{uav1} and \texttt{uav4} decreases rapidly as they select the informed~\ac{UAV} as their target. Since the mean relative distance is larger than $R^f$ for a significant part of the experiment, the uninformed~\acp{UAV} were able to select \texttt{uav3} as their target (see Fig.~\ref{fig:target_time}). Similarly, the order between the~\acp{UAV} also increases rapidly after the initial $\SI{105}{\second}$ (see Fig.~\ref{fig:order_time}). Note that, the order for all~\acp{UAV} increases to a value above $0.5$ after the initial $\SI{105}{\second}$ and remains so until $\SI{250}{\second}$. As the \acp{UAV} reach the goal near the end of the experiment, they slow down and just move around the goal position. This reduces the order drastically near the end of the experiment. This behavior can also be observed in the supplementary multimedia material. Since the order is a measure of the alignment degree of the swarm, it is apparent that the~\acp{UAV} were moving in the same direction during the experiment. This successfully demonstrates that the proposed approach can be used for collective navigation, even in the worst case scenario presented in this experiment.

The large number of trees and continuous reactive motion of other members of the swarm made the navigation challenging. While moving through the environment,~\acp{UAV} were often occluded by trees or other~\acp{UAV}. Since the~\ac{UVDAR} localization system uses cameras for relative localization, these occlusions affected its estimation accuracy. This is visible in Fig.~\ref{fig:dist_time} where the estimated distance between~\acp{UAV} has a high variance and often jumps by large values. The estimation inaccuracy also results in irregularity in the order (measure of velocity alignment) of the~\acp{UAV}, as shown in Fig.~\ref{fig:order_time}. However, as the proposed path similarity and persistence metrics (see Section~\ref{sec:sys_model}) depend on path history rather than single position estimates, the experiments show that proposed approach successfully allows the~\acp{UAV} to navigate in this challenging environment.

\begin{figure*}[tb]
\begin{center}
  \begin{subfigure}{0.45\textwidth}
    \adjincludegraphics[width=\textwidth, keepaspectratio, trim={{.0\width} {.0\height} {.0\width} {.1\height}}, clip]{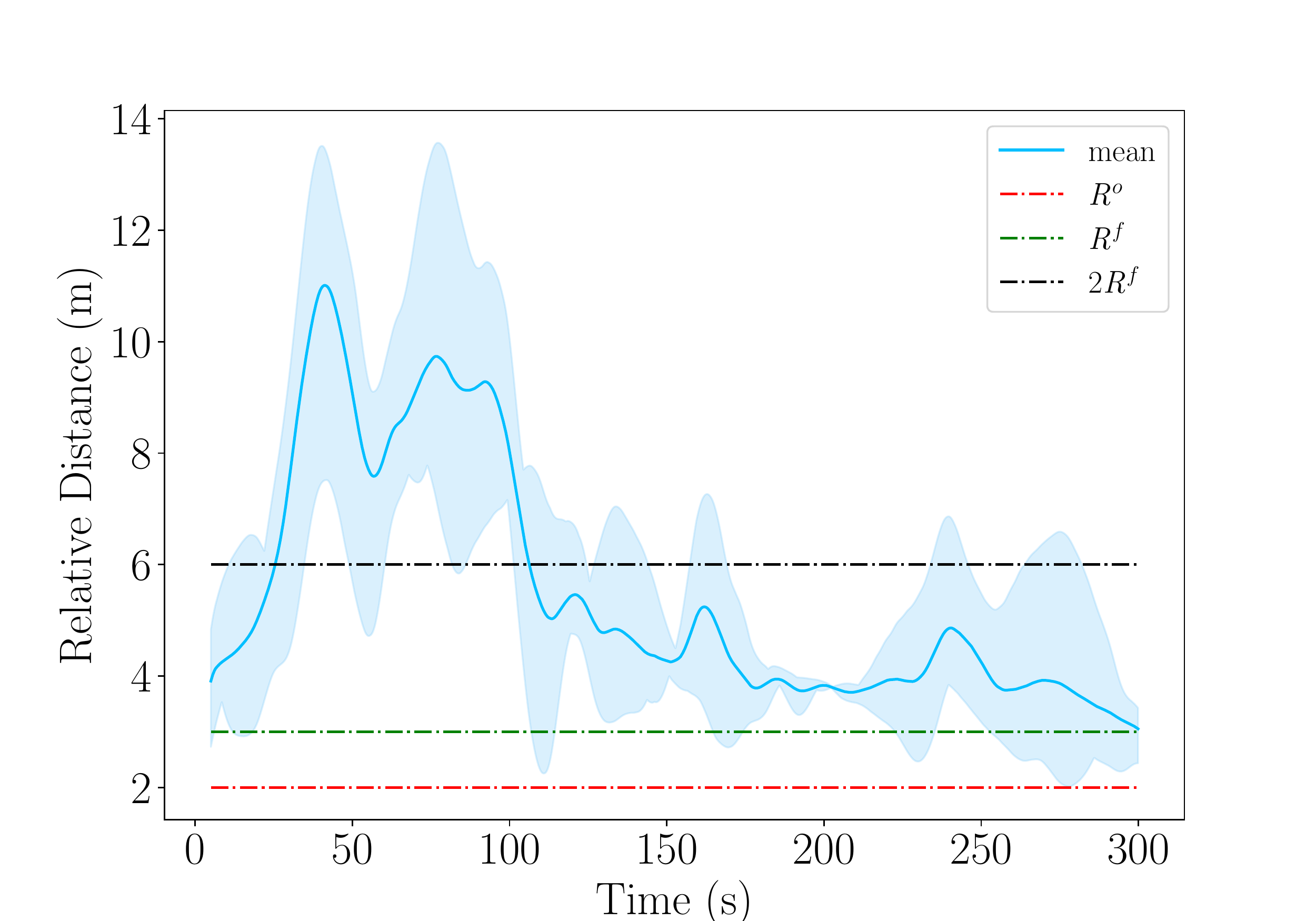}
    \caption{\texttt{uav1}.}
  \end{subfigure} 
  \hfil
  \begin{subfigure}{0.45\textwidth}
    \adjincludegraphics[width=\textwidth, keepaspectratio, trim={{.0\width} {.0\height} {.0\width} {.1\height}}, clip]{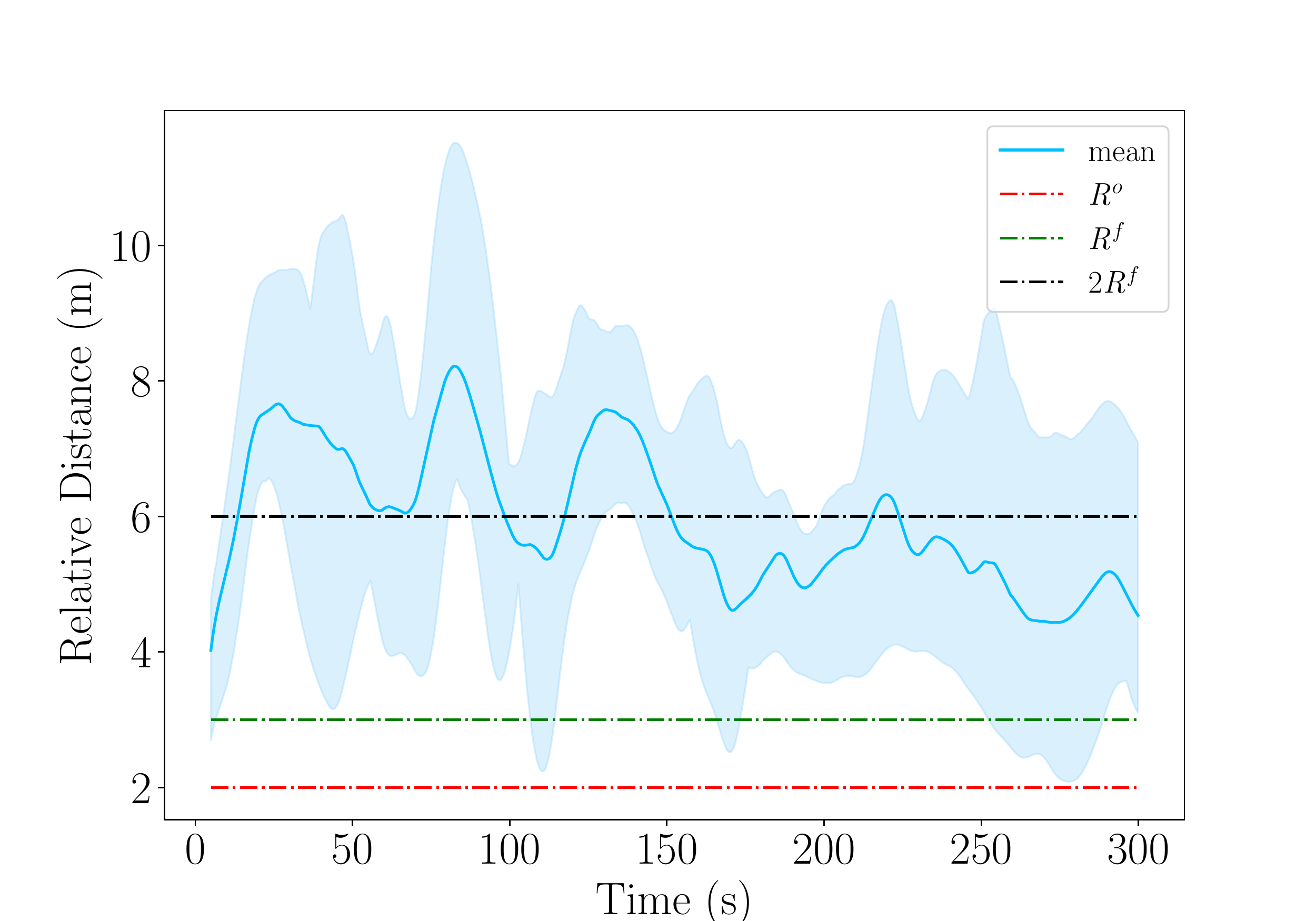}
    \caption{\texttt{uav2}.}
  \end{subfigure} 
  \medskip
  \begin{subfigure}{0.45\textwidth}
    \adjincludegraphics[width=\textwidth, keepaspectratio, trim={{.0\width} {.0\height} {.0\width} {.1\height}}, clip]{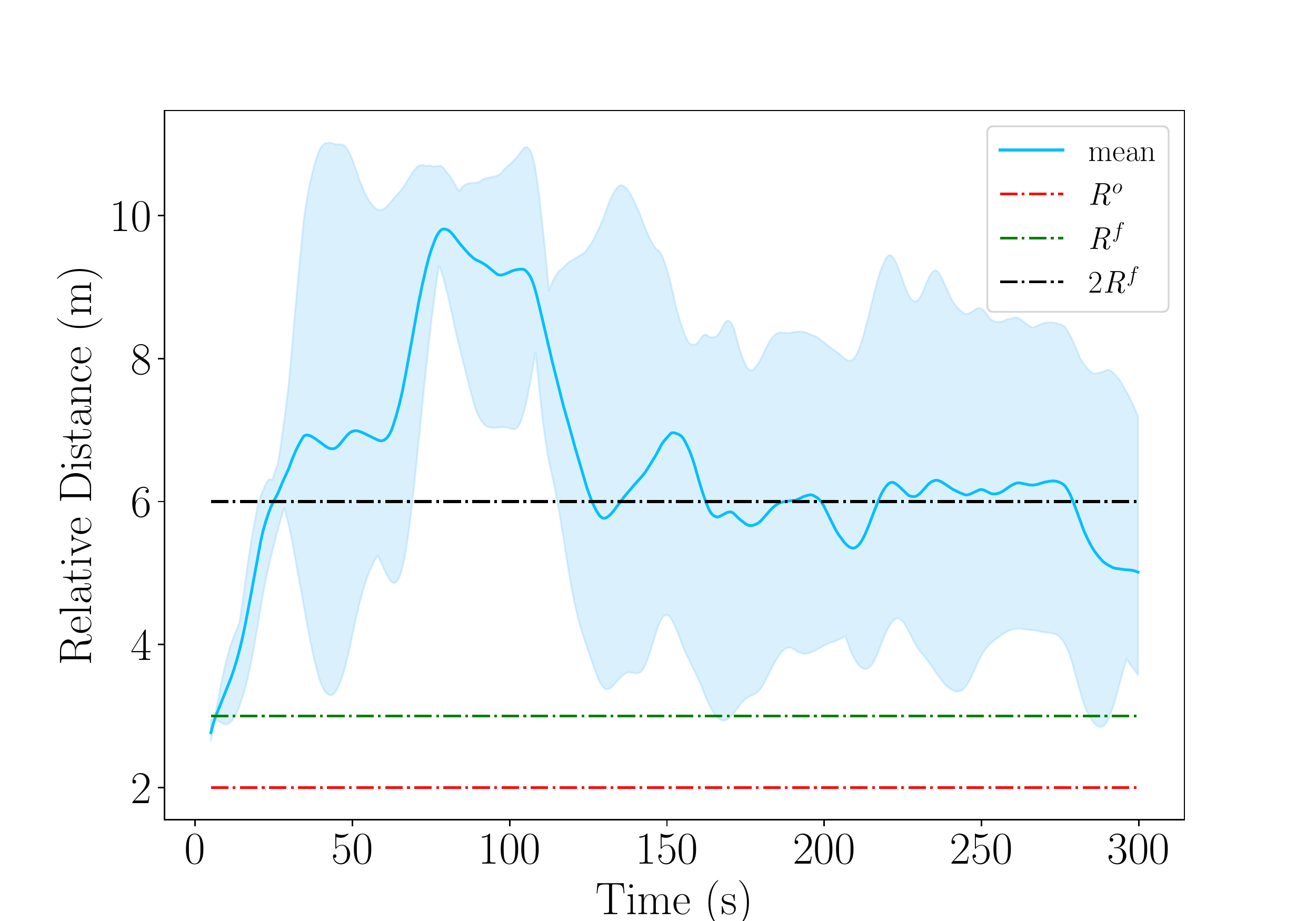}
    \caption{\texttt{uav3}.}
  \end{subfigure} 
  \hfil
  \begin{subfigure}{0.45\textwidth}
    \adjincludegraphics[width=\textwidth, keepaspectratio, trim={{.0\width} {.0\height} {.0\width} {.1\height}}, clip]{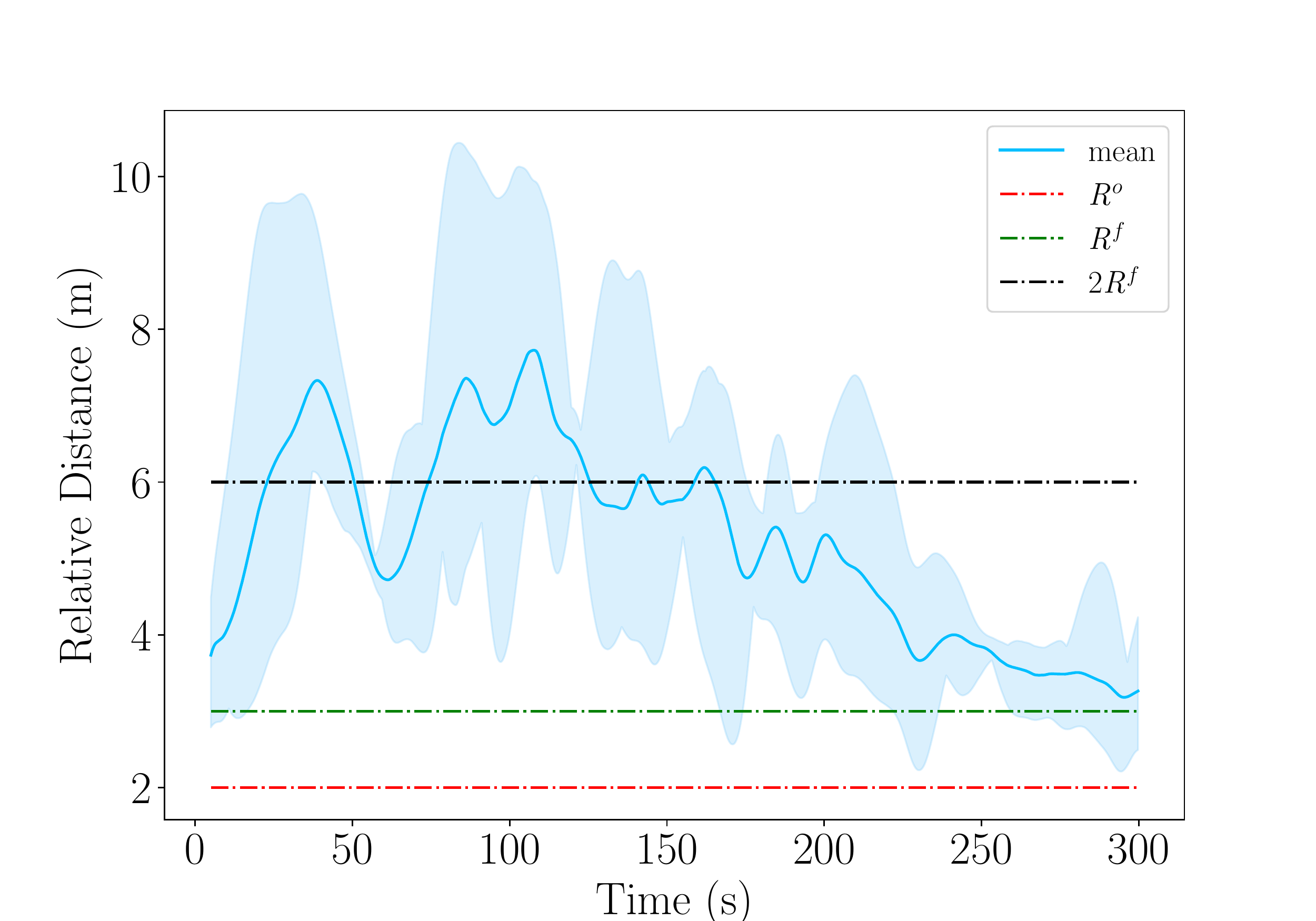}
    \caption{\texttt{uav4}.}
  \end{subfigure}
  \vspace{-1.0em}
  \caption{Recorded relative distances between the~\acp{UAV} in the real-world experiment. The distance is calculated using the position estimate from the~\ac{UVDAR} direct localization system. The shaded region contains all the distance measurements and the solid line represents their mean at any given time instant.}
  \label{fig:dist_time}%
  \vspace{-1.7em}
\end{center}
\end{figure*}

\section{Discussion}
\label{sec:discussion}
\subsection{2D vs 3D implementation}
\label{sec:2Dvs3Dimplementation}

Let us discuss the difference between the 2D and 3D implementation of the proposed approach. This article focuses on a 2D implementation of~\ac{PACNav}, although a 3D implementation would be straightforward, providing the~\acp{UAV} an additional dimension for motion and further facilitating collision avoidance. A 3D implementation of the method herein would only require expansion of dimensions for all the vector quantities. This change in dimension would result in a slight increase in the calculations involving vectors, but have no effect on the overall method. However, a 3D representation of the environment will be needed for~\ac{SLAM} and obstacle avoidance. This 3D representation would require heavy sensors and computationally expensive mapping algorithms. Moreover, 3D implementation of the $A^\star$ algorithm used for planning (Section \ref{sec:track_target:navigation}) would significantly increase the memory and computational load of the overall system when running onboard the~\ac{UAV}.

The addition of heavy sensors and more computational power would increase the size of the~\ac{UAV}, making it difficult to operate in obstacle-rich environments, such as forests. The 2D implementation of the~\ac{PACNav} approach is sufficient for solving the navigation problem in many practical scenarios (some presented in this article) while also being practically feasible. Thus, it is unclear whether the extra computational load and energy requirements for 3D implementation would justify its use.

\subsection{Scalability}
\label{sec:scalability}

In a scenario free of obstacles, there would be a consistent ~\ac{LoS} between most of the~\acp{UAV} throughout the entire mission. Therefore, each~\ac{UAV} would simply react to the motion of the other~\acp{UAV}. However, in cluttered environments, the~\acp{UAV} often loose~\ac{LoS} due to occlusions from environmental obstacles and other~\acp{UAV}. As a result, each~\ac{UAV} may only use local information (from~\acp{UAV} in~\ac{LoS}) to select a target. For a fixed number of informed~\acp{UAV}, increasing the number of uninformed~\acp{UAV} would mean that several~\acp{UAV} do not have~\ac{LoS} with any informed~\ac{UAV}. This would adversely affect the collective navigation of the swarm. 
    
For the forest simulated in our experiments, with a group size greater than six~\acp{UAV}, we observed splitting of the swarm into sub-groups.
Due to the loss of the target by some of the uninformed~\acp{UAV}, the~\acp{UAV} farther away from the informed ones were thus unable to move towards the goal, and consequently unable to guide the other uninformed~\acp{UAV} to the goal. Moreover, the uninformed~\acp{UAV} were more likely to get stuck or left behind, as they did not have an appropriate target to follow. Our experiments suggest that splitting can be avoided by composing the swarm with approximately 60 percent informed~\acp{UAV}.  

\subsection{Design parameters}
\label{sec:design_params}
The design parameters of the PACNAV approach are $R^f$, $R^o$, $K^n$, $K^c$, $K^m$, and $K^p$. In the following section, we briefly discuss their purpose and effect on the PACNAV approach.

    \begin{itemize}

        \item $R^f$ is used to discard potential target~\acp{UAV} that are too close. The $i$-th~\ac{UAV} would consider the $j$-th~\ac{UAV} as a potential target only if it is farther than the distance $R^f$. 
        
        \quad If $R^f$ is too small, most of the~\acp{UAV} close to the $i$-th~\ac{UAV} are considered potential targets. Since these~\acp{UAV} are close to each other, their trajectory is significantly influenced by the collision avoidance mechanism that increases the distance between the~\ac{UAV}. As a result, the~\acp{UAV} will not exhibit the guiding behavior that arises when the~\acp{UAV} are following a target. Thus, having a small $R^f$ in a cluttered environment can lead to frequent changes in the choice of target~\ac{UAV}, as motion due to collision avoidance will result in high path persistence for most of the potential targets. This rapid change slows down the collective movement of the swarm while the~\acp{UAV} try to follow the collision avoidance motion, rather than the goal-directed motion of the informed~\acp{UAV}.
        
        \quad Alternatively, if $R^f$ is too large, then the $i$-th~\ac{UAV} might not be able to find any~\ac{UAV} as a potential target, and the algorithm may fail.

        \item The parameter $R^o$ describes the threshold distance when reacting to the obstacles (\ref{eq:collision_avoidance_vec}). 
        
        \quad If $R^o\rightarrow\infty$, the~\ac{UAV} reacts to every single obstacle around it. These reactions can slow down movement as the~\ac{UAV} will unnecessarily try to avoid far-away obstacles.
        
        \quad However, if $R^o$ is too small, then the~\ac{UAV} will become short-sighted. Consequently, the~\ac{UAV} might not react in time to some of the surrounding obstacles, which can lead to collisions.

        \item The parameter $K^n$ is the coefficient of the navigation vector of an informed~\ac{UAV} (\ref{eq:nav_vec_informed}). 
        
        \quad When $K^n$ is large, the informed~\ac{UAV} will move to the goal with a high velocity, which may result in a loss of~\ac{LoS} with other~\acp{UAV} when operating in a cluttered environment. Moving at high velocities can also affect the stability of the~\ac{UAV} when it tries to avoid collisions with obstacles.
        
        However, when $K^n$ is too small, the~\ac{UAV} will move more slowly, which in turn, slows down the entire swarm.

        \item The parameter $K^c$ is the coefficient of the collision avoidance vector $\mathbf{c}$ (\ref{eq:collision_avoidance}). 
        Small values of $K^c$ would result in insufficient and slow reaction to obstacles, risking collisions.
        %
        Large values of $K^c$ would result in violent reaction of the~\ac{UAV} in the presence of obstacles. In the worst case (just as in the case of $K^n$), this could negatively affect the stability of the~\ac{UAV}.

        \item $K^m$ is introduced in Algorithm~\ref{alg:neighbor_update}. When the $i$-th~\ac{UAV} loses~\ac{LoS} with the $j$-th~\ac{UAV} for more than $K^m$ time instants, the $j$-th~\ac{UAV} is discarded from the set $\mathcal{N}_i$.
        
        \quad When the value of $K^m$ is large, the~\acp{UAV} will be part of the set $\mathcal{N}_i$ long after~\ac{LoS} has been lost. Thus, the position information of these~\acp{UAV} will also be outdated. As the navigation vector depends on the number of~\acp{UAV} in $\mathcal{N}_i$, the $i$-th~\ac{UAV} will react to the outdated position information, which can cause unnecessary delays in the collective motion of the entire swarm.
        
        \quad In contrast, when $K^m$ is small, the $i$-th~\ac{UAV} will remove the~\acp{UAV} as soon as the~\ac{LoS} is lost. As the environment is cluttered with obstacles, the loss of~\ac{LoS} would be frequent. Thus, at any time $k$, the $i$-th~\ac{UAV} will only have a few~\acp{UAV} in $\mathcal{N}_i$, which in turn reduces the options for potential targets. As a result, having a small $K^m$ can severely disrupt the target selection process and might produce erratic trajectories.
        
        \item $K^p$ is introduced in Algorithm~\ref{alg:path_hist}. It determines the length of the path history stored for each~\ac{UAV}. This length is crucial for the path similarity (\ref{eq:path_sim}) and path persistence (\ref{eq:path_pers}) metrics. The motion of most of the~\acp{UAV} is similar on very small and very large timescales. As these metrics are calculated on normalized vector quantities, the observed path histories of multiple~\acp{UAV} can have similar values for metrics. Thus, for both small and large values of $K^p$, the metrics would not reflect the path information, resulting in a frequent change in the selected target. When a~\ac{UAV} is frequently switching between target~\acp{UAV}, the overall motion of the swarm can be slowed down.
        
    \end{itemize}


\section{Conclusion}
\label{sec:conclusion}

This article has presented~\ac{PACNav} as a new bioinspired decentralized approach for navigating a~\ac{UAV} swarm to the desired goal as a compact group. In contrast to state-of-the-art methods, the presented approach does not require communication among the members or a global localization system. Such an approach is highly beneficial in demanding real-world conditions where global localization is unavailable or has a high uncertainty and the swarm size makes communication unfeasible. In the presented approach, each~\ac{UAV} determines its future motion using the information derived from only on-board sensors, mimicking the sensory organs of animals moving in a group. The resultant decentralized swarm is scalable as it is not limited by communication bandwidth and information sharing. 
The navigation method based on the metrics derived from the collective motion of animals in nature ensures coherent movement of the swarm and collision avoidance with the environment and other members. The simulated experiments in the Gazebo robotic simulator and a real-world flight in a natural forest validated the effectiveness of the presented approach. An intensive analysis carried out in simulation demonstrated the reliability of the algorithm concerning changes in the number of informed~\acp{UAV}, as well as different spatial distribution of trees in the forest. The software framework used to deploy the algorithm on a decentralized swarm of~\acp{UAV} is provided as open-source\footnoteref{fotnote:code} in order to facilitate further research and replication of the obtained results.





\section*{Acknowledgment}
\label{sec:acknowledgment}

This work was partially supported by the Czech Science Foundation (GA\v{C}R) under research project no. 20-10280S, by the European Union's Horizon 2020 research and innovation programme AERIAL-CORE under grant agreement no. 871479, by CTU grant no. SGS20/174/OHK3/3T/13, by the Technology Innovation Institute - Sole Proprietorship LLC, UAE, and by OP VVV project CZ.02.1.01/0.0/0.0/16 019/0000765 ``Research Center for Informatics". The authors would like to thank Daniel He\v{r}t for his help with the experimental setup.







\section*{ORCID}
Afzal Ahmad$^{\orcidicon{0000-0002-5889-0320}}$: 0000-0002-5889-0320\hfill\\
Daniel Bonilla Licea$^{\orcidicon{0000-0002-1057-816X}}$: 
0000-0002-1057-816X\hfill\\
Giuseppe Silano$^{\orcidicon{0000-0002-6816-6002}}$: 0000-0002-6816-6002\hfill\\
Tom\'{a}\v{s} B\'{a}\v{c}a$^{\orcidicon{0000-0001-9649-8277}}$: 0000-0001-9649-8277\hfill\\
Martin Saska$^{\orcidicon{0000-0001-7106-3816}}$: 0000-0001-7106-3816\hfill

\begin{acronym}
    \acro{CNN}[CNN]{Convolutional Neural Network}
    \acro{DBSCAN}[DBSCAN]{Density-based Spatial Clustering of Applications with Noise}
    \acro{EKF}[EKF]{Extended Kalman Filter}
    \acro{FIFO}[FIFO]{First-In-First-Out}
    \acro{FOV}[FoV]{Field of View}
    \acro{GNSS}[GNSS]{Global Navigation Satellite System}
    \acro{ICNIRP}[ICNIRP]{International Commission on Non-Ionizing Radiation Protection}
    \acro{IMU}[IMU]{Inertial Measurement Unit}
    \acro{IR}[IR]{InfraRed}
    \acro{LiDAR}[LiDAR]{Light Detection and Ranging}
    \acro{LoS}[LoS]{Line of Sight}
    \acro{MAV}[MAV]{Micro Aerial Vehicle}
    \acro{ML}[ML]{Machine Learning}
    \acro{MPC}[MPC]{Model Predictive Control}
    \acro{MOCAP}[MoCap]{Motion Capture}
    \acro{PACNav}[PACNav]{Persistence Administered Collective Navigation}
    \acro{RMSE}[RMSE]{Root Mean Square Error}
    \acro{ROS}[ROS]{Robot Operating System}
    \acro{RTK}[RTK]{Real-Time Kinematic}
    \acro{SLAM}[SLAM]{Simultaneous Localization and Mapping}
    \acro{UAV}[UAV]{Unmanned Aerial Vehicle}
    \acro{UV}[UV]{UltraViolet}
    \acro{UVDAR}[UVDAR]{UltraViolet Direction And Ranging}
    \acro{UT}[UT]{Unscented Transform}
    \acro{UWB}[UWB]{Ultra-Wideband ranging}
\end{acronym}



\section*{References}
\bibliographystyle{iopart-num}
\bibliography{main}

\providecommand{\newblock}{}
\begin{thebibliography}{10}
\expandafter\ifx\csname url\endcsname\relax
  \def\url#1{{\tt #1}}\fi
\expandafter\ifx\csname urlprefix\endcsname\relax\def\urlprefix{URL }\fi
\providecommand{\eprint}[2][]{\url{#2}}

\bibitem{ZhouScienceRobotics2022}
{Zhou} X, {Wen} X, {Wang} Z, {Gao} Y, {Li} H, {Wang} Q, {Yang} T, {Lu} H, {Cao}
  Y, {Xu} C and {Gao} F 2022 {Swarm of micro flying robots in the wild} {\em
  Science Robotics\/} {\bf 7} eabm5954

\bibitem{kumar_2018_tro}
{Chung} S, {Paranjape} A~A, {Dames} P, {Shen} S and {Kumar} V 2018 {A Survey on
  Aerial Swarm Robotics} {\em IEEE Transactions on Robotics\/} {\bf 34}
  837--855

\bibitem{Honig2018TRO}
Hönig W, Preiss J~A, Kumar T~K~S, Sukhatme G~S and Ayanian N 2018 {Trajectory
  Planning for Quadrotor Swarms} {\em IEEE Transactions on Robotics\/} {\bf 34}
  856--869

\bibitem{Augugliaro2014CSM}
Augugliaro F, Lupashin S, Hamer M, Male C, Hehn M, Mueller M~W, Willmann J~S,
  Gramazio F, Kohler M and D'Andrea R 2014 {The Flight Assembled Architecture
  installation: Cooperative construction with flying machines} {\em IEEE
  Control Systems Magazine\/} {\bf 34} 46--64

\bibitem{ElmokademIFAC2019}
{Elmokadem} T 2019 {Distributed Coverage Control of Quadrotor Multi-UAV Systems
  for Precision Agriculture} {\em IFAC-PapersOnLine\/} vol~52 pp 251--256

\bibitem{McGuireScienceRobotics2019}
{McGuire} K~N, {De Wagter} C, Tuyls K, {Kappen} H~J and {de Croon} G~C~H~E 2019
  {Minimal navigation solution for a swarm of tiny flying robots to explore an
  unknown environment} {\em Science Robotics\/} {\bf 4} eaaw9710

\bibitem{tagliabue_2019_sage}
Tagliabue A, Kamel M, Siegwart R and Nieto J 2019 {Robust collaborative object
  transportation using multiple MAVs} {\em The International Journal of
  Robotics Research\/} {\bf 38} 1020--1044

\bibitem{Inada2010IFAC}
{Inada} Y and {Takanobu} H 2010 {Flight-Formation Control of Air Vehicles Based
  on Collective Motion Control of Organisms} {\em IFAC-PapersOnLine\/} vol~43
  pp 386--391

\bibitem{Novak2021Bioinspired}
{Novak} F, {Walter} V, {Petr{\'{a}}{\v{c}}ek} P, {B{\'{a}}{\v{c}}a} T and
  {Saska} M 2021 {Fast collective evasion in self-localized swarms of unmanned
  aerial vehicles} {\em Bioinspiration {\&} Biomimetics\/} {\bf 16} 026009

\bibitem{rabbath_2012_autocontrol}
{Chamseddine} A, {Zhang} Y and {Rabbath} C~A 2012 {Trajectory planning and
  re-planning for fault tolerant formation flight control of quadrotor unmanned
  aerial vehicles} {\em America Control Conference\/} pp 3291--3296

\bibitem{petracek_2020_ral}
{Petr\'{a}\v{c}ek} P, {Kr\'{a}tk\'{y}} V and {Saska} M 2020 {Dronument: System
  for Reliable Deployment of Micro Aerial Vehicles in Dark Areas of Large
  Historical Monuments} {\em IEEE IEEE Robotics and Automation Letters\/} {\bf
  5} 2078--2085

\bibitem{vicsek_2010_nature}
Nagy M, {\'A}kos Z, Biro D and Vicsek T 2010 {Hierarchical group dynamics in
  pigeon flocks} {\em Nature\/} {\bf 464} 890--893

\bibitem{couzin_2007_nature}
Couzin I 2007 {Collective minds} {\em Nature\/} {\bf 445} 715--715

\bibitem{yomosa_2015_plos}
Yomosa M, Mizuguchi T, V{\'a}s{\'a}rhelyi G and Nagy M 2015 {Coordinated
  behaviour in pigeon flocks} {\em Plos one\/} {\bf 10} e0140558

\bibitem{reynolds_boids}
Reynolds C~W 1987 {Flocks, Herds and Schools: A Distributed Behavioral Model}
  {\em Proceedings of the 14th Annual Conference on Computer Graphics and
  Interactive Techniques\/} vol~21 pp 25--34

\bibitem{olfati_2006_tac}
{Olfati-Saber} R 2006 {Flocking for multi-agent dynamic systems: algorithms and
  theory} {\em IEEE Transactions on Automatic Control\/} {\bf 51} 401--420

\bibitem{cheng_2017_iros}
Cheng H, Zhu Q, Liu Z, Xu T and Lin L 2017 {Decentralized navigation of
  multiple agents based on ORCA and model predictive control} {\em IEEE/RSJ
  International Conference on Intelligent Robots and Systems\/} pp 3446--3451

\bibitem{gao_2021_icra}
Zhou X, Zhu J, Zhou H, Xu C and Gao F 2021 {EGO-Swarm: A Fully Autonomous and
  Decentralized Quadrotor Swarm System in Cluttered Environments} {\em IEEE
  International Conference on Robotics and Automation\/} pp 4101--4107

\bibitem{vicek_2018_scirob}
V{\'a}s{\'a}rhelyi G, Vir{\'a}gh C, Somorjai G, Nepusz T, Eiben A~E and Vicsek
  T 2018 {Optimized flocking of autonomous drones in confined environments}
  {\em Science Robotics\/} {\bf 3}

\bibitem{afzal_2021_icra}
{Ahmad} A, {Walter} V, {Petracek} P, {Petrlik} M, {Baca} T, {Zaitlik} D and
  {Saska} M 2021 {Autonomous Aerial Swarming in GNSS-denied Environments with
  High Obstacle Density} {\em IEEE International Conference on Robotics and
  Automation\/} pp 570--576

\bibitem{Dmytruk2021ICUAS}
Dmytruk A, Nascimento T, Ahmad A, Báča T and Saska M 2021 {Safe
  Tightly-Constrained UAV Swarming in GNSS-denied Environments} {\em
  International Conference on Unmanned Aircraft Systems\/} pp 1391--1399

\bibitem{Petracek2021RAL}
Petráček P, Krátký V, Petrlík M, Báča T, Kratochvíl R and Saska M 2021
  {Large-Scale Exploration of Cave Environments by Unmanned Aerial Vehicles}
  {\em IEEE Robotics and Automation Letters\/} {\bf 6} 7596--7603

\bibitem{wang_2017_aaai}
{Wang} Q, {Chen} M and {Li} X 2017 {Quantifying and Detecting Collective Motion
  by Manifold Learning} {\em AAAI Conference on Artificial Intelligence\/} pp
  4292--4298

\bibitem{swarm_survey2}
Anam T, Jari B, Mohammad-Hashem H, Hannu~T T and Juha P 2019 {Swarms of
  Unmanned Aerial Vehicles — A Survey} {\em Journal of Industrial Information
  Integration\/} {\bf 16} 100106

\bibitem{swarm_survey_mean_field}
Karthik E and Spring B 2019 {Mean-field models in swarm robotics: a survey}
  {\em Bioinspiration {\&} Biomimetics\/} {\bf 15} 015001

\bibitem{lablike_swarm1}
Verginis C~K, Xu Z and Dimarogonas D~V 2017 {Decentralized motion planning with
  collision avoidance for a team of UAVs under high level goals} {\em IEEE
  International Conference on Robotics and Automation\/} pp 781--787

\bibitem{lablike_swarm2}
Cotsakis R, St-Onge D and Beltrame G 2019 {Decentralized collaborative
  transport of fabrics using micro-UAVs} {\em IEEE International Conference on
  Robotics and Automation\/} pp 7734--7740

\bibitem{lablike_swarm3}
Kushleyev A, {Mellinger} D, {Powers} C and {Kumar} V 2013 {Towards A Swarm of
  Agile Micro Quadrotors} {\em Autonomous Robots\/} {\bf 35} 287--300

\bibitem{non_obst_swarm1}
Schilling F, Schiano F and Floreano D 2021 {Vision-Based Drone Flocking in
  Outdoor Environments} {\em IEEE Robotics and Automation Letters\/} {\bf 6}
  2954--2961

\bibitem{comm_swarm}
Dai F, Chen M, Wei X and Wang H 2019 {Swarm Intelligence-Inspired Autonomous
  Flocking Control in UAV Networks} {\em IEEE Access\/} {\bf 7} 61786--61796

\bibitem{vio_swarm_kumar}
{Weinstein} A, {Cho} A, {Loianno} G and {Kumar} V 2018 {Visual Inertial
  Odometry Swarm: An Autonomous Swarm of Vision-Based Quadrotors} {\em IEEE
  Robotics and Automation Letters\/} {\bf 3} 1801--1807

\bibitem{pavel_swarm}
Petr{\'{a}}{\v{c}}ek P, Walter V, B{\'{a}}{\v{c}}a T and Saska M 2020
  {Bio-inspired compact swarms of unmanned aerial vehicles without
  communication and external localization} {\em Bioinspiration {\&}
  Biomimetics\/} {\bf 16} 026009

\bibitem{WalterIEEECASE2018}
Walter V, Staub N, Saska M and Franchi A 2018 {Mutual Localization of UAVs
  based on Blinking Ultraviolet Markers and 3D Time-Position Hough Transform}
  {\em IEEE 14th International Conference on Automation Science and
  Engineering\/} pp 298--303

\bibitem{WalterIEEERAL2019}
Walter V, Staub N, Franchi A and Saska M 2019 {UVDAR System for Visual Relative
  Localization With Application to Leader–Follower Formations of Multirotor
  UAVs} {\em IEEE Robotics and Automation Letters\/} {\bf 4} 2637--2644

\bibitem{VrbaIEEERAL2020}
Vrba M and Saska M 2020 {Marker-Less Micro Aerial Vehicle Detection and
  Localization Using Convolutional Neural Networks} {\em IEEE Robotics and
  Automation Letters\/} {\bf 5} 2459--2466

\bibitem{Horyna2022ICUAS_UVDAR}
{Horyna} J, {Walter} V and {Saska} M 2022 {UVDAR-COM: UV-Based Relative
  Localization of UAVs with Integrated Optical Communication} {\em
  International Conference on Unmanned Aircraft Systems\/} pp 1341--1347

\bibitem{Kohlbrecher2011ISSSRR}
Kohlbrecher S, von Stryk O, Meyer J and Klingauf U 2011 {A flexible and
  scalable SLAM system with full 3D motion estimation} {\em IEEE International
  Symposium on Safety, Security, and Rescue Robotics\/} pp 155--160

\bibitem{orig_astar}
Hart P~E, Nilsson N~J and Raphael B 1968 {A Formal Basis for the Heuristic
  Determination of Minimum Cost Paths} {\em IEEE Transactions on Systems
  Science and Cybernetics\/} {\bf 4} 100--107

\bibitem{pot_obst_avoid}
Ma L, Bao W, Zhu X, Wu M, Wang Y, Ling Y and Zhou W 2020 {O-Flocking: Optimized
  Flocking Model on Autonomous Navigation for Robotic Swarm} {\em Advances in
  Swarm Intelligence\/} pp 628--639

\bibitem{gazebo_2004_iros}
{Koenig} N and {Howard} A 2004 {Design and Use Paradigms for Gazebo, An
  Open-Source Multi-Robot Simulator} {\em IEEE/RSJ International Conference on
  Intelligent Robots and Systems\/} pp 2149--2154

\bibitem{SilanoSCM19}
Silano G, Oppido P and Iannelli L 2019 {Software-in-the-loop simulation for
  improving flight control system design: a quadrotor case study} {\em IEEE
  International Conference on Systems, Man and Cybernetics\/} pp 466--471

\bibitem{order_def}
Soria E, Schiano F and Floreano D 2020 {SwarmLab: a Matlab Drone Swarm
  Simulator} {\em IEEE/RSJ International Conference on Intelligent Robots and
  Systems\/} pp 8005--8011

\bibitem{Baca2021mrs}
Baca T, Petrlik M, Vrba M, Spurny V, Penicka R, Hert D and Saska M 2021 {The
  MRS UAV System: Pushing the Frontiers of Reproducible Research, Real-world
  Deployment, and Education with Autonomous Unmanned Aerial Vehicles} {\em
  {Journal of Intelligent {\&} Robotic Systems}\/} {\bf 102} 1--28

\bibitem{MRS2022ICUAS_HW}
{Hert} D, {Baca} T, {Petracek} P, {Kratky} V, {Spurny} V, {Petrilik} M,
  {Matous} V, {Zaitlik} D, {Stoudek} P, {Walter} V, {Stepan} P, {Horyna} J,
  {Ptrizl} V, {Silano} G, {Bonilla Licea} D, {Stibinger} P, {Penicka} R,
  {Nascimento} T and {Saska} M 2022 {MRS Modular UAV Hardware Platforms for
  Supporting Research in Real-World Outdoor and Indoor Environments} {\em
  International Conference on Unmanned Aircraft Systems\/} pp 1303--1312

\bibitem{Pritzl2022ICUAS_Swarm}
{Pritzl} V, {Vrba} M, {Stepan} P and {Saska} M 2022 {Cooperative Navigation and
  Guidance of a Micro-Scale Aerial Vehicle by an Accompanying UAV using 3D
  LiDAR Relative Localization} {\em International Conference on Unmanned
  Aircraft Systems\/} pp 517--526

\end{thebibliography}

\end{document}